\documentclass[10pt,twocolumn,letterpaper]{article}

\usepackage[pagenumbers]{cvpr} 

\newcommand{\gf}[2]{{\color{gray}#1}{\color{purple}#2}}

\usepackage[normalem]{ulem} 

\usepackage{booktabs}
\usepackage{pifont}
\newcommand{\cmark}{\ding{51}}
\newcommand{\xmark}{\ding{55}}
\newcommand{\pmark}{$\sim$}
\newcommand{\promptfont}{\fontfamily{lmtt}\selectfont\frenchspacing} 
\newcommand{\prompt}[1]{{\promptfont "#1"}} 
\newcommand{\ph}[1]{{\promptfont\char`\{}\textrm{\textit{#1}}{\promptfont\char`\}}} 
\newcommand{\tok}[1]{\mbox{\prompt{\textvisiblespace #1}}} 
\newcommand{\concept}[1]{\ensuremath{\langle}#1\ensuremath{\rangle}} 
\usepackage{booktabs, multirow, graphicx}
\usepackage[table]{xcolor}
\usepackage{twemojis} 
\usepackage{tikz}
\newcommand{\iconline}{\tikz[baseline=-0.5ex]\draw[line width=1.6pt] (0,0)--(0.35,0);} 
\newcommand{\iconrectangle}{\tikz[baseline=0.2ex]\fill (0,0) rectangle (0.4,0.15);}
\newcommand{\iconsquare}{\tikz[baseline=0.2ex]\fill (0,0) rectangle (0.22,0.22);}
\definecolor{nosteercolor}{rgb}{0.773,0.901,0.761} 
\definecolor{steercolor}{rgb}{0.901,0.772,0.761} 
\newcommand{\nosteerhl}[1]{{\setlength{\fboxsep}{1pt}\colorbox{nosteercolor}{#1}}} 
\newcommand{\steerhl}[1]{{\setlength{\fboxsep}{1pt}\colorbox{steercolor}{#1}}} 
\usepackage{comment}

\makeatletter
\renewcommand{\paragraph}{\@startsection{paragraph}{4}{\z@}%
  {1ex}
  {-1em}
  {\normalfont\normalsize\bfseries}}
\makeatother

\definecolor{cvprblue}{rgb}{0.21,0.49,0.74}
\usepackage[pagebackref,breaklinks,colorlinks,allcolors=cvprblue]{hyperref}

\def\confName{CVPR}
\def\confYear{2027}

\title{
Decoder-as-a-Probe: 
Verbalizing and Steering 
Visual Representations 
in LVLMs 
}

\author{
\textbf{Adhemar de Senneville$^{1}$ \quad Xavier Bou$^{2}$ \quad J\'er\'emy Anger$^{1}$}\\
\textbf{Rafael Grompone$^{1}$ \quad Gabriele Facciolo$^{1,3}$}\\[0.7em]
\textnormal{$^{1}$Universit\'e Paris-Saclay, CNRS, ENS Paris-Saclay, Centre Borelli, Paris, France}\\
\textnormal{$^{2}$P\^ole recherche de l'AMIAD, Palaiseau, France}\\
\textnormal{$^{3}$Institut Universitaire de France, Paris, France}
}

\begin{document}
\maketitle


\begin{abstract}
It is commonly accepted that high-level concepts are represented linearly in the representation space of large neural networks. 
For Large Vision Language Models (LVLMs), finding those interpretable directions in visual representations is important for interpretability. 
To find them, the classical approach is to train a linear classifier on the representations, 
but it needs a large amount of labels and
only recovers what can be read from the representations, not what the model actually uses.
In this paper we start from the following insight:
The LLM Decoder of an LVLM is also a classifier whose classes are the possible next tokens.
So we use its first-order Taylor expansion as a linear classifier.
Hence, we introduce Decoder-as-a-Probe (DaP), which significantly outperforms classical probing and steering baselines especially when training data is sparse.
(1) It verbalizes what an LVLM perceives in a given image.
(2) It can surgically change the perception of the LVLM about specific attributes (e.g. change only the color, the clothes worn or the action done).
Due to its simplicity and data efficiency, we believe DaP can serve as a plug-and-play tool for LVLM research, such as interpretability, hallucination mitigation or instruction tuning.
\end{abstract}

\section{Introduction}
\label{sec:intro}

%
Representation learning is one of the great successes of computer vision, with models such as CLIP~\cite{radford2021clip}. 
These vision encoders are now the eyes of Large Vision Language Models (LVLMs).
Indeed, in an LVLM, CLIP visual representations of the image are given to a Large Language Model (LLM Decoder) to answer a wide range of questions about it in natural language~\cite{liu2023llava, qwen2025qwen3vl}. 

\begin{figure}[t] 
    \centering
    \includegraphics[width=\linewidth]{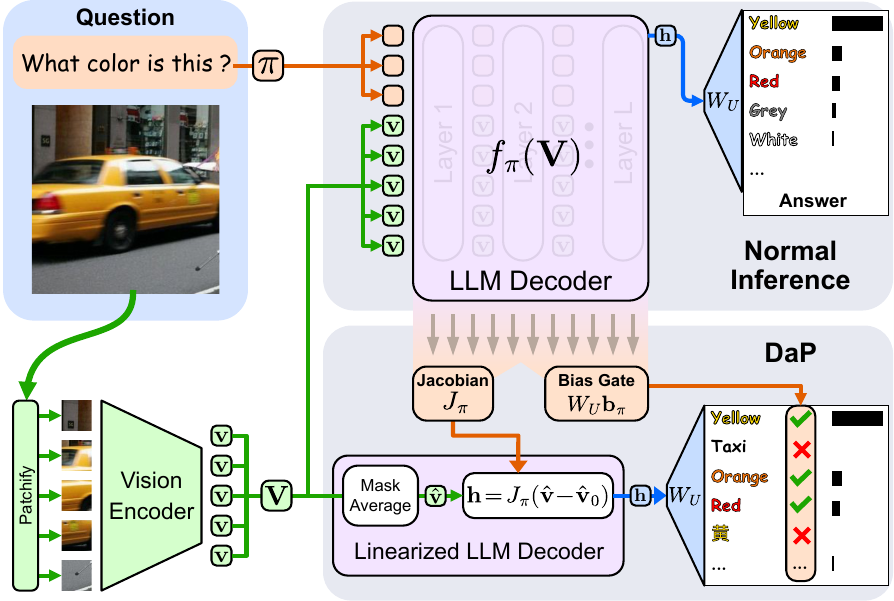}\hfill
    \caption{
        The top shows how, during a normal inference, an LVLM answers a question about an image.
        DaP approximates this process linearly with respect to visual input tokens, assuming the question is fixed.
        The resulting operator is a linear probe that answers the question.
        \Cref{fig:patch_readout} shows more question--answer examples.
    }
    \label{fig:method}
\end{figure}

\begin{figure*}[t]
    \centering
    \begin{minipage}[t]{0.19\textwidth}\centering
        \includegraphics[width=\linewidth, trim={187.6bp 83.9bp 80.9bp 55.8bp}, clip]{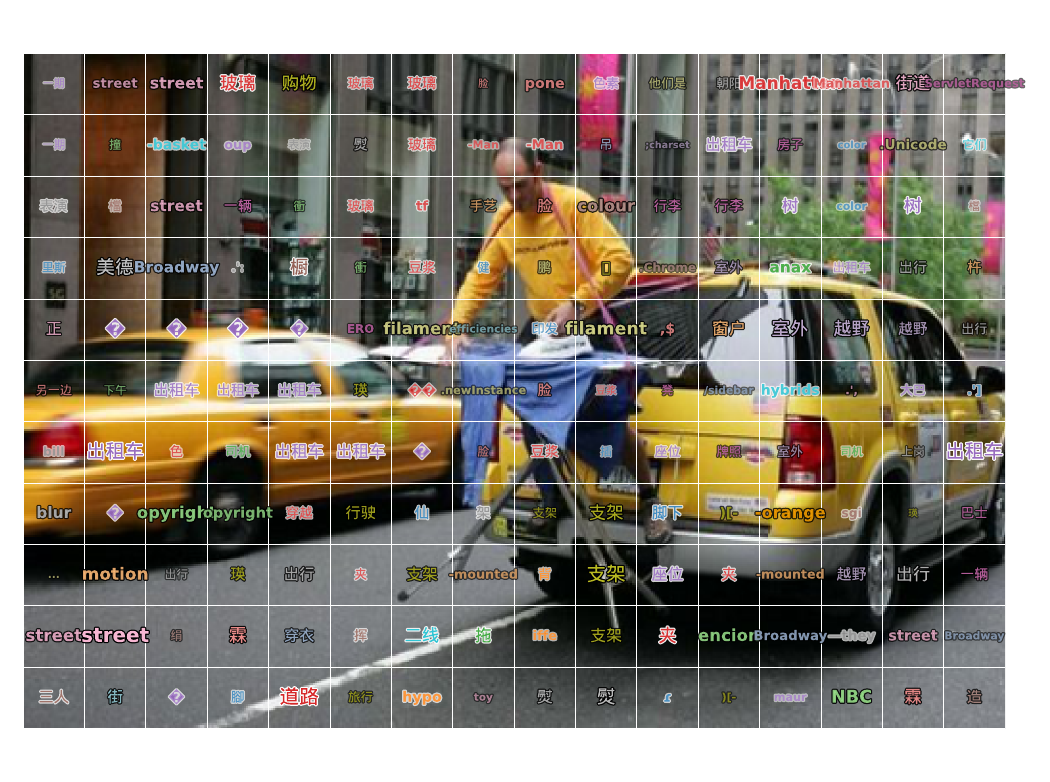}\\
        {\small Emb-Lens~\cite{fan2026visualtokens}}
    \end{minipage}\hfill
    \begin{minipage}[t]{0.19\textwidth}\centering
        \includegraphics[width=\linewidth, trim={187.5bp 86.5bp 86.5bp 59.0bp}, clip]{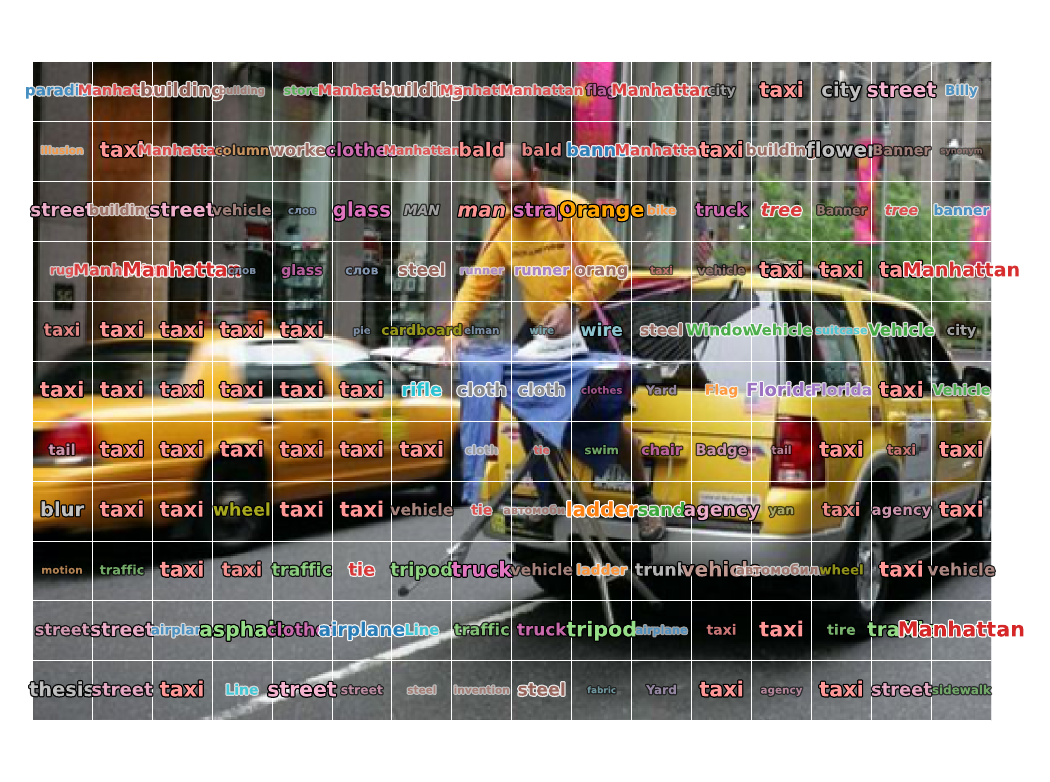}\\
        {\small DaP:}
        {\scriptsize\promptfont "Which word best names this object?"\par}
    \end{minipage}\hfill
    \begin{minipage}[t]{0.19\textwidth}\centering
        \includegraphics[width=\linewidth, trim={189.8bp 85.2bp 81.2bp 57.4bp}, clip]{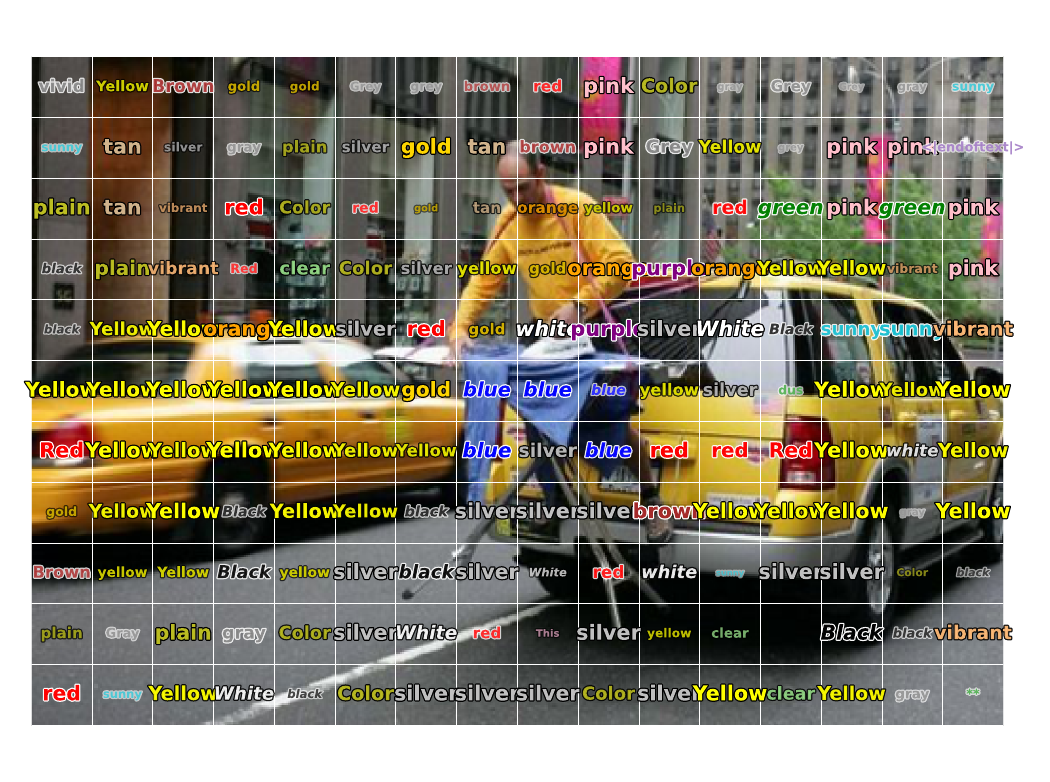}\\
        {\small DaP:}
        {\scriptsize\promptfont "Which word best describes the color of this object?"\par}
    \end{minipage}\hfill
    \begin{minipage}[t]{0.19\textwidth}\centering
        \includegraphics[width=\linewidth, trim={192.9bp 85.6bp 80.6bp 59.1bp}, clip]{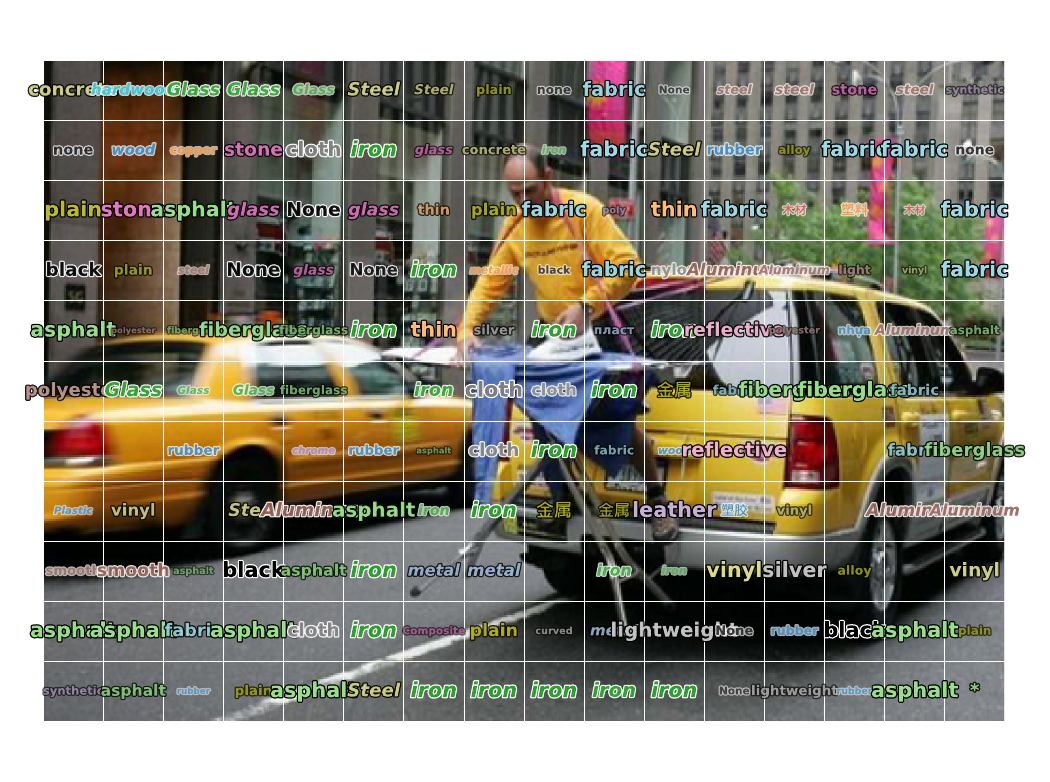}\\
        {\small DaP:}
        {\scriptsize\promptfont "Which word best describes the material of this object?"\par}
    \end{minipage}\hfill
    \begin{minipage}[t]{0.19\textwidth}\centering
        \includegraphics[width=\linewidth, trim={190.5bp 80.5bp 71.5bp 52.6bp}, clip]{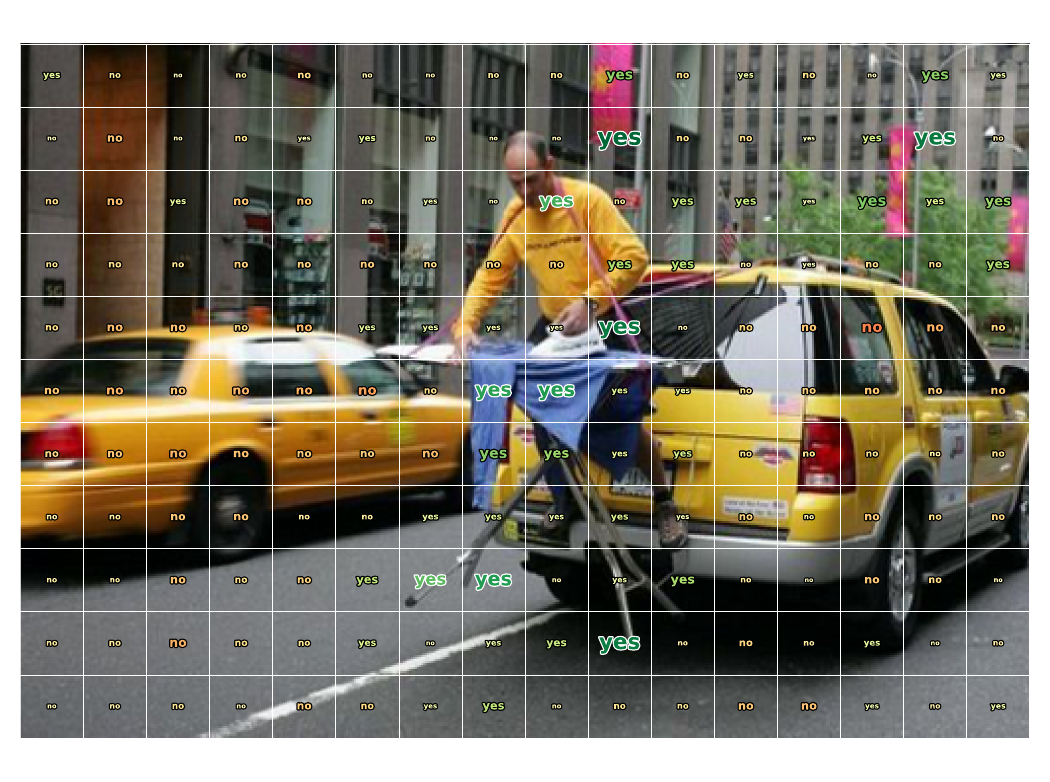}\\
        {\small DaP:}
        {\scriptsize\promptfont "Is this object hanging?"\par}
    \end{minipage}
    \caption{\textbf{Patch-level linear readouts.} 
    Top-1 token prediction from Emb-Lens~\cite{fan2026visualtokens} and from DaP estimated with different prompts.
    }
    \label{fig:patch_readout}
    \vspace{-8pt}
\end{figure*}

Yet, understanding how these representations encode high-level concepts is a key challenge in interpretability.
The dominant view, the 
Linear Representation Hypothesis~\cite{mikolov2013linguistic, park2024linear}, posits that high-level concepts are encoded as directions, 
so that an embedding decomposes as a weighted sum of concept directions (\eg, \concept{taxi} + \concept{yellow} + \concept{metallic}).
More recent work extends this view to take into account more elaborate structure of visual representations~\cite{trager2023linear, fel2026rabbit}.
They posit that concepts are grouped into multi-dimensional subspaces (\eg, \concept{taxi} among objects + \concept{yellow} among colors).
In any case, accessing those interpretable concepts requires by definition only a linear operator.
Yet, finding those operators that allow us to interpret and steer the representation is usually tedious. 

The standard tool to find such operators is the trained linear probe~\cite{savietto2026geometry, alain2016probes, kim2018tcav, li2023iti, rajaram2025line}. It fits a linear classifier on frozen representations to predict a concept, and then its weights are read as the concept direction.
However, linear probing suffers from three main limitations.
First, each probe requires a large labeled dataset, which must be collected for every concept of interest.
Second, and as a consequence, a probe only covers the classes it was trained on and does not generalize to unseen but connected concepts: a \emph{color} probe fitted on \concept{blue} says nothing about \concept{yellow}, although both could lie in the same color subspace.
Third, a probe only shows that a concept is linearly decodable, not that the LVLM actually uses this direction downstream~\cite{belinkov2022probing}.

In this work, we circumvent all three limitations with a simple insight: the LLM Decoder of an LVLM has already learned to read out visual representations in order to answer questions about the image.
For example, to answer \prompt{Is the taxi yellow?}, the decoder must retrieve the direction of \concept{yellow} from the visual tokens and turn it into the answer.
So why not use the decoder itself as the probe?
Inspired by LLM linearization methods~\cite{hernandez2024lre, chanin2024lrc, gurnee2026workspace},
we introduce \textbf{Decoder-as-a-Probe (DaP)},
which simply linearizes the LLM Decoder's next-token prediction with respect to the input visual tokens for a given question prompt (\eg, \prompt{Which word best describes the color of this object?}) (see \cref{fig:method}).
The resulting operator is a Jacobian matrix that maps representations of the vision encoder onto the model vocabulary, 
directly addressing the three limitations of probing.
(1) Only a few images are needed as anchor points for the linear approximation. 
It requires no labels on these images, 
 and uses the model's answer to the question as the reference. 
(2) It is not restricted to a fixed set of classes: since it maps visual representations onto the whole vocabulary, it can read out concept values never seen during estimation. 
(3) It is, by construction, an approximation of the computations the model performs itself to answer the question, making the resulting directions more efficient and precise at steering.

\textbf{In summary, our contribution is DaP}, a plug-and-play framework to estimate the direction of any concept in LVLM visual representation from a handful of images. These directions outperform previous baselines at steering with 256 times fewer training images, and extend to concepts never seen during estimation. 

\section{Related Work} 

\label{sec:related}

\paragraph{Linear Probing.}
%
When a large number of labels is available for a given concept,
to find its direction, the standard option is to fit a linear classifier
\cite{alain2016probes, kim2018tcav, li2023iti, rajaram2025line}.
A simpler option is the Nearest Centroid Classifier (NCC), whose directions are the class centroids~\cite{panickssery2024caa}. These directions steer better than those of trained probes~\cite{marks2024geometry}.
In LVLMs, both methods were used to study input visual tokens~\cite{savietto2026geometry}.

\paragraph{Lenses.}
%
Logit-Lens~\cite{nostalgebraist2020logitlens} can use the LLM unembedding matrix to read interpretable concepts in visual tokens of LVLMs~\cite{neo2025interpreting}.
These concepts are vocabulary text tokens that match the object the token covers (\eg, \tok{taxi}).
However, whereas last-layer visual tokens can be decoded, it works much less in the input layer, making it poorly suited to steering.
Recently, to better decode concepts at the input layer, Emb-Lens~\cite{jiang2025alignment, fan2026visualtokens, peng2024visualwords} uses the embedding matrix, since input visual tokens live in the same space as input text token embeddings.
The main problem of Lenses is that they mostly give the name of the object (see \cref{fig:patch_readout}). 
DaP instead can decompose a token into its attributes by asking a different question for each attribute.

\paragraph{SAEs.}
Sparse Autoencoders (SAEs) have gained significant traction in recent years~\cite{cunningham2024sae, fel2025archetypal, pach2025sae}.
By training on a large corpus of unlabeled data, SAEs decompose superposed concepts into a large dictionary of directions.
However, they are often unstable to train~\cite{paulo2025sae, fel2025archetypal} and 
naming these directions with interpretable concepts takes extra machinery~\cite{paulo2025automatically,zhang2025lmm,rao2024discover}.
Moreover, these directions offer limited steerability~\cite{bhalla2024unifying, menon2025analyzing}.
DaP, in contrast, is plug-and-play, as it needs only a few images, targets the concept chosen by the question and gives steerable directions.
Still, SAEs give a full sparse decomposition of the representation.

\paragraph{Jacobian in LLMs.} 
The Jacobian gives a local linear approximation of a network computation, and is used by the language community to find computations performed by LLMs that can be approximated by a linear operator.
LRE~\cite{hernandez2024lre} approximates how an LLM maps a fact about a subject.
Later, LRC~\cite{chanin2024lrc} inverted this linear map to get concept directions (\eg, \concept{located in France} for cities such as \concept{Paris}).
Our work builds on this approximation, which we adapt and extend to visual input representations in four ways.
(1) We propose a framework in which the prompt is an open or closed question, so that any concept can be probed.
(2) We extend the linearization to objects that span several image tokens.
(3) We use the unembedding matrix as a dictionary of concept direction anchors to reduce the reliance on labels, generalize to unseen concepts and remove the costly matrix inversions.
(4) We propose using the bias as a gate on the vocabulary.
%
Recently, J-Lens~\cite{gurnee2026workspace} has used the Jacobian to map hidden states from middle LLM layers to all future last-layer hidden states. 
By verbalizing these intermediate hidden states with the unembedding matrix, it reveals the reasoning the model carries out along its layers.

\section{Motivation}
\label{sec:prelim}

\paragraph{Linear Representation Structure.}
A concept is defined as any interpretable and high-level property, like \concept{wearing a hat}, \concept{being a taxi} or \concept{being yellow}.
%
The Linear Representation Hypothesis states that concepts are represented linearly in the representation space of neural networks~\cite{park2024linear}.
The power of this hypothesis lies in the fact that we can therefore use linear algebra to steer and interpret concepts inside the model. 
This is why finding concept directions is essential to interpret model activations.
It has been studied and largely confirmed in LLMs~\cite{marks2024geometry}.
Moreover, prior work~\cite{park2025hierarchi} shows that concept directions are not arranged arbitrarily in the representation space but follow a \textbf{hierarchical} structure (\eg, \concept{mammal} decomposes as \concept{animal} plus an orthogonal \concept{mammal} part).
 
Still, for representations derived from images, we can expect even more structure than for text.
Visual inputs are \textbf{continuous}, meaning two banana pictures are never exactly the same shade of yellow and the same length. By contrast, the text token embedding of \tok{yellow} is discrete and fixed around a finite vocabulary.
This was first formalized in cognitive science as \textit{conceptual spaces}~\cite{gardenfors2000conceptual}. 
Continuity should have implications for the induced representations.
This is why \citet{fel2026rabbit}
recently proposed a refined view in which image representations are composed of a sum of convex mixtures
(\eg, \concept{taxi} among objects, \concept{yellow} among colors, and \concept{metallic} among materials).
%
Following this view, we consider in this paper that any attribute type (\eg, colors or materials) can be read with a single linear projection.

\paragraph{Superposition.}
A related but distinct idea is the superposition hypothesis~\cite{elhage2022superposition}.
It states that the number of dimensions of a hidden state is much smaller than its total concept representational capacity.
This is possible if we assume that
(1) concepts are encoded as quasi-orthogonal directions forming an overcomplete frame~\cite{strohmer2003grassmannian};
and (2) only a sparse set of concepts is activated at a given time.
%
The unembedding matrix $W_U$ of an LLM gives a direct example of superposition as
it up-projects the hidden dimension of the LLM to a token vocabulary that is usually many times bigger (\eg, $60\times$ for Qwen3-VL-4B~\cite{qwen2025qwen3vl}, from 2{,}560 to 151{,}936).
In this paper, we leverage the fact that 
most concepts can be represented by a token (\concept{yellow} $\to$ \tok{yellow}, \concept{metallic} $\to$ \tok{metallic}).
Thus, we could use the unembedding matrix as a free, labeled dictionary of concept directions. However, directly applying $W_U$ to input visual embeddings 
gives poor results~\cite{neo2025interpreting} 
and does not allow selecting a given attribute subspace.
This is why, in the following section, we introduce DaP, which overcomes these limitations by finding, given a prompt, the linear operator $J_\pi$ that maps visual embeddings to an aligned space for $W_U$ readout.
%

\section{The Decoder-as-a-Probe Framework} 
\label{sec:method}
%
First, we show that an LLM Decoder can be interpreted as a classifier, then that an LLM Decoder can be linearized.
Lastly, we show how to combine both to get a linear probe programmable by a prompt.


For the purpose of this paper, we consider an LVLM to consist of three parts (see \cref{fig:method}): a vision encoder $\phi_v(\cdot)$\footnote{Here, we make the projector part of the vision encoder, but most of our work can be applied to visual representations before the projector, especially if it is a linear projection.}, an LLM Decoder $\mathrm{LLM}(\cdot)$ and an unembedding matrix $W_U$.
Given an image $x$, the vision encoder produces a sequence of $N$ visual tokens $\mathbf{V} = \phi_v(x) = [\mathbf{v}_1, \dots, \mathbf{v}_N] \in \mathbb{R}^{N \times D}$.
In order to predict the next text token, the LLM Decoder takes the sequence of visual tokens concatenated with the prompt $\pi$, and returns $\mathbf{h} = \mathrm{LLM}([\mathbf{V}; \pi]) \in \mathbb{R}^{D}$,
where $\mathbf{h}$ is the output of the last layer at the last token position.
Finally, the unembedding matrix has one row $\mathbf{w}_t \in \mathbb{R}^{D}$ per token $t$ of the vocabulary $\mathcal{V}$.
The logit of token $t$ is $z_t = \mathbf{w}_t^\top \mathbf{h}$, and the probability that $t$ is the next token is
\begin{equation}
  p_\pi(t \mid \mathbf{V}) = \frac{e^{z_t}}{\sum_{t' \in \mathcal{V}} e^{z_{t'}}}, \quad t \in \mathcal{V}.
\end{equation}

This next-token prediction naturally lets us interpret the LLM Decoder as a classifier.

\subsection{Decoder as a Classifier}
\label{sec:classifier}

\paragraph{Closed Classifier.}
The prompt $\pi$ can be formulated so that the next-token prediction of the LLM Decoder is the answer to a question about the image, with $\mathcal{A} \subseteq \mathcal{V}$ the set of possible answers.
For example, \prompt{Is the man wearing a hat? Answer yes or no.} with $\mathcal{A} = \{\mbox{\prompt{yes}}, \mbox{\prompt{no}}\}$.
Then the next-token prediction can be turned into a classifier by
restricting the softmax to the answer tokens $\mathcal{A}$ giving: 
\begin{equation}
  p_\pi(t \mid \mathbf{V}) = \frac{e^{z_t}}{\sum_{t' \in \mathcal{A}} e^{z_{t'}}}, \quad t \in \mathcal{A}
  .
\end{equation}

\paragraph{Open Classifier.}
We can also ask an open question to the model, with $\mathcal{A} = \mathcal{V}$,
\eg, \prompt{Which word best describes the color of this object?}.
This holds for any concept category (\eg, object name, shape, material, activity).
If a set of colors to probe is known in advance, the answer space can be restricted to those colors. However, for interpretability it is important to be able to discover representations present in the model that could not have been anticipated; this is why keeping the answer set open is valuable.

In both the open and closed cases, the classifier is the LLM Decoder itself, which is not linear in the visual tokens. Turning it into a linear probe requires linearizing it with respect to the visual tokens. 
Unlike in text~\cite{hernandez2024lre, gurnee2026workspace}, a concept in an image is spread over many visual tokens, which makes this linearization not straightforward.

\subsection{Decoder as a Linear Function}
\label{sec:linear}

Our objective is to estimate, for a given prompt, a single linear operator $J_\pi$ that maps the visual tokens $\mathbf{V}$ to $\mathbf{h}$, which is read by the unembedding matrix $W_U$.
This operator should be shared across all visual tokens and images.

If the prompt is a question about the image, it is fixed, so we rewrite $f_\pi(\mathbf{V}) = \mathrm{LLM}([\mathbf{V}; \pi])$
%
and compute the first-order Taylor expansion of $f_\pi(\cdot)$ around an image $\mathbf{V}$: 
\begin{equation}
  f_\pi(\mathbf{V}') = f_\pi(\mathbf{V}) + \sum_{i=1}^{N} J_{\pi,i,\mathbf{V}} \, (\mathbf{v}'_i - \mathbf{v}_i) + o(\lVert \mathbf{V}' - \mathbf{V} \rVert),
\end{equation}
where $J_{\pi,i,\mathbf{V}} \in \mathbb{R}^{D \times D}$ is the partial Jacobian of $f_\pi(\cdot)$ with respect to its $i$-th visual token.
As in a total derivative, each visual token contributes its own term $J_{\pi,i,\mathbf{V}} (\mathbf{v}'_i - \mathbf{v}_i)$ to the change of $\mathbf{h}$.

\paragraph{{Making $J_\pi$ independent of the token}.} 
Using two simplifications, we reduce all partial Jacobians $J_{\pi,i,\mathbf{V}}$ to a single linear operator $J_{\pi,\mathbf{V}}$ shared across visual tokens.
First, we make sure that the answer to the prompt is not affected by the image background.
Let $\mathcal{M}$ be the set of visual tokens covered by the object; then $J_{\pi,i,\mathbf{V}} \, (\mathbf{v}'_i - \mathbf{v}_i) \approx \mathbf{0}$ for $i \notin \mathcal{M}$.
%
Now that all tokens are part of the object, 
none carries information contradicting the answer (\eg, another color). 
{Second, since the answer should not depend on the object position and size}, we assume the decoder applies an equal share of the same map to every token, $J_{\pi,i,\mathbf{V}} = \frac{1}{|\mathcal{M}|} J_{\pi,\mathbf{V}}$ for $i \in \mathcal{M}$, giving
\begin{align}
f_\pi(\mathbf{V}') \approx f_\pi(\mathbf{V}) + J_{\pi,\mathbf{V}} \, (\hat{\mathbf{v}}' - \hat{\mathbf{v}}), \quad \text{with} \\
  \hat{\mathbf{v}} = \frac{1}{|\mathcal{M}|} \sum_{i \in \mathcal{M}} \mathbf{v}_i \in \mathbb{R}^{D},
  \quad
  J_{\pi,\mathbf{V}} = \frac{\partial f_\pi(\mathbf{V})}{\partial \hat{\mathbf{v}}}.
\end{align}


\paragraph{{Making $J_\pi$ independent of the image}.}
If the LLM Decoder were linear, using only one image estimate would perfectly identify the linear function of the decoder.
Since the LLM Decoder is nonlinear, even with a fixed prompt, estimating $J_\pi$ from a single sample is noisy. 
Averaging over more samples (when available) improves performance by reducing the dependence on the images used for estimation:
\begin{align}
      f_\pi(\mathbf{V}') \approx \mathbf{b}_\pi + J_\pi \left( \hat{\mathbf{v}}' - \hat{\mathbf{v}}_0 \right),\quad  \text{with}
    \label{eq:final} \\ 
        \mathbf{b}_\pi = \mathbb{E}_{\mathbf{V}} [ f_\pi(\mathbf{V}) ],
    \;\;
    J_\pi = \mathbb{E}_{\mathbf{V}} [ J_{\pi,\mathbf{V}} ],
    \;\;
    \hat{\mathbf{v}}_0 = \mathbb{E}_{\mathbf{V}} [ \hat{\mathbf{v}}].
\end{align}
%


\paragraph{Using the Penultimate Layer.}
%
%
Training neural networks with cross-entropy loss makes the last-layer representations of same label inputs nearly identical, a phenomenon called neural collapse~\cite{papyan2020collapse}.
In LLMs, this collapse makes inputs that lead to the same next token end up with almost identical last-layer hidden states~\cite{wu2024collapse,li2025hidden}. 
This nonlinearity behaves as a smooth approximation of a step function and leads to noisy Jacobian estimates~\cite{gurnee2026workspace}.
Building on this observation, in open classification, we estimate $\mathbf{h}$ at the penultimate layer, which improves performance.

\subsection{Decoder as a Probe (DaP)}
\begin{table*}
\centering
\small\renewcommand{\arraystretch}{1}\setlength{\tabcolsep}{5pt}
\newsavebox{\acctable}
\sbox{\acctable}{%
\begin{tabular}{llc|c|cccc|c}
\toprule
 & Method & Shots & All Attributes & Color & Material & Shape & Activity & Names \\
\midrule
 & NCC~\cite{panickssery2024caa} & 16 & 68.0 & 49.4 & 59.9 & 39.9 & 70.3 & \underline{84.0} \\
 & Probe~\cite{savietto2026geometry} & 16 & \underline{69.6} & 55.0 & 63.6 & 51.9 & 80.3 & \textbf{86.8} \\
 & Emb-Lens~\cite{fan2026visualtokens} & 0 & -- & 51.7 & 64.5 & 41.3 & 53.3 & 61.4 \\
 & \cellcolor{gray!15}DaP & \cellcolor{gray!15}1 & \cellcolor{gray!15}63.6 & \cellcolor{gray!15}\underline{69.3} & \cellcolor{gray!15}\underline{73.5} & \cellcolor{gray!15}\underline{63.3} & \cellcolor{gray!15}\underline{82.0} & \cellcolor{gray!15}77.6 \\
 & \cellcolor{gray!15}DaP & \cellcolor{gray!15}16 & \cellcolor{gray!15}\textbf{73.7} & \cellcolor{gray!15}\textbf{72.5} & \cellcolor{gray!15}\textbf{78.9} & \cellcolor{gray!15}\textbf{65.5} & \cellcolor{gray!15}\textbf{84.9} & \cellcolor{gray!15}78.7 \\
\cmidrule(l){2-9}
\multirow{-6}{*}{\rotatebox[origin=c]{90}{\shortstack{\textit{Seen (4-way)}}}} & \textit{Model} & \textit{0} & \textit{83.0} & \textit{90.0} & \textit{89.5} & \textit{74.2} & \textit{90.2} & \textit{82.3} \\
\midrule\midrule
 & \# concepts &  & -- & 27 & 10 & 17 & 21 & 1313 \\
\cmidrule(l){2-9}
& Emb-Lens~\cite{fan2026visualtokens} & 0 & -- & 25.5 & 39.4 & 24.3 & \underline{24.5} & 10.6 \\
 & \cellcolor{gray!15}DaP & \cellcolor{gray!15}1 & \cellcolor{gray!15}-- & \cellcolor{gray!15}\underline{44.3} & \cellcolor{gray!15}\underline{59.0} & \cellcolor{gray!15}\underline{52.9} & \cellcolor{gray!15}23.6 & \cellcolor{gray!15}\underline{20.0} \\
 & \cellcolor{gray!15}DaP & \cellcolor{gray!15}16 & \cellcolor{gray!15}-- & \cellcolor{gray!15}\textbf{51.2} & \cellcolor{gray!15}\textbf{65.8} & \cellcolor{gray!15}\textbf{53.9} & \cellcolor{gray!15}\textbf{42.3} & \cellcolor{gray!15}\textbf{24.3} \\
\cmidrule(l){2-9}
\multirow{-5}{*}{\rotatebox[origin=c]{90}{\shortstack{\textit{Unseen}}}} & \textit{Model} & \textit{0} & \textit{--} & \textit{72.1} & \textit{77.4} & \textit{63.4} & \textit{64.6} & \textit{28.5} \\
\bottomrule
\end{tabular}}%
\usebox{\acctable}\hfill
\begin{minipage}[c]{\dimexpr\linewidth-\wd\acctable-1em\relax}
\centering
\includegraphics[width=\linewidth, height=\dimexpr\ht\acctable+\dp\acctable\relax, keepaspectratio]{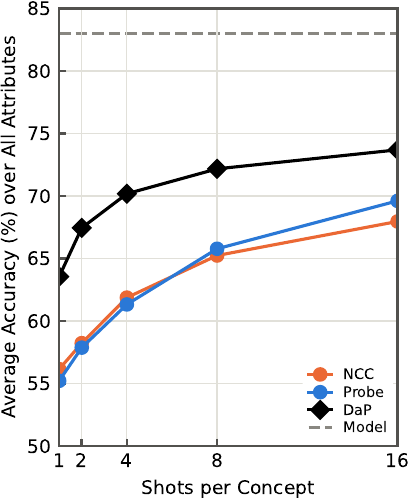}
\end{minipage}
\caption{ 
\textbf{Linear probe accuracy (\%)}. How often the probe detects the correct concept.
All Attributes are the 570 attributes binary classification accuracy averaged.
Seen concepts accuracy is on the 4 concepts seen during estimation. Unseen concepts are never seen during estimation.
\textbf{Bold}: best, \underline{underlined}: second best.
Right shows accuracy over All Attributes against the number of shots.
}
\vspace{-5pt}
\label{tab:benchmark}
\end{table*}

Here we describe how to build DaP.

\paragraph{Estimation.}
First, we choose a prompt $\pi$ that asks about the concept to probe, either a closed yes-no question for a single concept or an open question for a whole attribute type (\cref{sec:classifier}).
Second, we estimate $J_\pi$, $\mathbf{b}_\pi$ and $\hat{\mathbf{v}}_0$ by averaging over a set of images with object masks (\cref{sec:linear}).
These images can be drawn at random, which requires no label, or chosen as positive and negative instances of the concept to probe.
Now, replacing the LLM Decoder with its linear approximation of \cref{eq:final} gives a linear probe whose logit of each answer $t \in \mathcal{A}$ is
\begin{equation}
  z_t = \mathbf{w}_t^\top \mathbf{b}_\pi + \mathbf{w}_t^\top J_\pi (\hat{\mathbf{v}} - \hat{\mathbf{v}}_0),
\end{equation}
where $\mathbf{w}_t$ is the $t$-th row of the unembedding matrix $W_U$.
%
In its current form, the linear approximation is not faithful to the model answers.
If we look at the current probe prediction, the bias acts as a prior probability over the class associated with the token $t$
\begin{equation}
  p_{\pi}(t \mid \mathbf{V}) \propto \underbrace{e^{\mathbf{w}_t^\top J_\pi (\hat{\mathbf{v}} - \hat{\mathbf{v}}_0)}}_{\text{likelihood}} \; \underbrace{e^{\mathbf{w}_t^\top \mathbf{b}_\pi}}_{\text{prior}}.
\end{equation}
However, the likelihood term has a low impact on the answer and is not powerful enough to counterbalance the prior. 
This prior has two negative effects:
First, it biases answers toward classes seen during the estimation (\eg, $p_{\text{color}}(\text{blue}) \sim 33 \, p_{\text{color}}(\text{red})$).
Second, it biases answers toward words that are more common than others
(\eg, $p_{\text{color}}(\text{gray}) \sim 49 \, p_{\text{color}}(\text{silver})$).
%
%
For these reasons, we set the bias to $\mathbf{b}_\pi = \mathbf{0}$, giving
\begin{equation}
  z_t = \mathbf{w}_t^\top J_\pi (\hat{\mathbf{v}} - \hat{\mathbf{v}}_0)
  .
  \label{eq:dap}
\end{equation}

\paragraph{Probing (Closed Classifier).}
When the prompt asks a closed question, DaP predicts its answer among the possible answers ($\mathcal{A} \subseteq \mathcal{V}$).
Given a region $\mathcal{M}$ to probe, we average its visual tokens into $\hat{\mathbf{v}}$, which DaP classifies by the highest logit $z_t$ over $\mathcal{A}$ (\cref{eq:dap}).
The mask can be a single token, giving a single patch readout.
In this setup, we only need one direction per answer $t$, $J_\pi^\top \mathbf{w}_t$, instead of the full operator $J_\pi$. Since each direction estimation takes a backward pass, this reduces the estimation cost from $D$ backward passes to $|\mathcal{A}|$.
Note that for a yes-no question, the answer direction \prompt{no} then plays the role of the counterfactual, which is known to help isolate the concept~\cite{park2024linear}.

\paragraph{Verbalizing (Open Classifier).}
When the prompt asks an open question, DaP verbalizes its answer over the whole vocabulary ($\mathcal{A} = \mathcal{V}$).
We noticed that when the bias is removed,
words related to the object but not part of the answer space appear (\eg, \concept{taxi} instead of \concept{yellow})
and some answers are correct but in other languages, such as Chinese.
As the LVLM does not produce those artifacts, we know they are due to the linearization. 
Hence, 
\textbf{the bias plays the role of the attribute subspace selector},
giving a low prior to tokens that are not part of the answer space.
To address this issue, we use the bias as a gate in open classification.
More precisely, we restrict the answer space $\mathcal{A}$ to the top-$k$ tokens with the highest bias logits $W_U \mathbf{b}_\pi$ (see \cref{fig:method}).
This gate multiplies each answer probability by a binary mask, equal to one for these $k$ tokens and zero for the others.
We use $k = 200$ for attribute types and $k = 5000$ for naming objects, which have more possible answers.


\paragraph{Steering.}
To steer the model perception of the image, 
we directly modify visual embeddings 
in the selected mask region $\mathcal{M}$ of choice.
Let $\mathbf{w}_{\text{source}}$, $\mathbf{w}_{\text{target}}$ be the source and target concepts to steer and $\delta = J_\pi^\top (\mathbf{w}_{\text{target}} - \mathbf{w}_{\text{source}})$ be the steering direction, then we apply
%
\begin{equation}
  \mathbf{v}_i \leftarrow \mathbf{v}_i + \alpha \, \lVert \hat{\mathbf{v}} \rVert \, \frac{\delta}{\lVert \delta \rVert}, \quad i \in \mathcal{M}
  ,
  \label{eq:steer}
\end{equation}
where $\alpha\in \mathbb{R}^+$ is the chosen steering strength.
%
If the linear approximation is accurate, steering increases the difference between the target and source answer logits by $\alpha \lVert \hat{\mathbf{v}} \rVert \lVert \delta \rVert$, which is the maximum for an edit of this norm.
This explains why DaP has a strong steering efficiency.


\section{Quantitative Experiments} 
\begin{table*}
\centering
\small\renewcommand{\arraystretch}{1}\setlength{\tabcolsep}{5pt}
\newsavebox{\steertable}
\sbox{\steertable}{%
\begin{tabular}{llc|c|cccc|c}
\toprule
 & Method & Shots & All Attributes & Color & Material & Shape & Activity & Names \\
\midrule
 & NCC & 16 & 39.8 & 19.8 & 27.0 & 21.8 & 14.9 & 21.3 \\
 & NCC & 256 & -- & 39.8 & 42.3 & 35.0 & 31.7 & \underline{28.0} \\
 & Probe & 16 & 38.6 & 26.0 & 22.5 & 23.3 & 15.4 & 9.7 \\
 & Probe & 256 & -- & 57.0 & 50.7 & 42.8 & 44.2 & 5.0 \\
 & Emb-Lens & 0 & -- & 6.8 & 15.0 & 15.3 & 4.3 & 4.7 \\
 & \cellcolor{gray!15}DaP & \cellcolor{gray!15}1 & \cellcolor{gray!15}\underline{74.4} & \cellcolor{gray!15}\underline{75.2} & \cellcolor{gray!15}\underline{76.5} & \cellcolor{gray!15}\underline{82.0} & \cellcolor{gray!15}\underline{57.2} & \cellcolor{gray!15}27.5 \\
\multirow{-7}{*}{\rotatebox[origin=c]{90}{\shortstack{\textit{Seen}}}} & \cellcolor{gray!15}DaP & \cellcolor{gray!15}16 & \cellcolor{gray!15}\textbf{89.5} & \cellcolor{gray!15}\textbf{83.5} & \cellcolor{gray!15}\textbf{82.2} & \cellcolor{gray!15}\textbf{86.3} & \cellcolor{gray!15}\textbf{80.3} & \cellcolor{gray!15}\textbf{43.8} \\
\midrule\midrule
 & \# targets &  & -- & 27 & 10 & 17 & 21 & 1453 \\
 \cmidrule(l){2-9}
 & Emb-Lens & 0 & -- & 0.7 & 1.5 & 0.7 & 0.0 & 0.5 \\
 & \cellcolor{gray!15}DaP & \cellcolor{gray!15}1 & \cellcolor{gray!15}-- & \cellcolor{gray!15}\underline{23.0} & \cellcolor{gray!15}\underline{33.3} & \cellcolor{gray!15}\underline{21.5} & \cellcolor{gray!15}\underline{17.8} & \cellcolor{gray!15}\underline{14.5} \\
\multirow{-4}{*}{\rotatebox[origin=c]{90}{\shortstack{\textit{Unseen}}}} & \cellcolor{gray!15}DaP & \cellcolor{gray!15}16 & \cellcolor{gray!15}-- & \cellcolor{gray!15}\textbf{25.2} & \cellcolor{gray!15}\textbf{42.3} & \cellcolor{gray!15}\textbf{42.0} & \cellcolor{gray!15}\textbf{35.1} & \cellcolor{gray!15}\textbf{20.7} \\
\bottomrule
\end{tabular}}%
\usebox{\steertable}\hfill
\begin{minipage}[c]{\dimexpr\linewidth-\wd\steertable-1em\relax}
\centering
\includegraphics[width=\linewidth, height=\dimexpr\ht\steertable+\dp\steertable\relax, keepaspectratio]{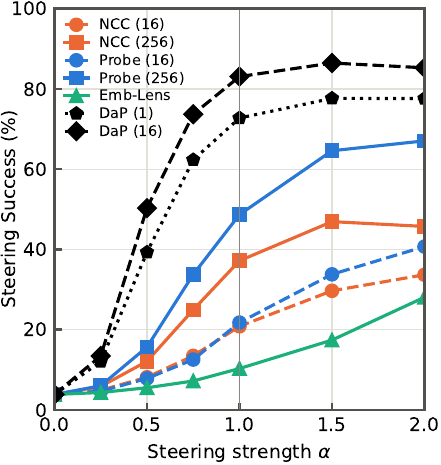}
\end{minipage}
\caption{
\textbf{Steering success (\%) at $\alpha=1$}.
How often steering makes the model answer the target concept.
For All Attributes, steering aims to flip the yes-no answer.
Seen and Unseen rows steer toward seen and unseen concepts, respectively.
\textbf{Bold}: best, \underline{underlined}: second best.
Right shows the Steering success as a function of steering strength. 
}
\vspace{-5pt}
\label{tab:steering}
\end{table*}

\paragraph{Dataset.}
We use the Visual Attributes in the Wild (VAW) dataset~\cite{pham2021vaw}.
It consists of 261k object instances annotated with an object name, a bounding box and a list of attributes, using images from Visual Genome~\cite{krishna2017visual}.
VAW contains 620 different attributes%
, which are hierarchically grouped into attribute types, the four largest being Colors, Materials, Shapes and Activities.
We crop images around the object bounding box with a 20\% margin.

\paragraph{Benchmark.}
\textbf{Model:} We use Qwen3-VL-4B. Results with other models are shown in the supplementary material~\ref{sec:app:models}.
%
\textbf{Tasks:} We evaluate three concept probing tasks, defined by the following prompts:
(1) {Object Names.} \prompt{Which word best names the thing in the middle of the image?};
(2) {Attribute Types.} \prompt{Which word best describes the \ph{attribute type} of the \ph{object name}?}, with \textit{attribute type} being color, material, shape and activity;
(3) {All Attributes.} 
Most attributes are not part of an attribute type or part of a small one, so we probe all attributes independently using a binary classification task
with the prompt \prompt{Is the \ph{object name} \ph{attribute}? Answer yes or no.}
%
%
\textbf{Baselines:}
%
We compare against a trained linear probe~\cite{savietto2026geometry}, NCC~\cite{panickssery2024caa} and Emb-Lens~\cite{fan2026visualtokens}.
We evaluate in a few-shot setup, where regularization plays an important role in performance~\cite{zhai2022scaling}. For the trained linear probe, we thus use closed-form ridge regression~\cite{hoerl1970ridge, balestriero2025lejepa} with the regularization hyperparameter that gives the best validation accuracy for each task and shot count.
%
%
%
%

\paragraph{Probe Estimation.}
First, for Object Names and Attribute Types, the probe is fitted on the four concepts with the most training instances. We refer to them as the seen concepts. 
We fit each method on 16 images (shots) per seen concept. 
Only the trained probe and NCC need labels at estimation.
Only DaP, Emb-Lens and the model can be evaluated on unseen concepts.
For each task, we estimate $J_\pi$ with the prompt $\pi$ of that task.

While VAW contains 927k concept annotations, it is highly unbalanced, as the median concept has only 94 training instances.
This is why we evaluate DaP at 16 shots, which we consider a realistic setting.
For All Attributes, VAW also provides explicit negative labels, so we sample 16 positive and 16 negative training instances per concept.
This lets us probe the 570 of the 620 VAW attributes that have at least 16 training and test samples.
There are no unseen concepts, and Emb-Lens cannot be evaluated on All Attributes.
Full experimental details are given in the supplementary material (\cref{sec:app:setup}).
%
%
%

\paragraph{Probing Benchmark.}
We evaluate the capacity of each linear probe to detect the presence of a concept. 
\Cref{tab:benchmark} shows the closed accuracy on seen and unseen concepts.
For a fair comparison, we show closed accuracy for DaP, Emb-Lens and the LVLM for reference. Otherwise, since Emb-Lens is unable to select attributes, it would score near 0.
The trained probe and NCC, with few labeled instances and no prompt to specify what to look for, rely on the most salient information in the visual embedding, which is the object identity. As a result, \uline{DaP outperforms all baselines on every attribute type, even at 1 shot.}

The aim of DaP is not to be the best detector of a concept, but to be faithful to the model's perception, \ie, to give the answer the model gives for each object, making DaP bounded by the model accuracy. For example, on Object Names, DaP (78.7) is less accurate than the LVLM itself (82.3), which is less accurate than the trained probe (86.8)~\cite{savietto2026geometry}.
We provide faithfulness results, which favor DaP by design, in the supplementary material~\ref{sec:app:faithfulness}.
Yet, on the 1,313 object names never seen during estimation, DaP reaches a top-1 accuracy of 24.3\%, close to that of the model (28.5\%), and more than twice that of Emb-Lens (10.6\%).
More generally, \uline{DaP at 16 shots largely outperforms Emb-Lens.}

\paragraph{Steering Benchmark.}
We evaluate the capacity of each linear probe to steer the answer of the model toward a target concept.
Steering is successful when the answer of the steered model is the target concept.
To avoid favoring DaP, we steer the model with a task question different from the one used to estimate DaP, and we show in the supplementary material~\ref{sec:app:steering_language} the results when steering with a question in another language.
As in \cref{eq:steer}, each probe steers the visual tokens of the object along the difference between its target and source concept directions ($\delta$), for a given steering strength $\alpha$.
\Cref{tab:steering} (left) shows the steering success rate for $\alpha = 1$, and \cref{tab:steering} (right) shows the steering success rate for Attribute Types as a function of $\alpha$.
\uline{For Attribute Types, DaP 1-shot needs ${\sim}2\times$ less steering strength than the linear probe trained on 256 shots.} 
Interestingly, Emb-Lens can probe concepts but is remarkably poor at steering, which may explain why previous work only uses it for readout~\cite{jiang2025alignment, fan2026visualtokens, peng2024visualwords}.
\uline{As a result, among all methods, only DaP can steer toward unseen concepts.}
Also, steering the object name is harder, as the object identity takes up a large part of the visual embedding.

We additionally measure the steering specificity of each probe. Specificity measures how much the steering leaves the LVLM perception of unrelated concepts unchanged. 
For this, we ask the model questions about unrelated concepts before and after steering.
For a fair comparison, each method steers at the $\alpha$ that gives a steering success rate of 40\% (\cref{tab:surgical}).
\uline{For All Attributes, DaP changes the object name $4\times$ less often than the trained probe and NCC.}

\section{Qualitative Experiments}
\begin{figure*}[t]
    \centering
    \includegraphics[width=\linewidth]{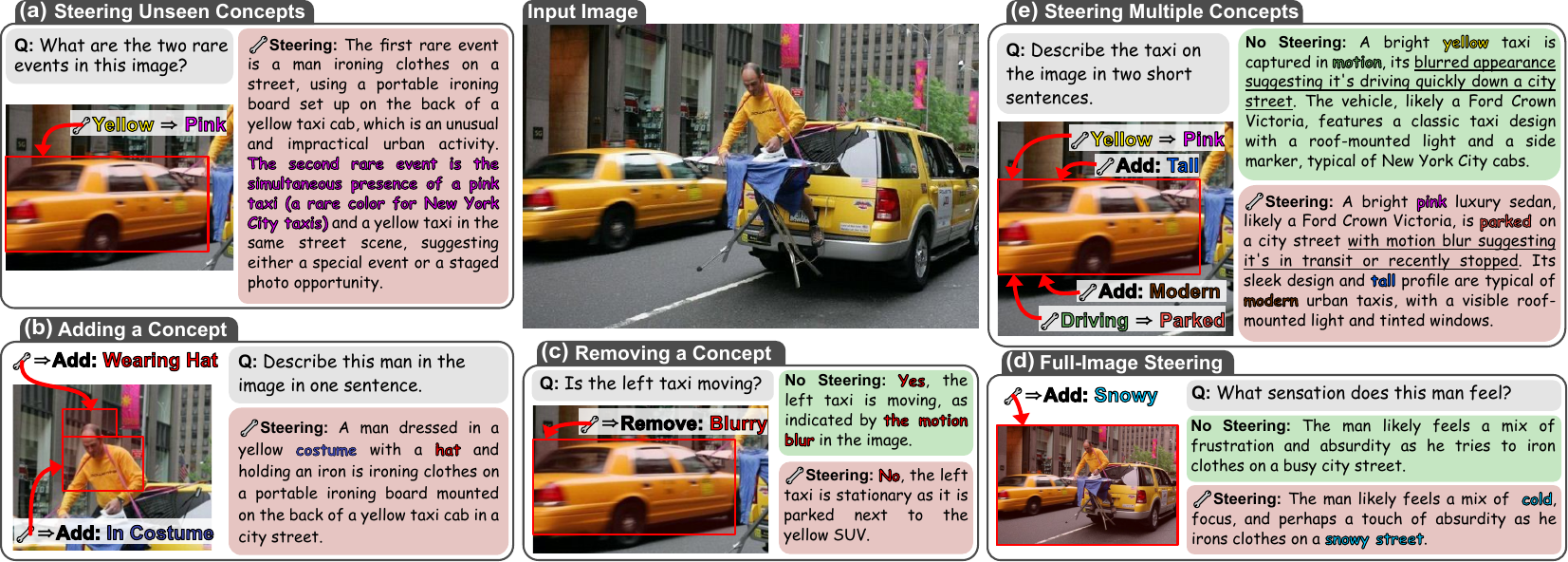}
    \caption{\textbf{Examples of steering visual input embeddings with DaP.}
    \nosteerhl{Green} marks the answer of the model without steering, and \steerhl{red} the answer of the steered model.
    The red box highlights the visual patches where the steering is performed.
    }
    \label{fig:steer}
    \vspace{-5pt}
\end{figure*}

\begin{table}
\centering
\small\renewcommand{\arraystretch}{1}\setlength{\tabcolsep}{4pt}
\begin{tabular}{l|cc|cc}
\toprule
 & \multicolumn{2}{c}{All Attributes} & \multicolumn{2}{c}{Attribute Types} \\
\cmidrule(lr){2-3} \cmidrule(lr){4-5}
Method & \shortstack{Attribute \\ flip} & \shortstack{Name \\ change} & \shortstack{Attribute \\ flip} & \shortstack{Name \\ change} \\
\midrule
NCC & 6.3 & 19.7 & 9.5 & 23.1 \\
Probe & 6.2 & 16.8 & 5.9 & 10.9 \\
\cellcolor{gray!15}DaP (1-shot) & \cellcolor{gray!15}\underline{3.2} & \cellcolor{gray!15}\underline{6.0} & \cellcolor{gray!15}\underline{4.7} & \cellcolor{gray!15}\underline{10.0} \\
\cellcolor{gray!15}DaP (16-shot) & \cellcolor{gray!15}\textbf{2.6} & \cellcolor{gray!15}\textbf{4.7} & \cellcolor{gray!15}\textbf{4.3} & \cellcolor{gray!15}\textbf{8.7} \\
\bottomrule
\end{tabular}
\caption{\textbf{Steering specificity (\%)}.
How much steering changes the model perception of unrelated attributes.
Attribute flip: how often the yes-no answer about an unrelated attribute flips.
Name change: how often the object name changes.
NCC and the probe use 16 shots for All Attributes and 256 shots for the mean over the 4 Attribute Types.
}
\vspace{-5pt}
\label{tab:surgical}
\end{table}

In this section, we provide concrete examples of using DaP to probe and steer visual embeddings, as we think that this is more insightful than numbers on a benchmark. 
However, in an effort to prevent cherry-picking and to show the versatility and robustness of the method,
all experiments are done using the standard ``ironing man'' image~\cite{openai2023gpt4, liu2023llava}.
We provide additional qualitative results in the supplementary material~\ref{sec:app:patch_readouts} and~\ref{sec:app:steered_examples}.

\subsection{Verbalizing} 

Although DaP is estimated to answer about the concept of a given object,
we can use it to decode concepts at every patch of the image ($\mathcal{M} = \{i\}$).
\Cref{fig:patch_readout} shows readouts of DaP for Object Names, Color and Material concept subspaces, as well as the \concept{hanging} concept direction.
Text tokens predicted are scaled by the probe token probability.
By comparison with Emb-Lens, DaP outputs are more consistent, mainly in English, and more interpretable.
%
%
%
%

\begin{figure}[t]
    \centering
    \includegraphics[width=0.25\linewidth]{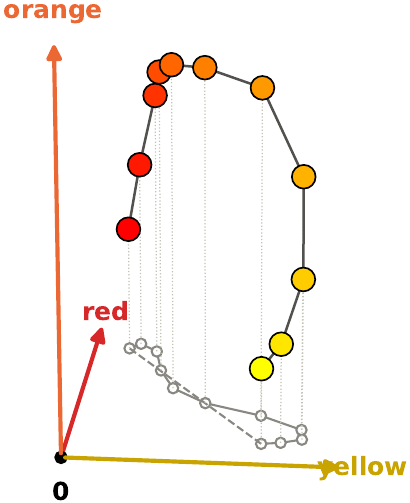}\hfill
    \includegraphics[width=0.32\linewidth]{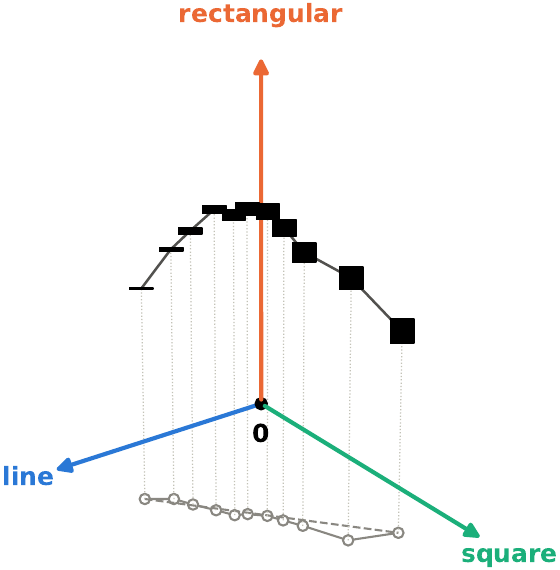}\hfill
    \includegraphics[width=0.32\linewidth]{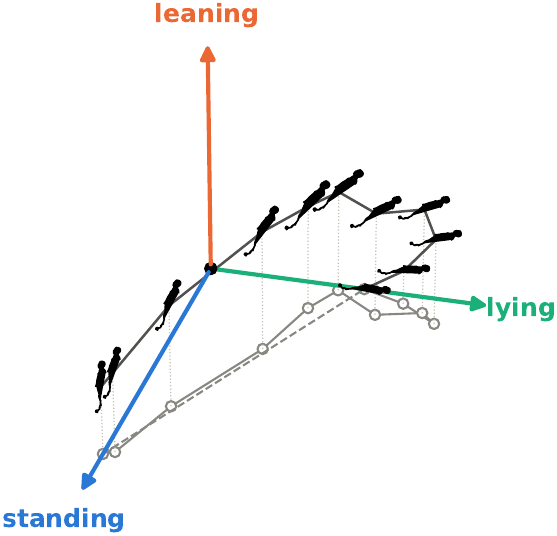}
    \caption{\textbf{Continuity of visual representations.}
    Three sequences of 11 images change from one concept to another, by interpolating the color in pixel space (left), growing the rectangle height (middle) and rotating the silhouette from $0^\circ$ to $90^\circ$ (right).
    Their embeddings are projected on the DaP concept directions shown as axes. The floor shows the shadow of the trajectory (solid) and the linear input path (dashed).
    }
    \label{fig:continuity_3d}
    \vspace{-5pt}
\end{figure}

\subsection{Steering}
 
One possible concern is that DaP only steers successfully in its restricted fitting setup
when directly asking the model about the attributes of interest.
This is why, in this section, we test steering the model by only asking indirect questions that do not directly query the attributes of interest.

\paragraph{Steering unseen concepts.}
For the first example, we ask the standard question \prompt{What is unusual about this image?}, but this time asking for two unusual events, while steering the color of the yellow taxi toward \concept{pink}.
Results are shown in \cref{fig:steer}(a).
Here, the DaP probe was estimated on neither \concept{yellow} nor \concept{pink} objects,
demonstrating its ability to generalize to unseen concept directions.

\paragraph{Removing a concept.}
Next, we noticed that the model justifies that the left taxi is moving based on its motion blur.
Conveniently, one of the VAW concepts we estimated was \concept{blurry}.
Once we remove the \concept{blurry} direction, the model says that the car has stopped (see \cref{fig:steer}(c)).

\paragraph{Steering several concepts.}
Next, we test modifying 2 concepts while adding 2 more concepts to the taxi on the same visual tokens.
For honesty, we noticed that when steering 4 attributes at the same time while asking indirect questions, the model may not surface all concepts in its answer, and the result is sensitive to prompt variations.
Here we selected the result in \cref{fig:steer}(e) as it shows:
(1) The unsteered answer surfaces the 2 modified concepts.
(2) The steered answer successfully surfaces the 4 new concepts.
(3) We also steer the model to perceive the taxi as \concept{parked} while keeping the blur, which results in the model hallucinating a cause for the blur (underlined).

\paragraph{Preserving other concepts.}
In most examples, steering does not impair the model's ability to perceive other concepts in the steered area.
For example, when adding \concept{in costume}, it continues to perceive the clothes as yellow (see \cref{fig:steer}(b)).
Or, more notably, the model retrieves
\prompt{with motion blur \dots{} visible roof-mounted light and tinted windows} after 4 attributes are added on top of the taxi.
Also, the coherence and fluency of the model's sentences seem unaffected by the steering. Notably, the steered and unsteered answers diverge at the moment the model reads the attribute 
(\eg, \prompt{The man likely feels a mix of frustration} $\rightarrow$ \prompt{\dots{} of cold}, see \cref{fig:steer}(d)).
%
%

\subsection{Case Study: Continuity} 

We synthetically generate image sequences that continuously morph from one concept to another.
We show the direction of the in-between concepts
(\concept{yellow}~\texttwemoji{yellow circle} $\rightarrow$ \concept{orange}~\texttwemoji{orange circle} $\rightarrow$ \concept{red}~\texttwemoji{red circle}, \concept{line}~\iconline{} $\rightarrow$ \concept{rectangle}~\iconrectangle{} $\rightarrow$ \concept{square}~\iconsquare{}, and \concept{standing man} $\rightarrow$ \concept{leaning man} $\rightarrow$ \concept{lying man}).
The results shown in \cref{fig:continuity_3d} are consistent with continuity.
This is not obvious because LVLM training could collapse visual representations to be discrete to match the language structure.
The trajectories project on a straight line in the plane of the source and target concepts, meaning the two concepts act as the two ends of a single axis~\citep{fel2026rabbit}. 

\section{Ablation}

\paragraph{Estimation.}
In \cref{tab:ablation_closed}, we estimate DaP at the last or the penultimate layer, and on all image tokens instead of the object mask.
The main benchmark samples object instances based on concept labels. 
To assess the impact of that sampling on the performance, we additionally estimate DaP on the same number of object instances drawn at random from the VAW training split.
For open questions, the diverse set of random instances slightly improves accuracy, while it degrades All Attributes accuracy.
Random instances rarely show a given VAW attribute, so DaP performance degrades without positive instances.
Thus, to probe a rare concept outside attribute types and object names, DaP needs positive instances of that concept.
Additionally, the supplementary material~\ref{sec:app:transfer} reports how Jacobian estimates transfer from one attribute type to another. 

\paragraph{Bias Gate.}
\Cref{tab:ablation} shows that the bias gate roughly doubles DaP faithfulness to the LVLM answers in open classification. 
While it is impressive that a linear readout can exactly predict the answer of the LLM Decoder one time out of three, results remain low. The remaining disagreements are often near misses (\tok{vehicle} instead of \tok{car}), so we additionally report top-5 agreement in \cref{tab:ablation}.
 
 
\begin{table}
\centering
\small\renewcommand{\arraystretch}{1}\setlength{\tabcolsep}{2.5pt}
\begin{tabular}{llccccc}
\toprule
 &  & All & \multicolumn{2}{c}{Attribute Types} & \multicolumn{2}{c}{Names} \\
\cmidrule(lr){4-5} \cmidrule(lr){6-7}
Layer & Estimation & Attributes & Seen & Unseen & Seen & Unseen \\
\midrule
$L-1$ & default & 73.5 & 75.5 & 53.3 & \textbf{78.7} & 24.3 \\
$L$ & default & \textbf{73.7} & 61.9 & 40.5 & \textbf{78.7} & 22.8 \\
\midrule
best & w/o mask & 70.6 & 75.7 & 52.9 & 71.6 & 22.7 \\
best & random & 66.7 & \textbf{75.8} & \textbf{55.0} & 75.5 & \textbf{25.0} \\
\bottomrule
\end{tabular}
\caption{Closed top-1 accuracy (\%), where each row changes one choice of the estimation method. 
best: best layer per column.
}
\vspace{-10pt}
\label{tab:ablation_closed}
\end{table}

\begin{table}
\centering
\small\renewcommand{\arraystretch}{1}\setlength{\tabcolsep}{4pt}
\begin{tabular}{lccccc}
\toprule
 & All & \multicolumn{2}{c}{Attribute Types} & \multicolumn{2}{c}{Names} \\
\cmidrule(lr){3-4} \cmidrule(lr){5-6}
Method & Attributes & Seen & Unseen & Seen & Unseen \\
\midrule
\cellcolor{gray!15}$J_\pi (\hat{\mathbf{v}} - \hat{\mathbf{v}}_0)$ & \cellcolor{gray!15}\textbf{76.0} & \cellcolor{gray!15}13.3 & \cellcolor{gray!15}14.7 & \cellcolor{gray!15}16.3 & \cellcolor{gray!15}24.2 \\
\midrule
+ $\mathbf{b}_\pi$ & 63.4 & \textbf{35.9} & 7.4 & 13.5 & 1.2 \\
+ $W_U \mathbf{b}_\pi$ gate & -- & 32.1 & \textbf{30.3} & \textbf{28.4} & \textbf{38.2} \\
\quad Top-5 & -- & \textit{64.7} & \textit{51.6} & \textit{51.8} & \textit{59.4} \\
\bottomrule
\end{tabular}
\caption{
Faithfulness (agreement of DaP with the LVLM next-token prediction).
The last row reports agreement with the top-5 answers of DaP.
}
\label{tab:ablation}
\end{table}

\section{Conclusion}

We introduce DaP, a general framework to verbalize, steer and study any concept direction given limited data.
%
Probes and Lenses are widely used in LVLM research, to improve both our understanding and our methods.
For example, they are used to improve training~\cite{jiang2025alignment, peng2024visualwords}, for interpretability~\cite{neo2025interpreting, fan2026visualtokens, min2026farlooksup, savietto2026geometry}, to mitigate hallucinations~\cite{jiang2025hallucinations, li2025hidden, liu2024reducing} or for safety monitoring~\cite{jiang2025hiddendetect}.
We believe that, due to its simplicity and data efficiency, DaP could directly apply and benefit each one of these research  lines. 
We release the code as well as some interactive demos.
 

\paragraph{Ethical Statement.}
As often in interpretability research, DaP can be a double-edged sword.
While we believe it enables better monitorability,
we are surprised by how efficiently DaP can enable white-box pixel-level attacks.
Examples are shown in the supplementary material\gf{}{~\ref{sec:app:attack}}, where DaP makes those attacks (1) cheaper in labels, (2) cheaper in compute, (3) less detectable and (4) robust to prompt variations.
We do not release the code for those experiments.

\paragraph{Limitations.}
The gradient backward pass necessitates a larger memory and compute footprint at the fitting stage (for a given image on a Tesla V100: open DaP: 185s, closed DaP: 0.37s).
While closed DaP can probe multi-token concepts such as \concept{riding a horse}, open DaP is limited to one-token concepts.
More DaP failure cases are stated in the experimental part and in the supplementary material~\ref{sec:app:failing}.

\section*{LLM Usage Statement}
We used Claude Code (Anthropic) in both the research and the writing of this paper.
In the research, it helped write and debug the experiment code, the analysis scripts, and the scripts that generate the tables and figures.
In the writing, it was used for grammar and spelling corrections and to suggest reformulations of individual sentences.
The method, the experimental design and all claims are the authors' own, and the authors checked every result, table and sentence of the paper.

{
    \small
    \bibliographystyle{ieeenat_fullname}
    \bibliography{main}
}

\clearpage
\appendix
\maketitlesupplementary

\section*{LLM Usage Statement}
We used Claude Code (Anthropic) in both the research and the writing of this paper.
In the research, it helped write and debug the experiment code, the analysis scripts, and the scripts that generate the tables and figures.
In the writing, it was used for grammar and spelling corrections and to suggest reformulations of individual sentences.
The method, the experimental design and all claims are the authors' own, and the authors checked every result, table and sentence of the paper.

More specifically:
\begin{itemize}[leftmargin=*, itemsep=2pt, topsep=2pt]
    \item \textbf{Writing:} in the main paper, we used it only for grammar corrections and reformulations. In the supplementary material, we used it for more extensive editing.
    \item \textbf{Review:} we gave the manuscript to ChatGPT (OpenAI) and Claude (Anthropic) to get reviewer-style feedback.
    \item \textbf{Citations:} we used it to copy the BibTeX entries of the cited papers and to check on the internet that the citations are correct.
    \item \textbf{Code:} it wrote and edited the code following the authors' instructions and coding rules, and the authors reviewed every change.
    \item \textbf{Experiments:} we used it to write the job scripts, and it sometimes gave a preliminary analysis of the results, which the authors checked.
    \item \textbf{Figures:} we used it to write the scripts that generate the tables, plots and figures, which the authors reviewed and edited.
    \item \textbf{Data:} we did not generate any data with AI models. We used it to help curate the label categories of VAW (merging synonyms and shades, excluding ambiguous labels) without changing the annotations of the dataset, and to translate the steering questions into Chinese.
    \item \textbf{Research ideas:} the ideas, hypotheses, claims and conclusions are the authors' own, with no use of AI tools.
\end{itemize}

\section{Summary}

\begin{itemize}
    \item \ref{sec:app:setup} Experimental Setup
    \item \ref{sec:app:faithfulness} Faithfulness
    \item \ref{sec:app:attack} Additional Experiments
    \begin{itemize}
        \item \ref{sec:app:pixel_attack} Pixel-Level Attack
        \item \ref{sec:app:patch_readouts} Patch-Level Readouts
        \item \ref{sec:app:steered_examples} Steering examples
        \item \ref{sec:app:transfer} Transferability
    \end{itemize}
    \item \ref{sec:app:models} Other LVLMs
    \item \ref{sec:app:steering_language} Steering in Another Language
    \item \ref{sec:app:failing} Failing Cases
    \item \ref{sec:app:linearization} Previous Linearization Methods
\end{itemize}

\section{Experimental Setup} \label{sec:app:setup}

\subsection{Details} 

The object region $\mathcal{M}$ contains the visual tokens whose center lies inside the bounding box.
We chose the gate size $k$ on validation sets.
Before $W_U$, we apply the final RMSNorm of the LLM Decoder, as the model does. Its weight is a diagonal linear map that folds into $W_U$, and its normalization only rescales each vector by a positive factor, which leaves the ranking of the tokens, and thus every prediction, unchanged.
This holds for most LVLMs and for every LVLM we tested, since their LLM Decoders all end with an RMSNorm.

\subsection{Prompt Design}

Each prompt ends with \prompt{Answer in one word.} (\prompt{Answer yes or no.} for All Attributes), and we start the model's reply with \prompt{Answer:}. 
The answer token then has a leading space, which makes more answer words a single token (\eg, \tok{teal} instead of \prompt{te} + \prompt{al}).
We note that we use the object name in some of the prompts, making the prompt not fixed at estimation.
However, we find that this improves the performance of the model by removing the ambiguity about which object the question refers to.
We did not really optimize prompts, and only kept those few design choices that improved the accuracy of the LVLM itself on the validation set.

\subsection{Data Config}

Since VAW labels were not designed to score free-form answers, and some attributes are synonyms or do not belong to their attribute type, we adapt them to our benchmark.
For example, we exclude \concept{murky} from the color attribute type, which VAW mostly assigns to water (\eg, a murky river).
Some materials that are not the material of the object (\eg, \concept{snowy}, \concept{furry}) or have too few instances are also excluded.
We merge shade labels into their base color (\eg, \concept{light yellow} and \concept{bright yellow} into \concept{yellow}, \concept{dark pink} into \concept{pink}), since a shade is a case of its base color, which the model answers and VAW usually labels as well. Also \concept{dark} and \concept{light} are separate VAW attributes evaluated in All Attributes.
The objective of these changes is to make the benchmark more reliable and consistent.

\subsection{Probing}
Each concept is scored by the logit of its answer token. Since the model may answer with a synonym, we add a few synonyms per concept and keep the highest logit among them (\eg, \tok{grey} for \concept{gray}). We score DaP, Emb-Lens and the LVLM with this same readout, so that their accuracies are directly comparable.

\subsection{Steering}

For Attribute Types and Object Names, we steer 100 test objects per seen concept (400 in total) (52 for Activity, whose smallest seen concept has only 52 test objects (208 in total)), each toward one target concept drawn at random, the same for every method. For All Attributes, we steer 8 positive and 8 negative test objects per attribute toward the answer opposite to their label (9,120 steering tests). Steering succeeds when the top-1 answer of the model over the whole vocabulary is the target concept.
The steering questions reword the estimation questions into \prompt{What is the \ph{attribute type} of the \ph{object name}?}, \prompt{Name the thing in the middle of the image.} and \prompt{Would you describe the \ph{object name} as \ph{attribute}?}.

\section{Faithfulness} \label{sec:app:faithfulness}
\Cref{tab:faithfulness} reports the faithfulness of each linear probe, measured as the share of test objects on which its top-1 concept is the top-1 concept of the LVLM.
\begin{table*}
\centering
\small\renewcommand{\arraystretch}{1}\setlength{\tabcolsep}{4pt}
\begin{tabular}{llc|c|cccc|c}
\toprule
 & Method & Shots & All Attributes & Color & Material & Shape & Activity & Names \\
\midrule
 & NCC~\cite{panickssery2024caa} & 16 & 66.6 & 48.2 & 61.2 & 36.7 & 71.6 & 78.8 \\
 & Probe~\cite{savietto2026geometry} & 16 & \underline{68.2} & 54.5 & 64.8 & 48.9 & 80.2 & \textbf{81.5} \\
 & Emb-Lens~\cite{fan2026visualtokens} & 0 & -- & 51.2 & 64.5 & 49.9 & 55.5 & 51.8 \\
 & \cellcolor{gray!15}DaP & \cellcolor{gray!15}1 & \cellcolor{gray!15}65.0 & \cellcolor{gray!15}\underline{69.7} & \cellcolor{gray!15}\underline{75.8} & \cellcolor{gray!15}\underline{65.5} & \cellcolor{gray!15}\underline{86.0} & \cellcolor{gray!15}\underline{79.0} \\
 & \cellcolor{gray!15}DaP & \cellcolor{gray!15}16 & \cellcolor{gray!15}\textbf{76.0} & \cellcolor{gray!15}\textbf{73.8} & \cellcolor{gray!15}\textbf{82.2} & \cellcolor{gray!15}\textbf{73.4} & \cellcolor{gray!15}\textbf{87.9} & \cellcolor{gray!15}75.0 \\
\cmidrule(l){2-9}
\multirow{-6}{*}{\rotatebox[origin=c]{90}{\shortstack{\textit{Seen (4-way)}}}} & \textit{Chance} & \textit{--} & \textit{58.6} & \textit{25.9} & \textit{37.5} & \textit{45.9} & \textit{46.7} & \textit{62.6} \\
\midrule\midrule
 & \# classes &  & -- & 27 & 10 & 17 & 21 & 1313 \\
 & Emb-Lens~\cite{fan2026visualtokens} & 0 & -- & 25.9 & 42.6 & 19.4 & 25.3 & 11.1 \\
 & \cellcolor{gray!15}DaP & \cellcolor{gray!15}1 & \cellcolor{gray!15}-- & \cellcolor{gray!15}\underline{47.3} & \cellcolor{gray!15}\underline{66.6} & \cellcolor{gray!15}\underline{50.4} & \cellcolor{gray!15}\underline{26.4} & \cellcolor{gray!15}\underline{36.1} \\
 & \cellcolor{gray!15}DaP & \cellcolor{gray!15}16 & \cellcolor{gray!15}-- & \cellcolor{gray!15}\textbf{54.9} & \cellcolor{gray!15}\textbf{74.1} & \cellcolor{gray!15}\textbf{53.7} & \cellcolor{gray!15}\textbf{50.2} & \cellcolor{gray!15}\textbf{41.0} \\
\cmidrule(l){2-9}
\multirow{-5}{*}{\rotatebox[origin=c]{90}{\shortstack{\textit{Unseen}}}} & \textit{Chance} & \textit{--} & \textit{--} & \textit{22.4} & \textit{25.0} & \textit{14.9} & \textit{13.9} & \textit{2.8} \\
\bottomrule
\end{tabular}
\caption{Agreement (\%) with the model on VAW: the classifier's top-1 class is the model's top-1 class, object by object. Rows and columns as in Table~\ref{tab:benchmark}. Chance always predicts the concept that the model predicts most often in the column (for All Attributes, its most frequent answer for each attribute), so its agreement is the share of objects that receive this answer. Bold: best, underlined: second best (chance excluded).}
\label{tab:faithfulness}
\end{table*}

\section{Additional Experiments} \label{sec:app:attack}

\subsection{Pixel-Level Attack} \label{sec:app:pixel_attack}
Pixel-level attacks steer the model without editing its internal activations, by modifying the pixels of the input image instead.
Traditionally, pixel-level attacks on LVLMs take the answer of the model as the objective.
With DaP, the objective can be set on the output of the vision encoder $\mathbf{V}$ instead of the next-token prediction $\mathbf{h}$, so that only the model perception of the image is altered.
This makes the attack robust to prompt variations, since $\mathbf{V}$ does not depend on the prompt.
It also makes the attack cheaper, since during the costly optimization of the pixels, the backward pass goes through the vision encoder only and not through the LLM Decoder.
Compared to the trained probe, DaP gives directions that need less steering strength, so the attack needs a smaller budget.
DaP also needs fewer labels of the concept to attack.
Here we demonstrate an attack on 2 examples.

\begin{figure*}[t]
    \centering
    \includegraphics[width=0.32\linewidth]{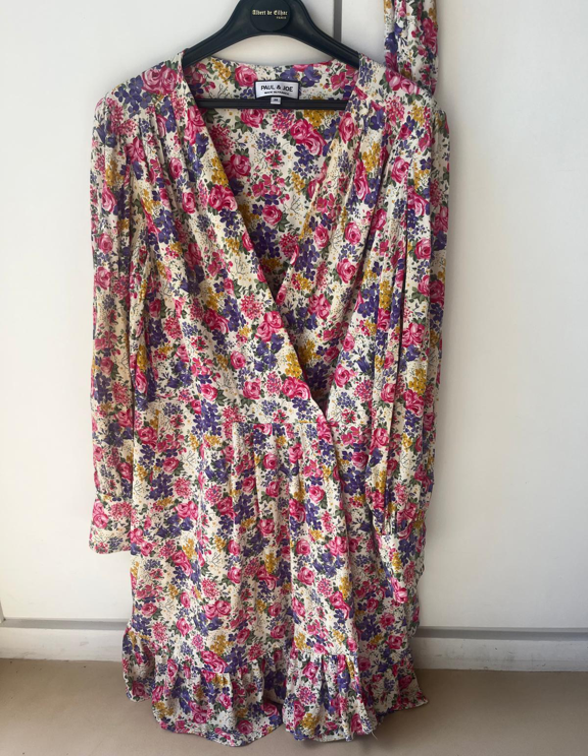}\hfill
    \includegraphics[width=0.32\linewidth]{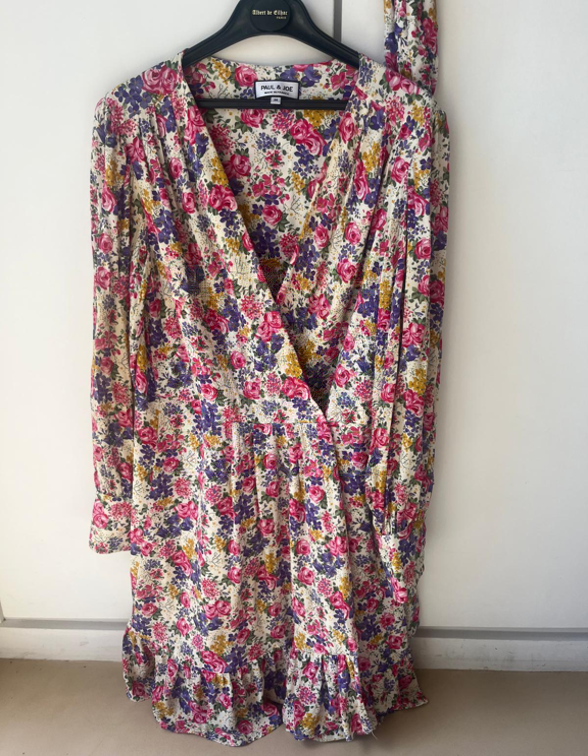}\hfill
    \includegraphics[width=0.32\linewidth]{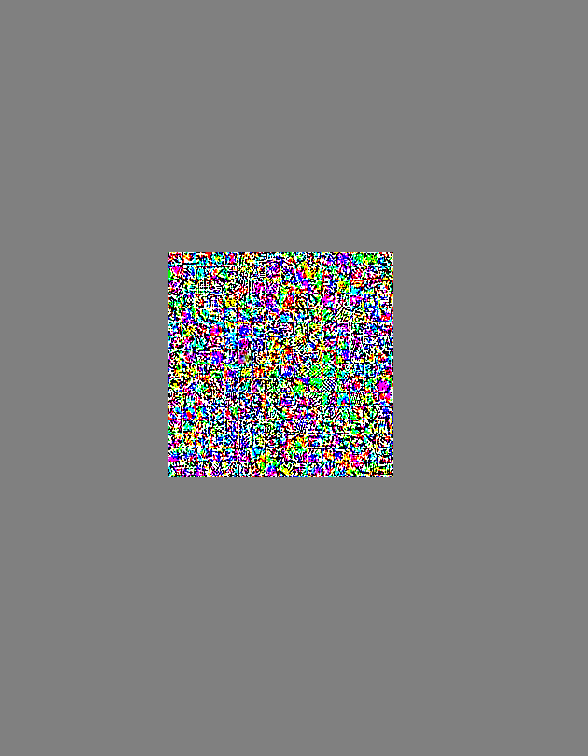}
    \caption{\textbf{Pixel-level attack of an attribute with DaP.}
    Clean image (left), image attacked toward \concept{damaged} on Ministral 3 3B (middle), and amplified perturbation (right).}
    \label{fig:pixel_attack_dress}
\end{figure*}

\begin{figure}[tp]
    \centering
    \includegraphics[width=\linewidth]{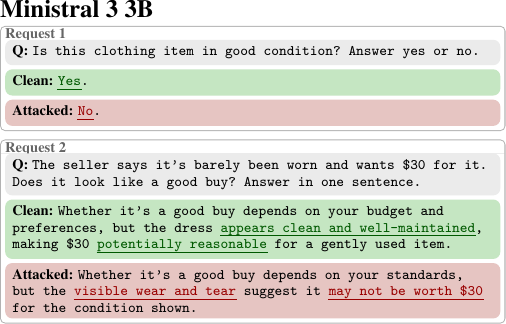}\\[8pt]
    \includegraphics[width=\linewidth]{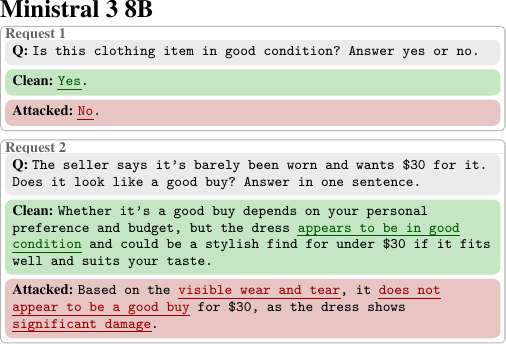}\\[8pt]
    \includegraphics[width=\linewidth]{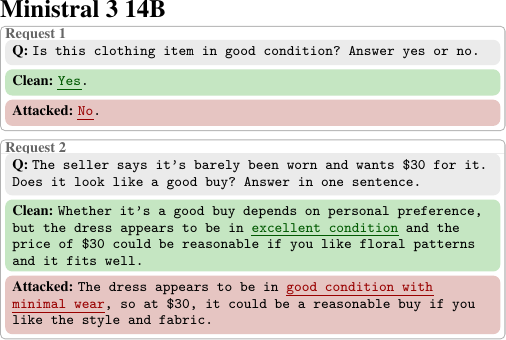}
    \caption{\textbf{API answers to the attacked dress.}
    Each request is separate and contains only the image and its question.
    \nosteerhl{Green} marks the answer for the clean image, and \steerhl{red} the answer for the attacked image.}
    \label{fig:pixel_attack_dress_api}
\end{figure}

First, we attack an attribute. Specifically, we make the model perceive a dress that could be sold on a marketplace as damaged.
\Cref{fig:pixel_attack_dress} shows a dress attacked toward \concept{damaged} with DaP on the vision encoder of Ministral 3 3B, and \cref{fig:pixel_attack_dress_api} the answers of the Ministral 3 models through the OpenRouter API.
The attack is invisible to the human eye and works.
Interestingly, models of the same family share a similar vision encoder, so the attack works on larger models of the same family.
The attack is also 2.5 times faster than attacking with the LLM Decoder, since the forward and backward passes of each step go through the vision encoder only.
It is even more efficient, since we attack a small model and use the attacked image on larger ones.
The attack is robust to prompt variations, as it works here on 2 prompts.

\begin{figure}[t]
    \centering
    \includegraphics[width=\linewidth]{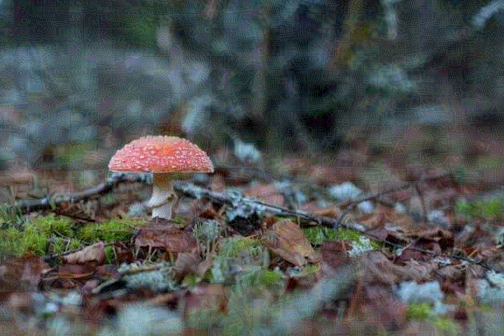}
    \caption{\textbf{Pixel-level attack with DaP.} Image attacked on Ministral 3 3B (with a visible attack: $L_\infty$ budget of $8/255$). Through the OpenRouter API, Ministral 3 14B describes this mushroom as young morels (edible mushroom) instead of a fly agaric (poisonous mushroom).}
    \label{fig:pixel_attack}
\end{figure}

Second, we change the identity of an object. Specifically, we make the model perceive a poisonous fly agaric as an edible mushroom.
We estimate DaP on a small dataset of mushroom photos, with 16 photos of each species.
Changing the identity needs a bigger budget than changing an attribute, so we let the attack modify every pixel of the image.
The resulting noise makes the attack visible in the blurry areas of the image.
\Cref{fig:pixel_attack} shows an image attacked with DaP on Ministral 3 3B ($L_\infty$ budget of $8/255$), then sent to Ministral 3 14B through the OpenRouter API.
For the question \prompt{Describe this mushroom in two sentences.}, the API response for the clean image is \prompt{This mushroom features a small, conical cap covered with white, scale-like spots on a reddish-orange background. Its slender, white stem rises from a bed of forest debris, giving it a delicate yet distinctive appearance.}, whereas the API response for the attacked image is \prompt{The mushroom in the image appears to be a cluster of young morel mushrooms, identifiable by their distinctive conical shape and honeycomb-like texture on the caps. These mushrooms typically grow in forested areas during spring and are highly prized by foragers for their unique flavor.}

\refstepcounter{subsection}\label{sec:app:sampling}

\subsection{Patch-Level Readouts} \label{sec:app:patch_readouts}

\begin{figure*}[p]
    \centering
    \includegraphics[height=0.379\linewidth]{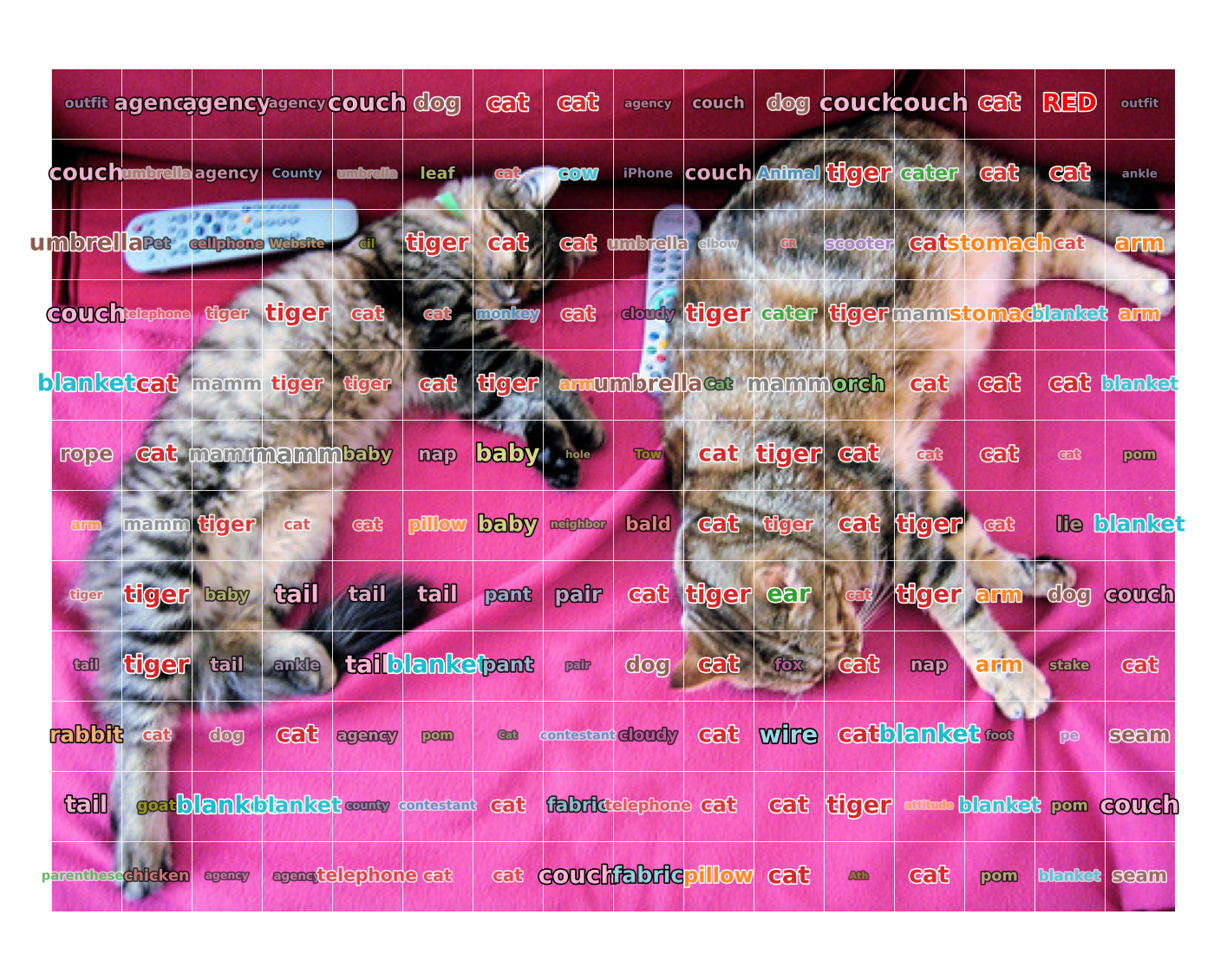}\hfill
    \includegraphics[height=0.379\linewidth]{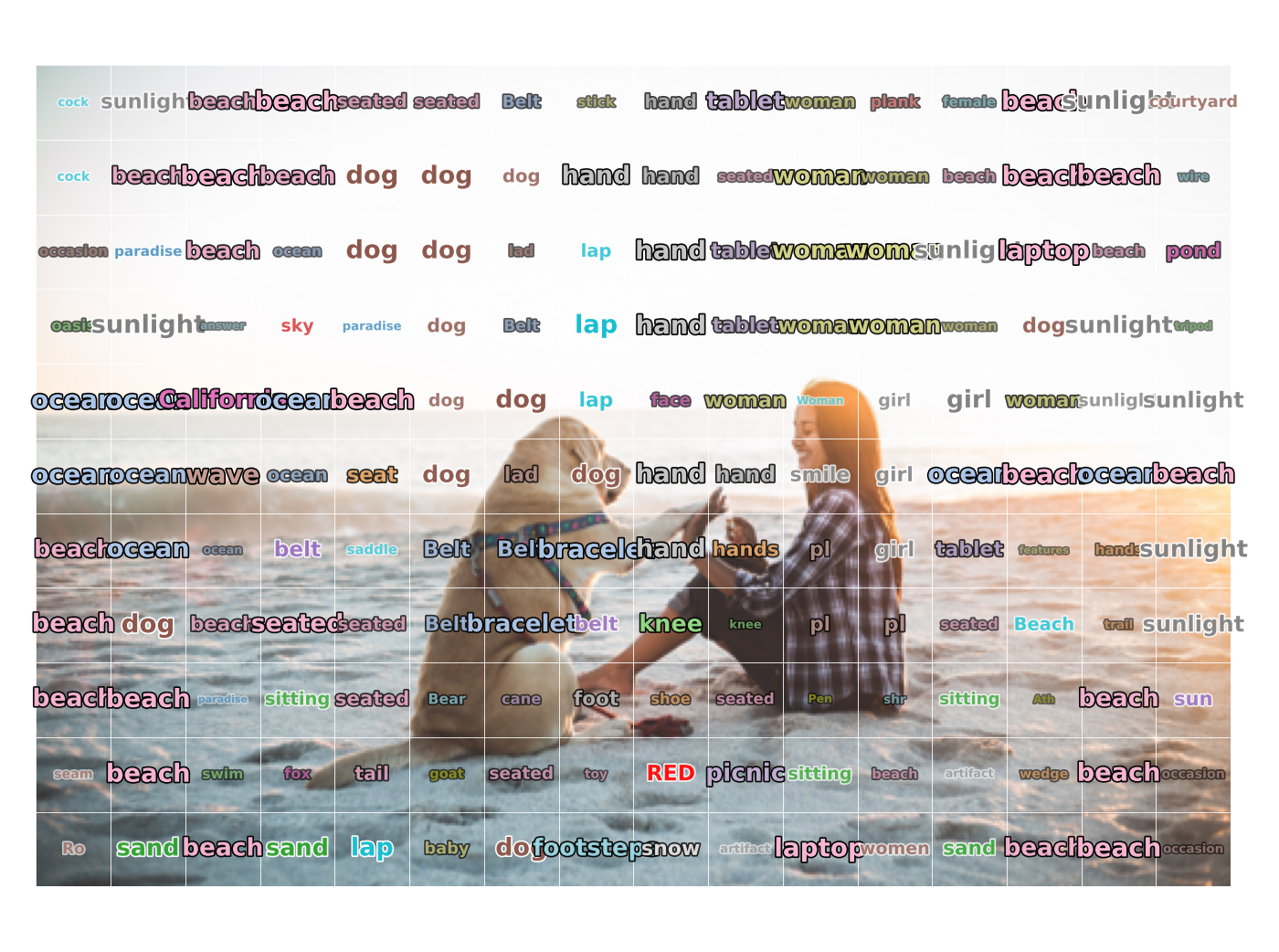}\\[4pt]
    \includegraphics[height=0.364\linewidth]{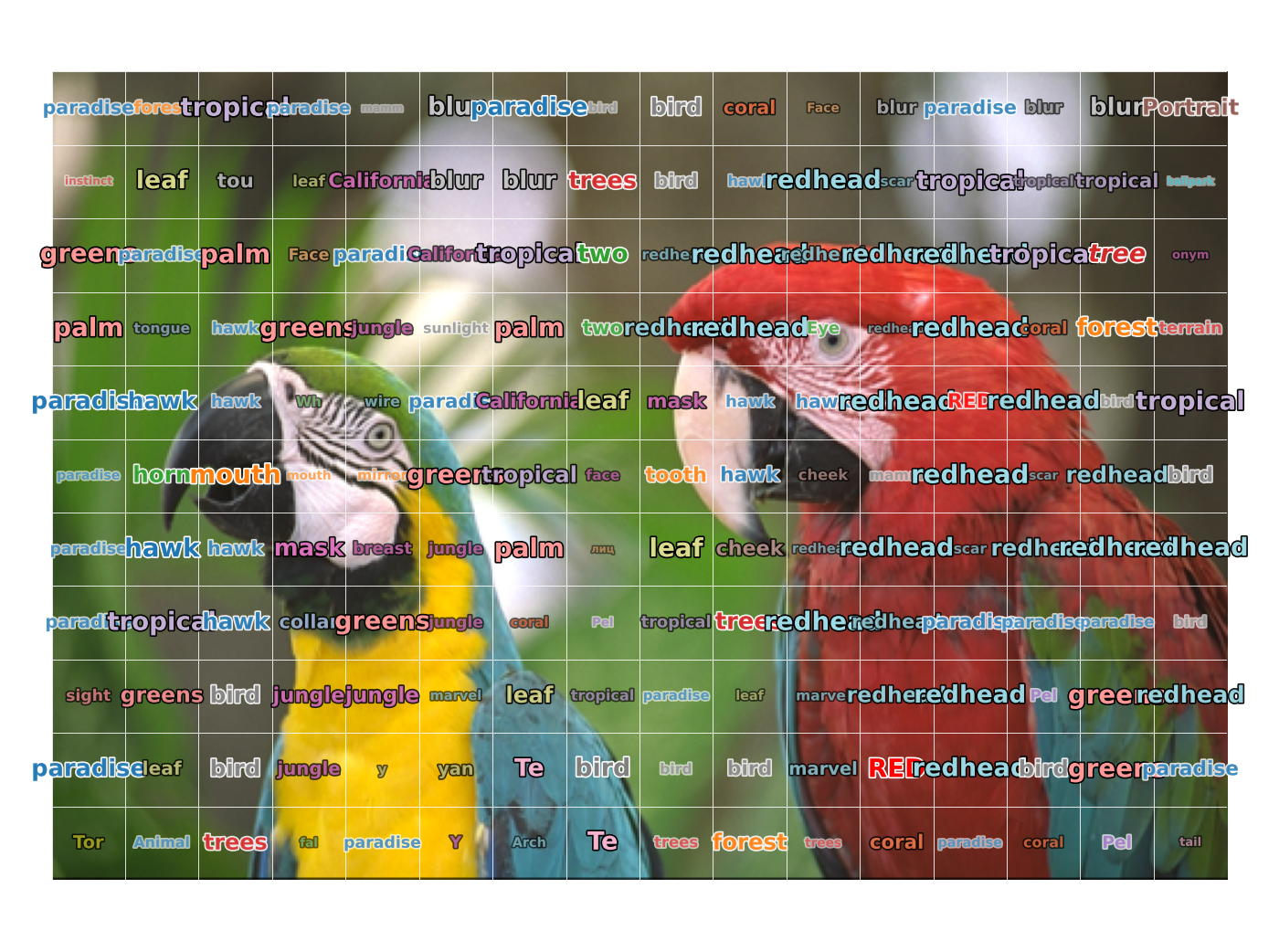}\hfill
    \includegraphics[height=0.364\linewidth]{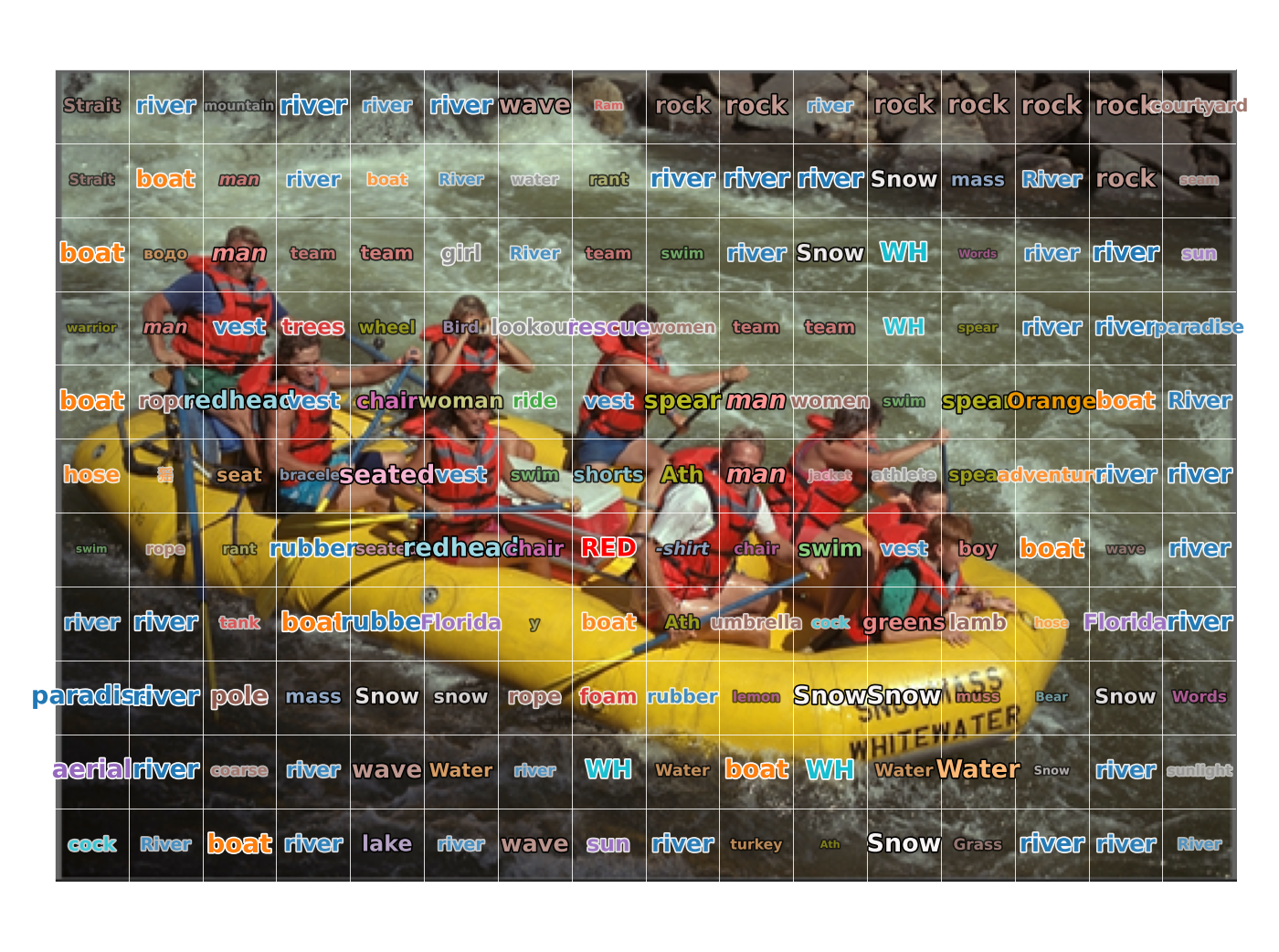}\\[4pt]
    \includegraphics[height=0.289\linewidth]{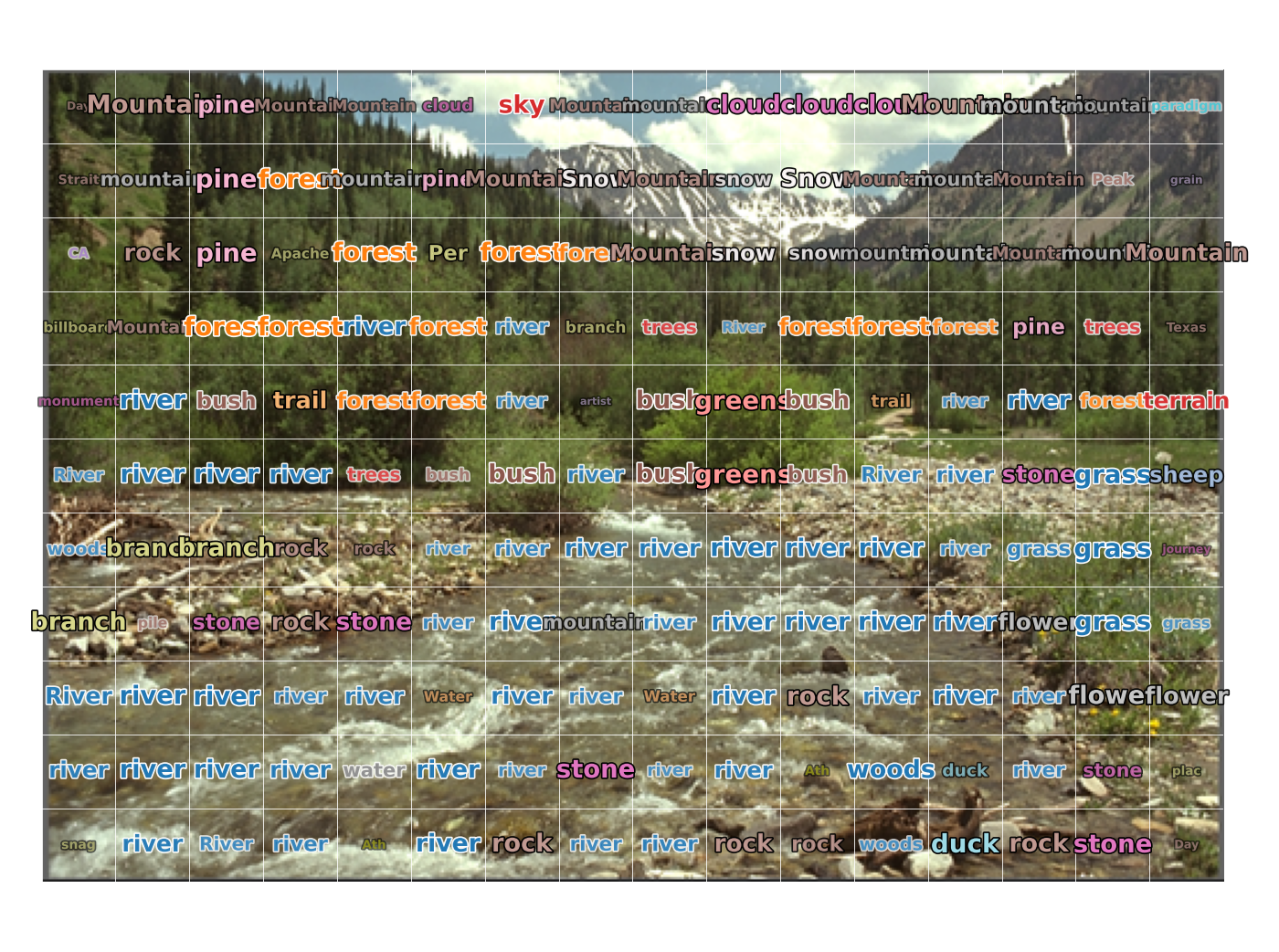}\hfill
    \includegraphics[height=0.289\linewidth]{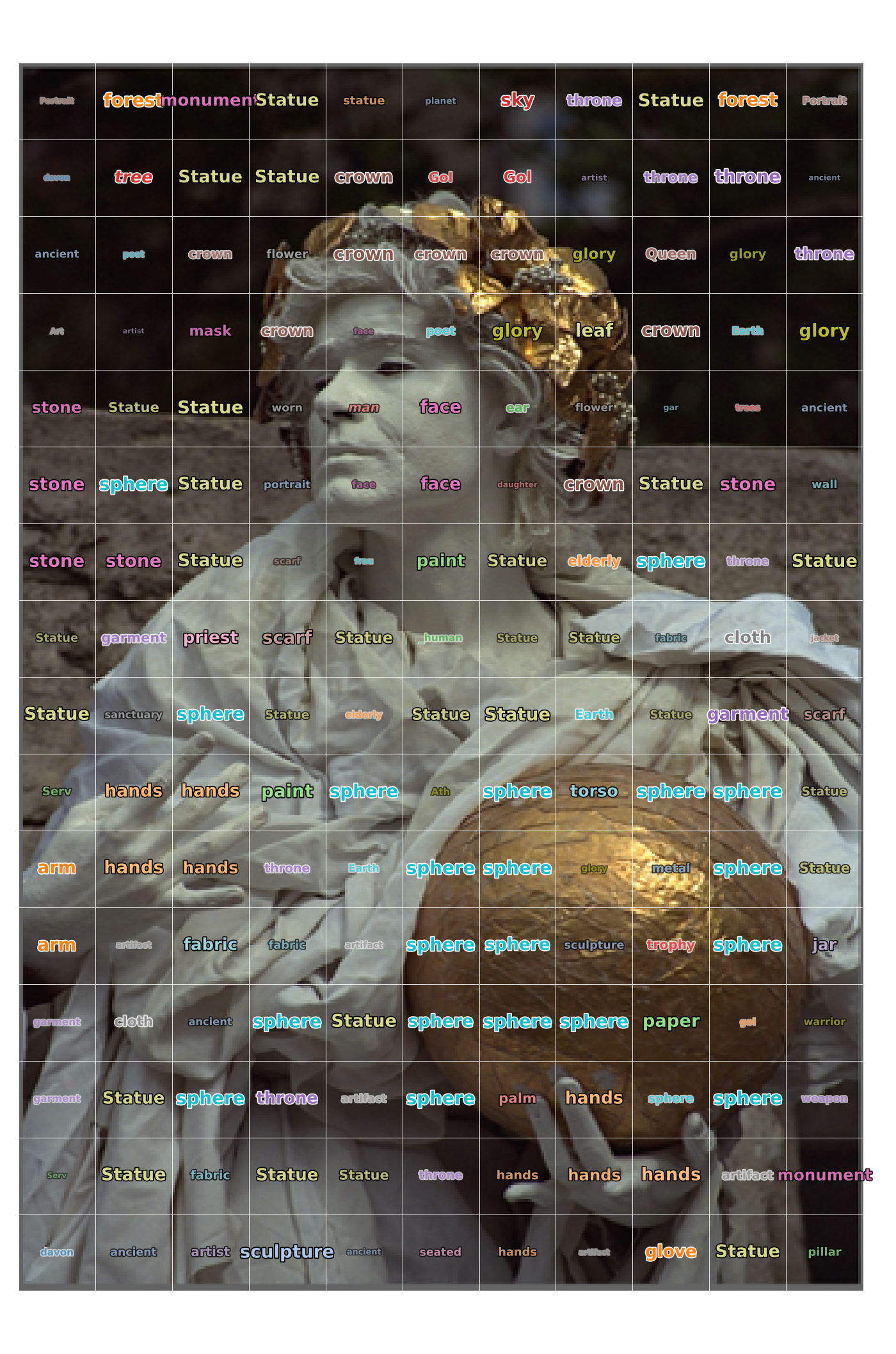}\hfill
    \includegraphics[height=0.289\linewidth]{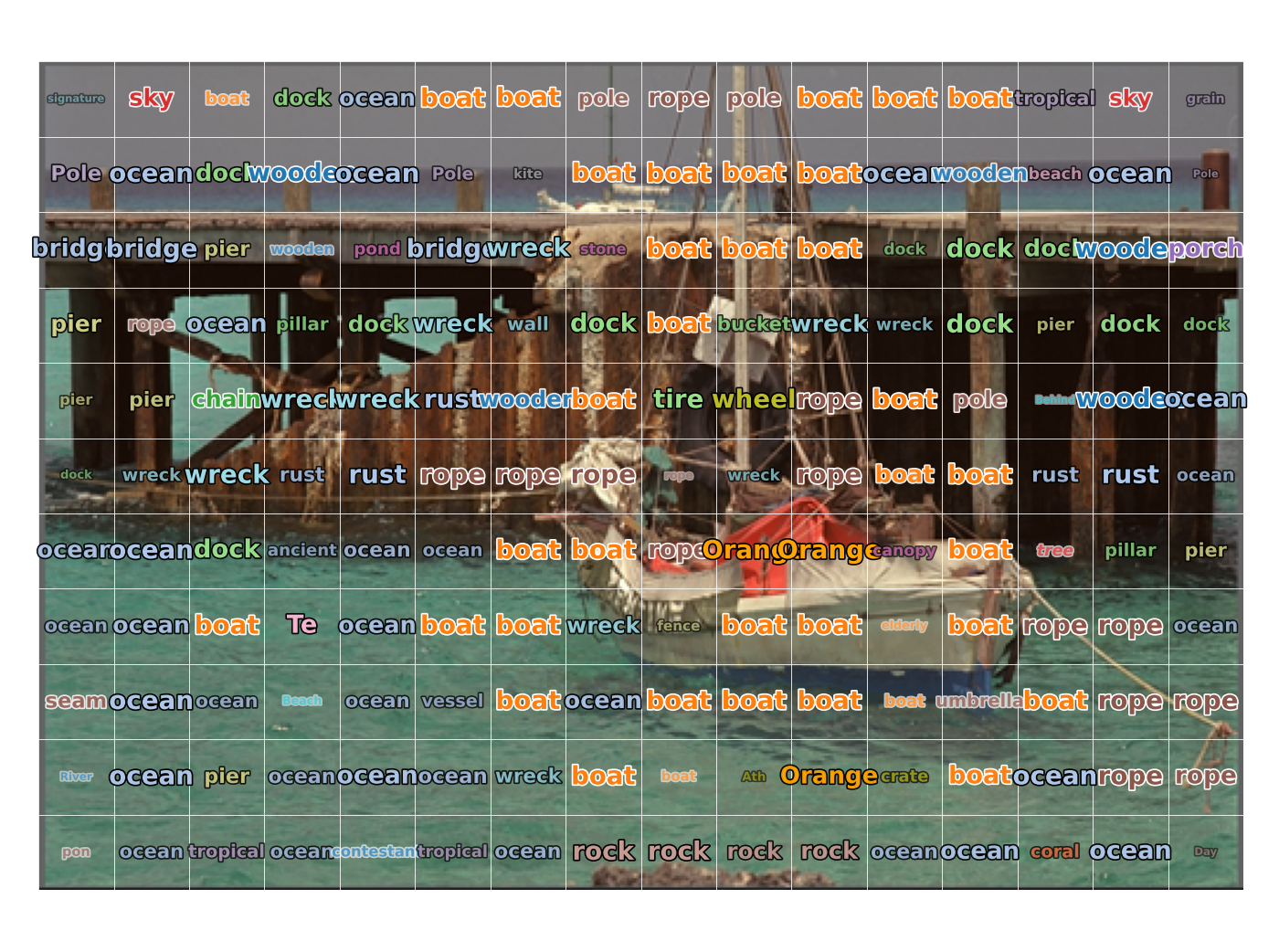}
    \caption{\textbf{Patch-level readouts of object names.}
    Top-1 token prediction of DaP estimated with \protect\prompt{Which word best names the thing in the middle of the image?} on each visual token.
    Word size and opacity grow with the probability of the token.
    Italics mark the answer tokens of the seen concepts.}
    \label{fig:supp_readout_object}
\end{figure*}

\begin{figure*}[p]
    \centering
    \includegraphics[height=0.379\linewidth]{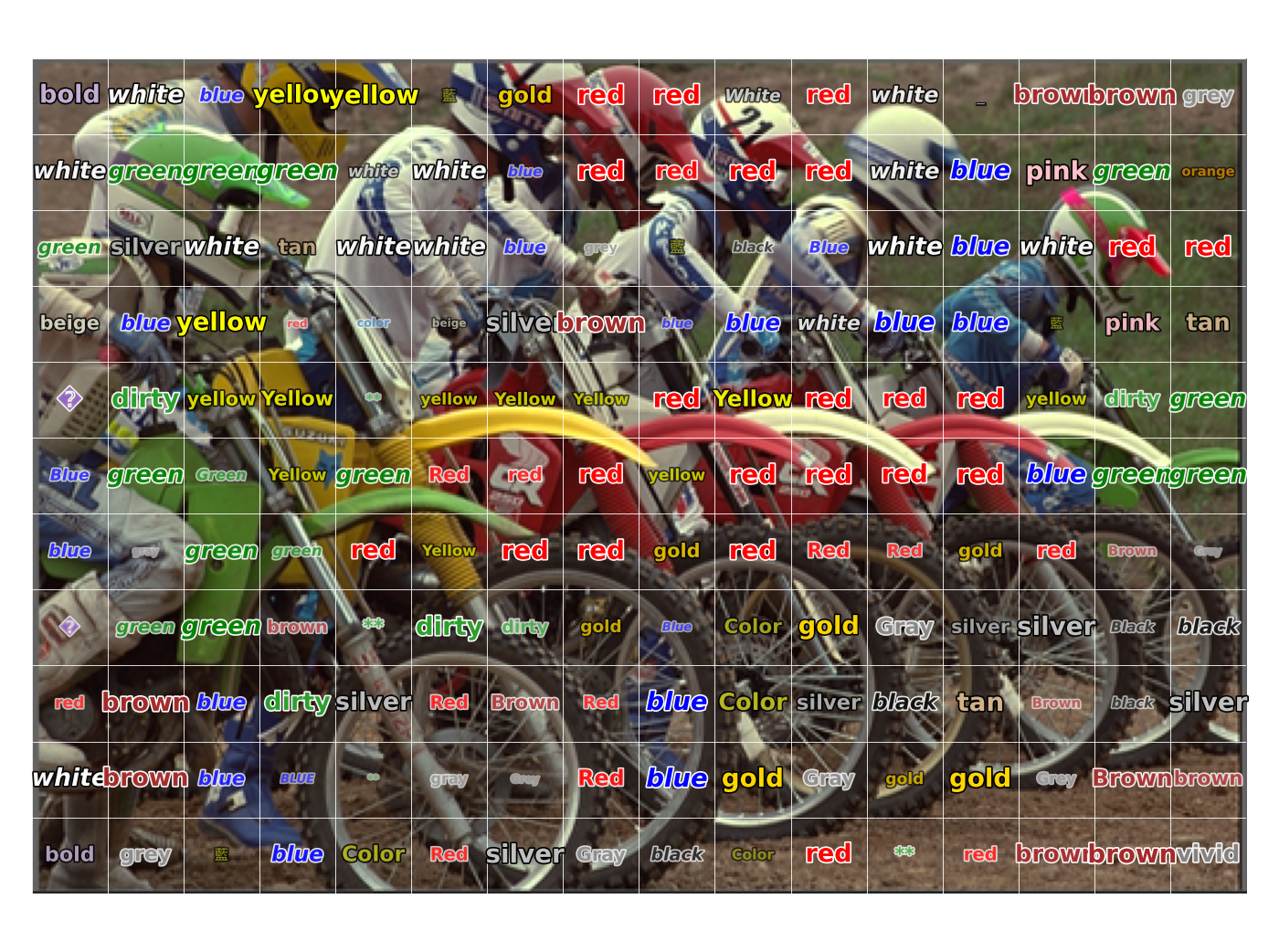}\hfill
    \includegraphics[height=0.379\linewidth]{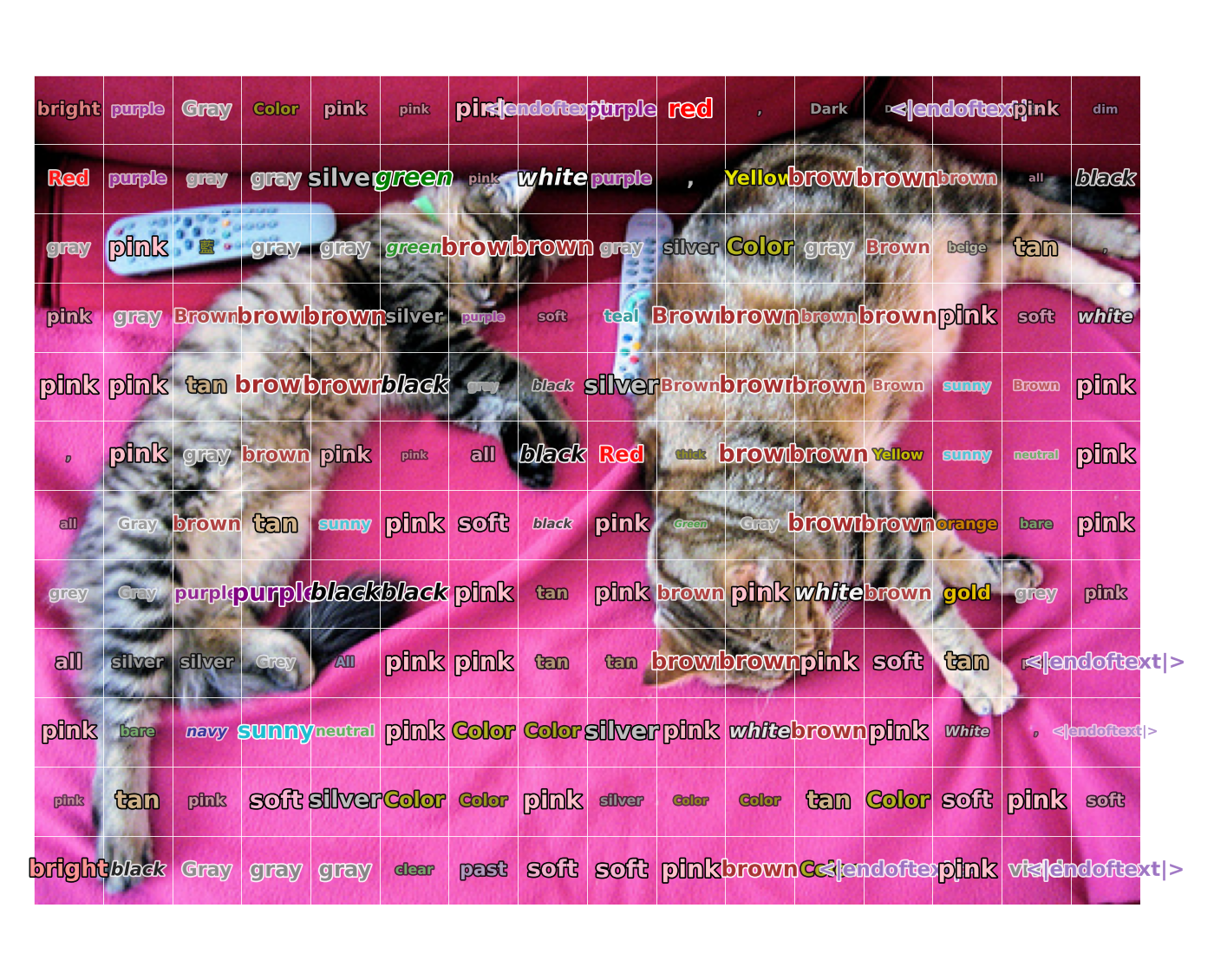}\\[4pt]
    \includegraphics[height=0.363\linewidth]{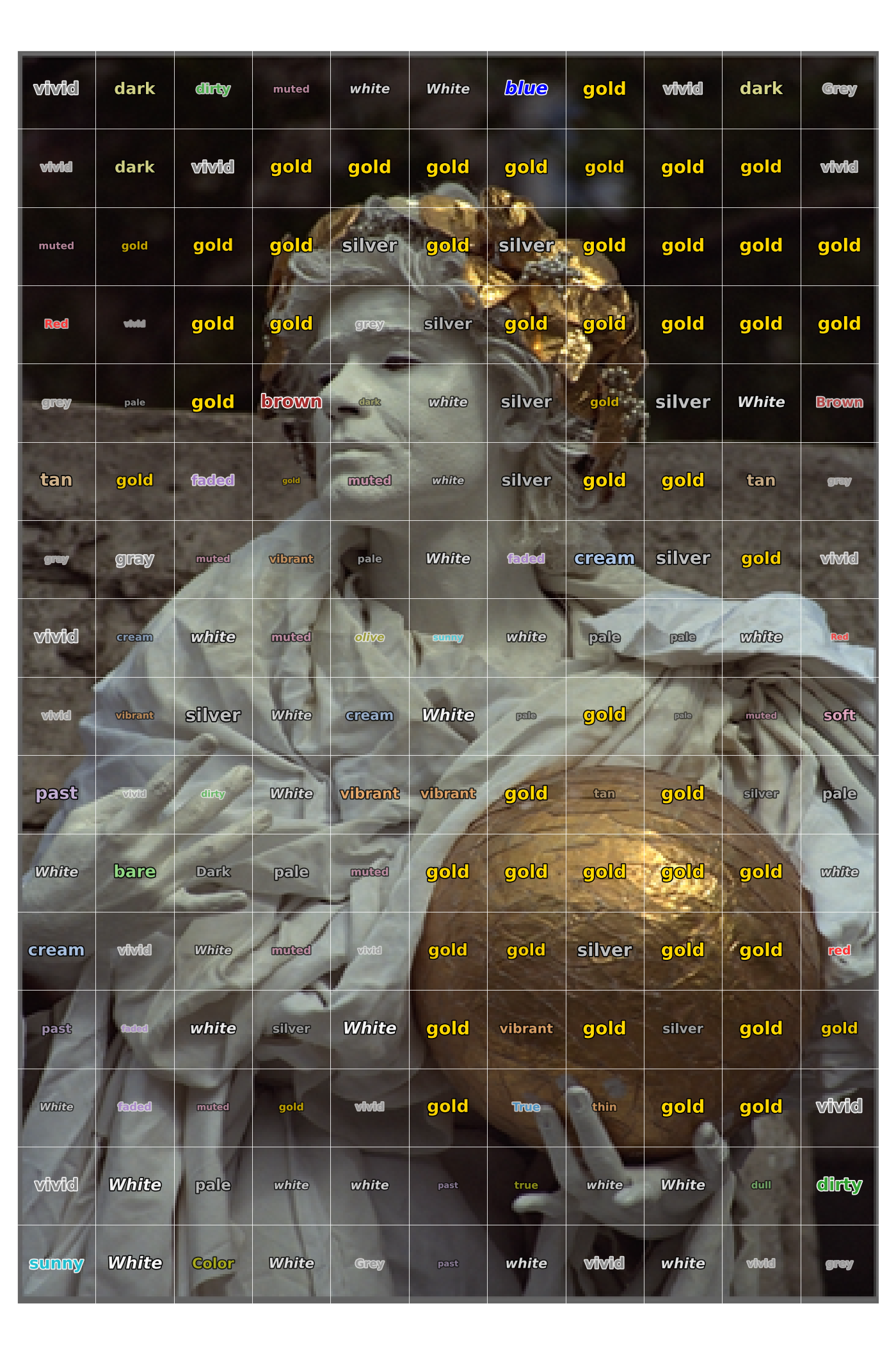}\hfill
    \includegraphics[height=0.363\linewidth]{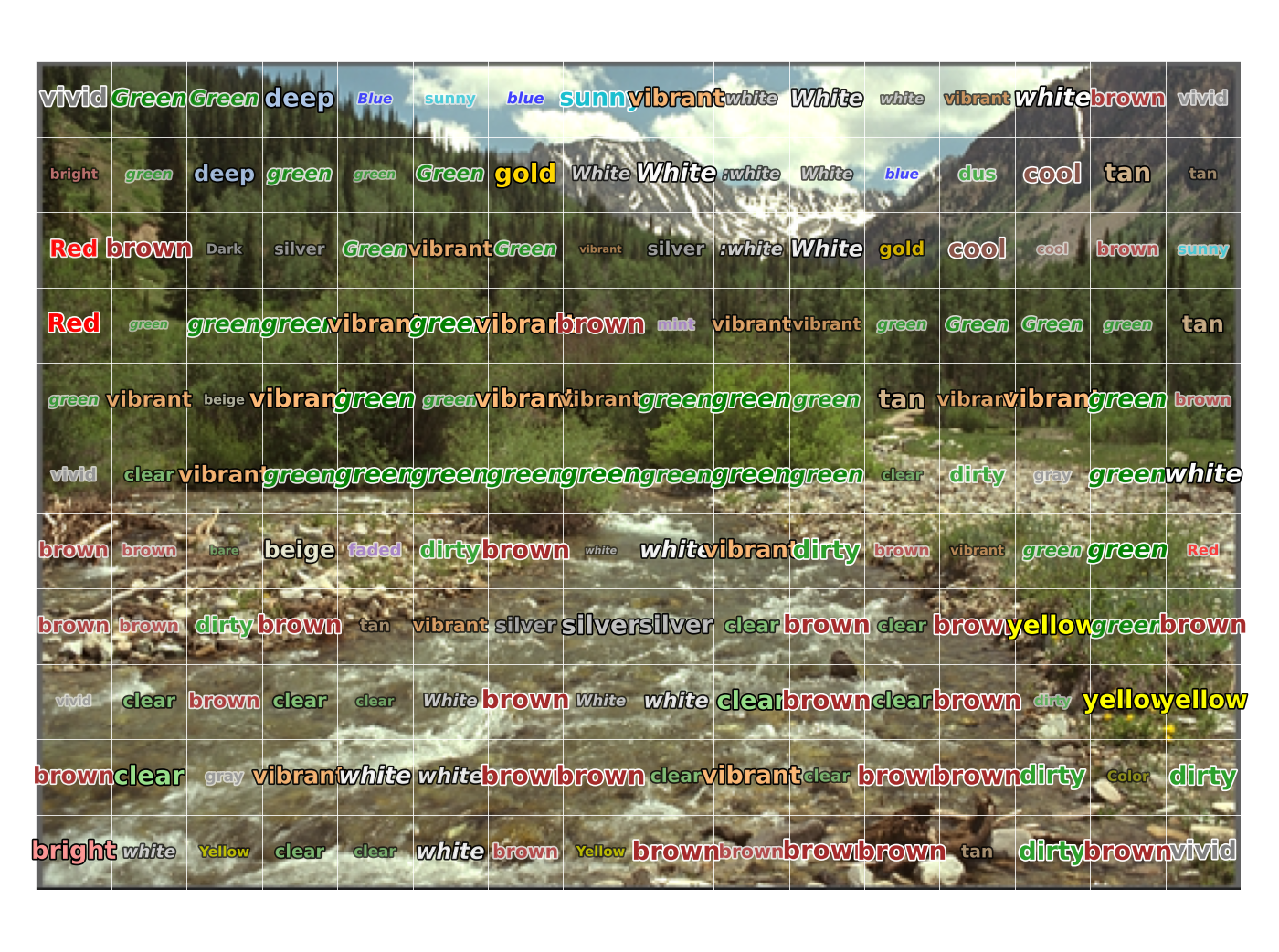}\hfill
    \includegraphics[height=0.363\linewidth]{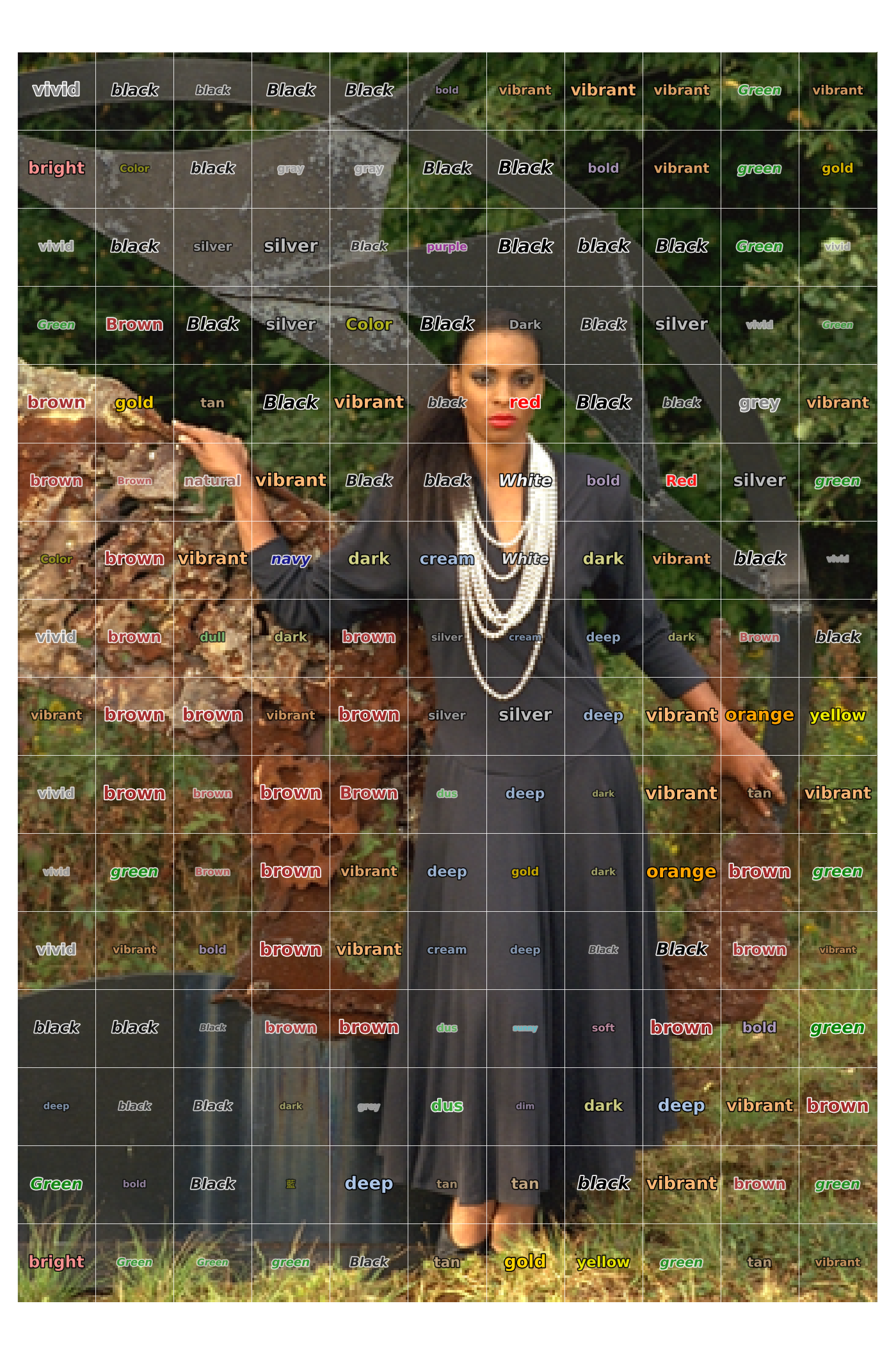}
    \caption{\textbf{Patch-level readouts of colors.}
    Top-1 token prediction of DaP estimated with \protect\prompt{Which word best describes the color of the \protect\ph{object name}?} on each visual token, as in \cref{fig:supp_readout_object}.}
    \label{fig:supp_readout_color}
\end{figure*}

\begin{figure*}[p]
    \centering
    \includegraphics[height=0.364\linewidth]{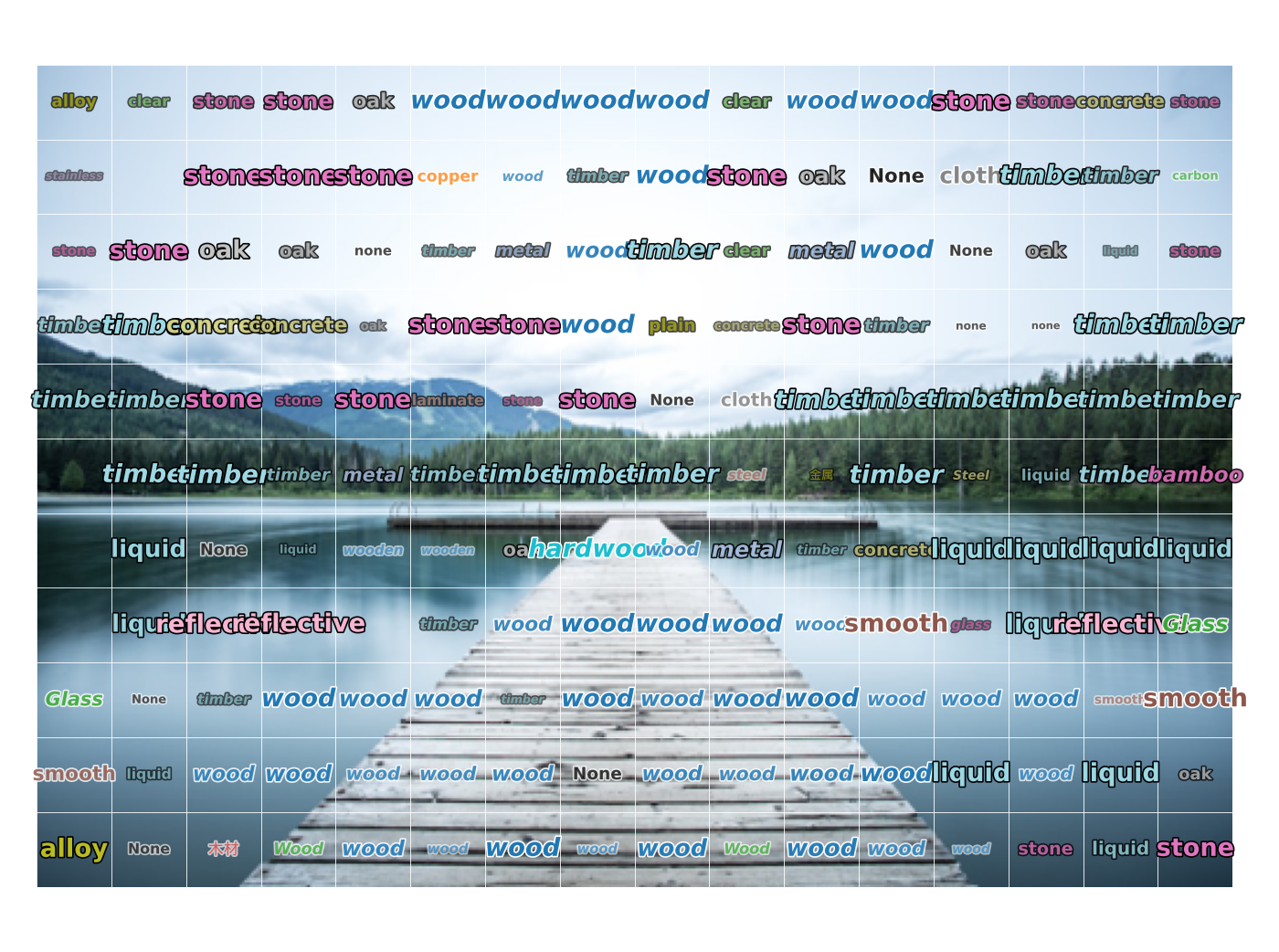}\hfill
    \includegraphics[height=0.364\linewidth]{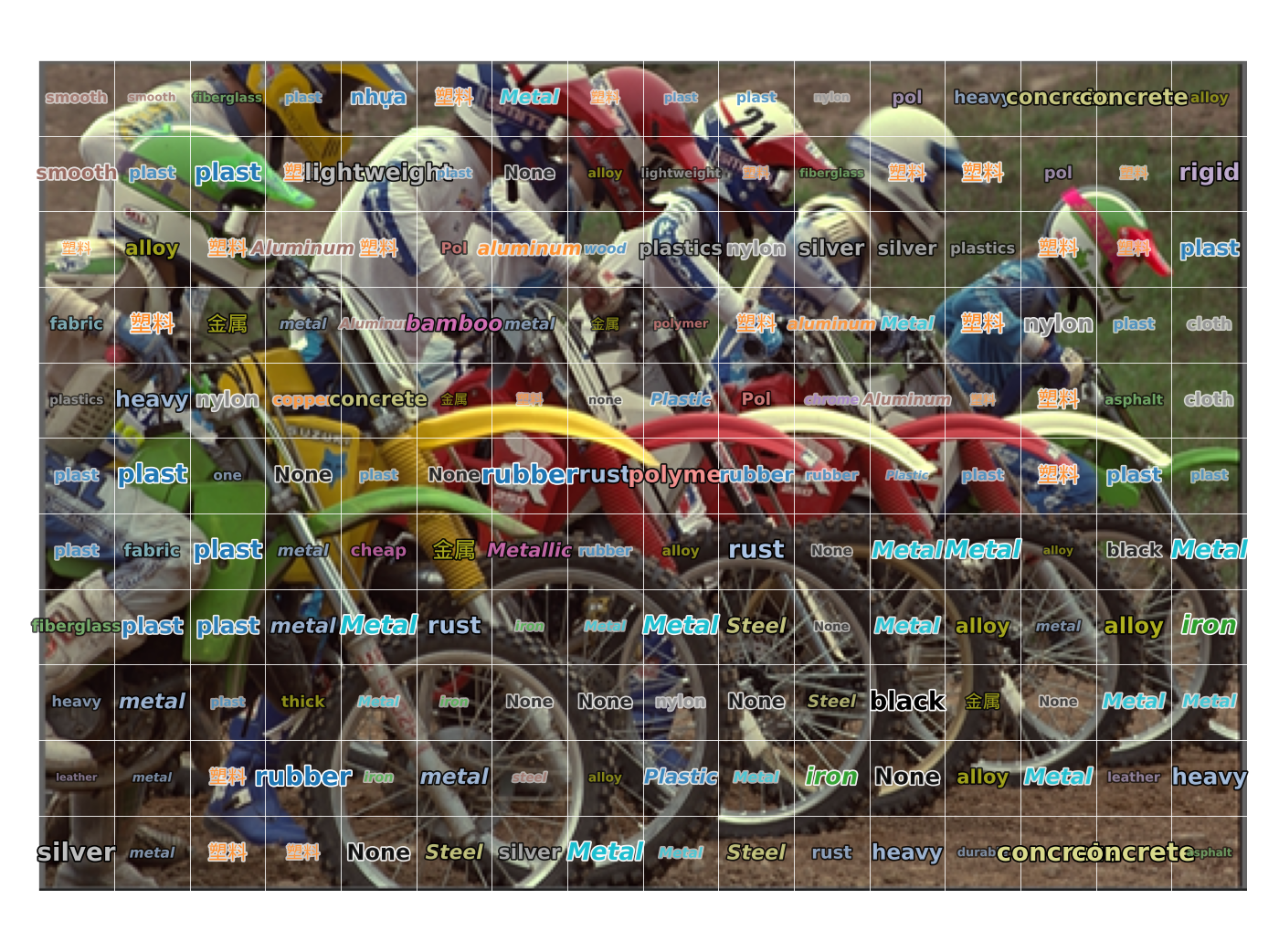}\\[4pt]
    \includegraphics[height=0.298\linewidth]{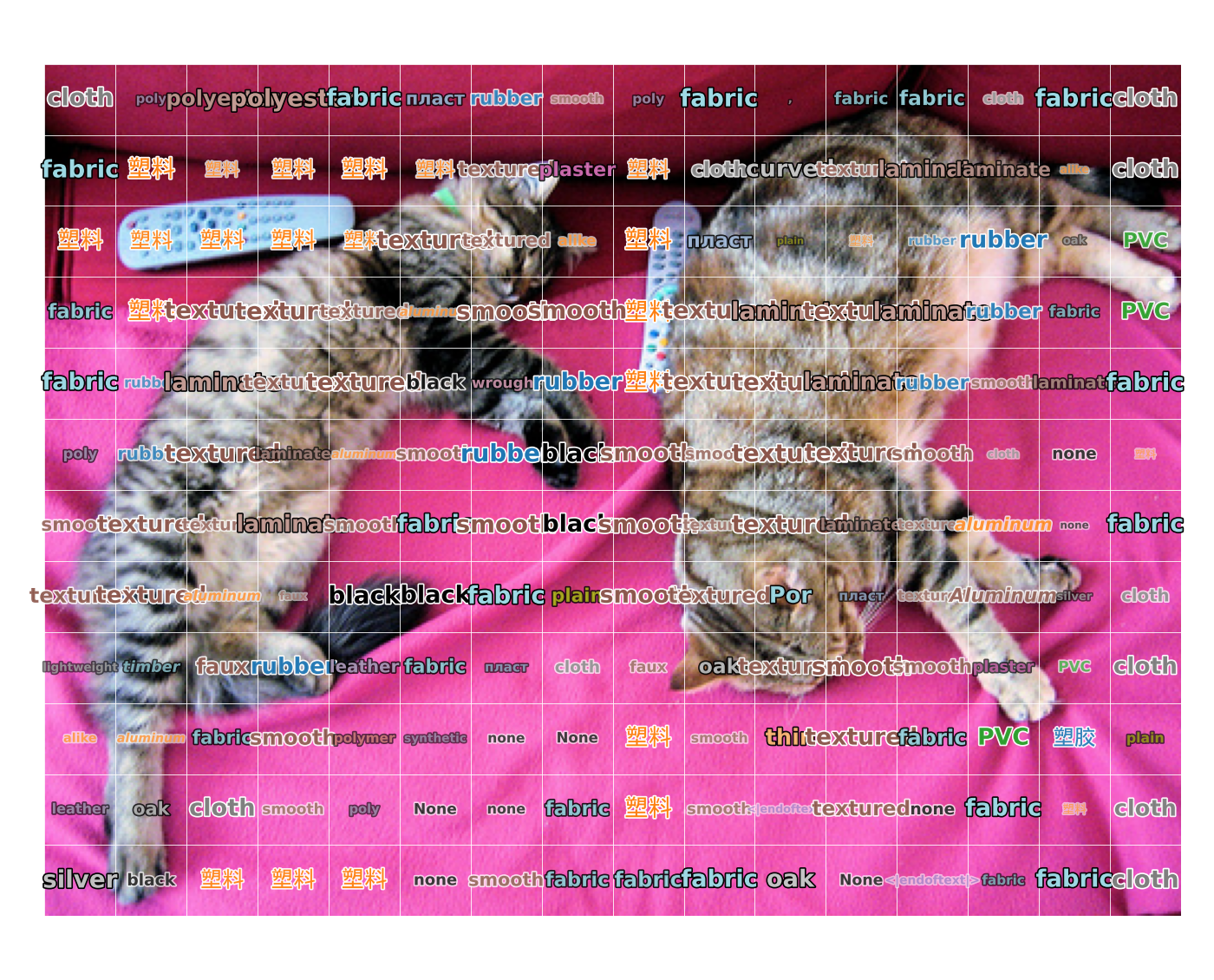}\hfill
    \includegraphics[height=0.298\linewidth]{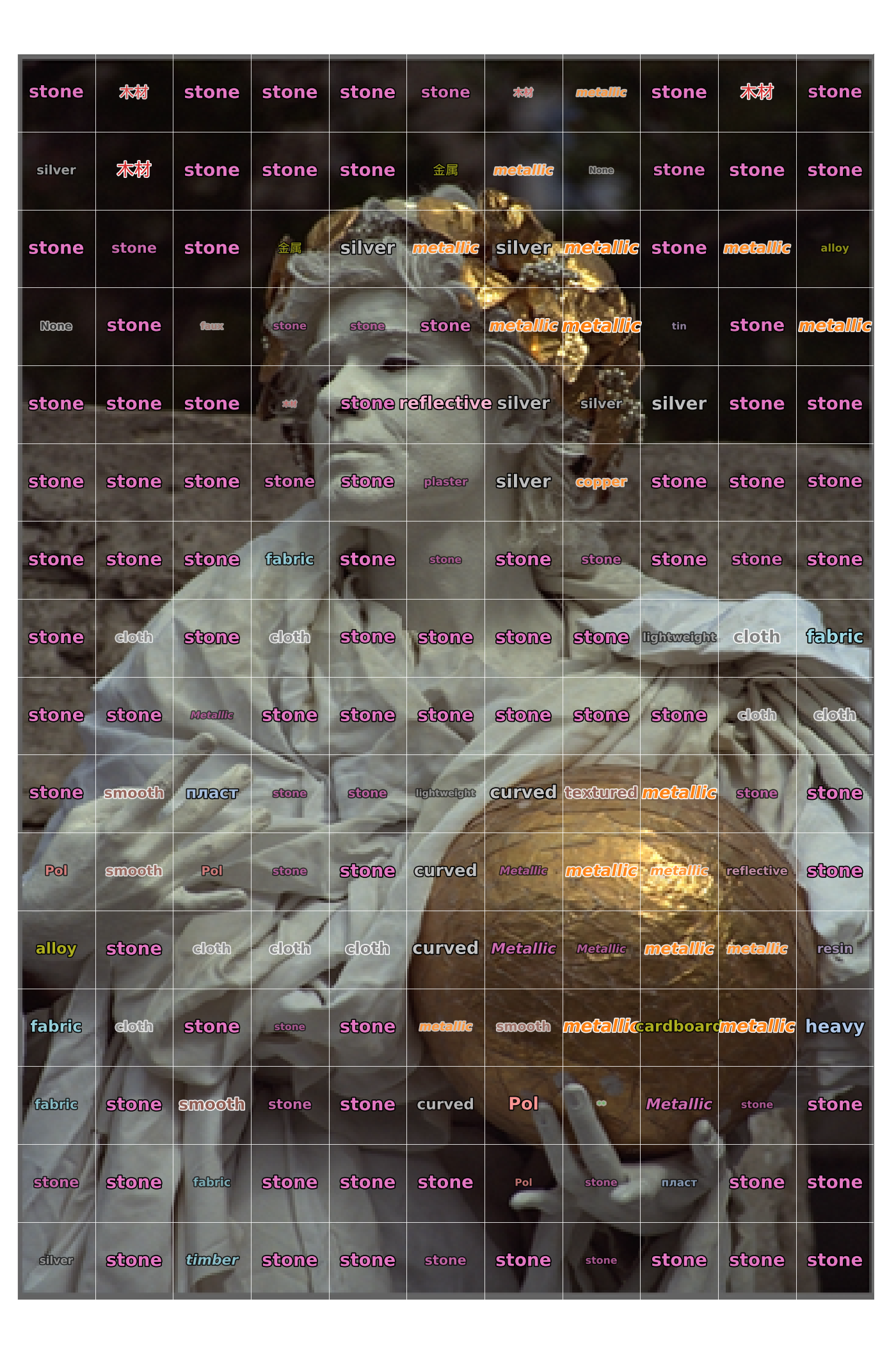}\hfill
    \includegraphics[height=0.298\linewidth]{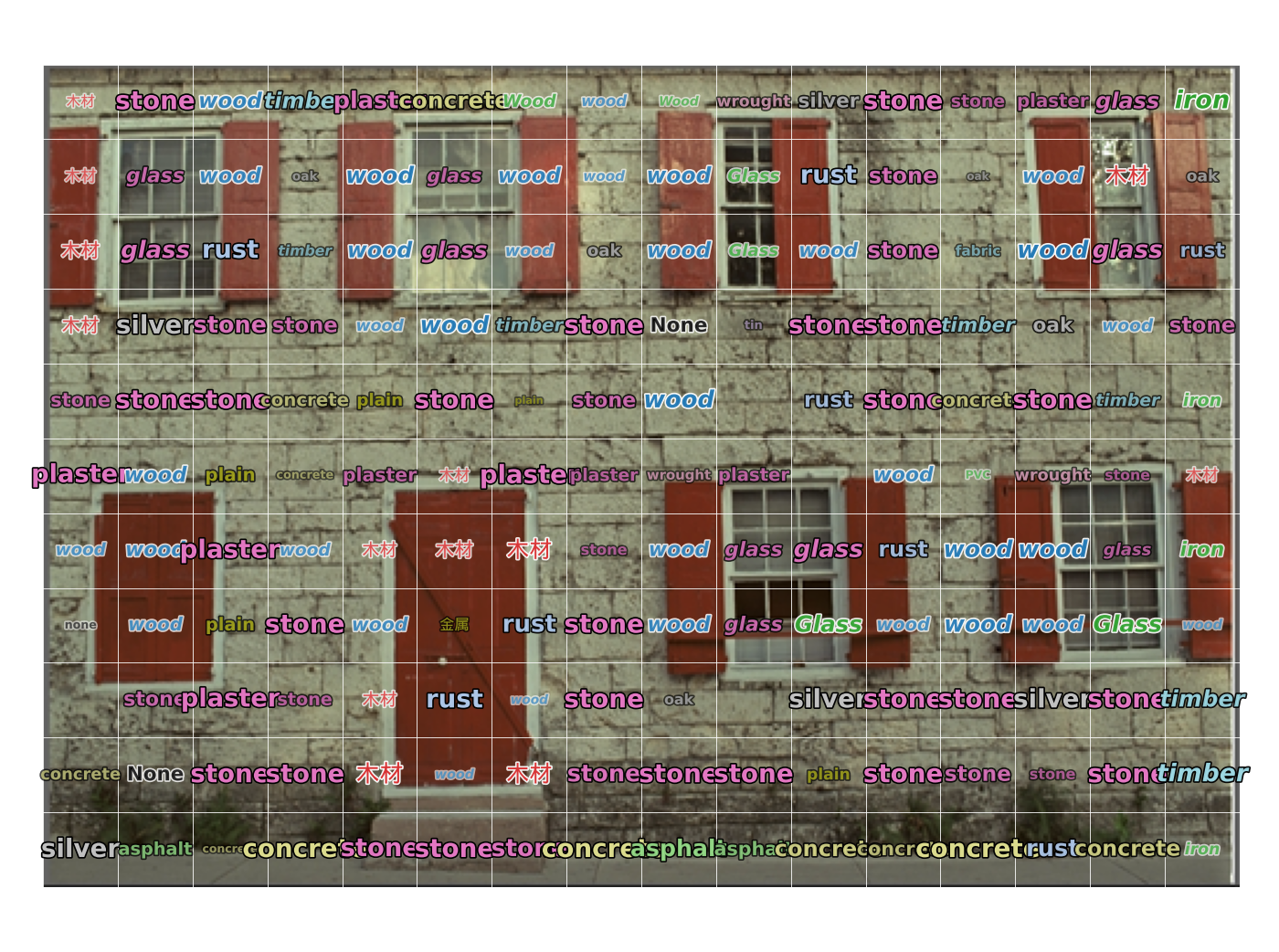}
    \caption{\textbf{Patch-level readouts of materials.}
    Top-1 token prediction of DaP estimated with \protect\prompt{Which word best describes the material of the \protect\ph{object name}?} on each visual token, as in \cref{fig:supp_readout_object}.}
    \label{fig:supp_readout_material}
\end{figure*}

\begin{figure*}[p]
    \centering
    \includegraphics[height=0.364\linewidth]{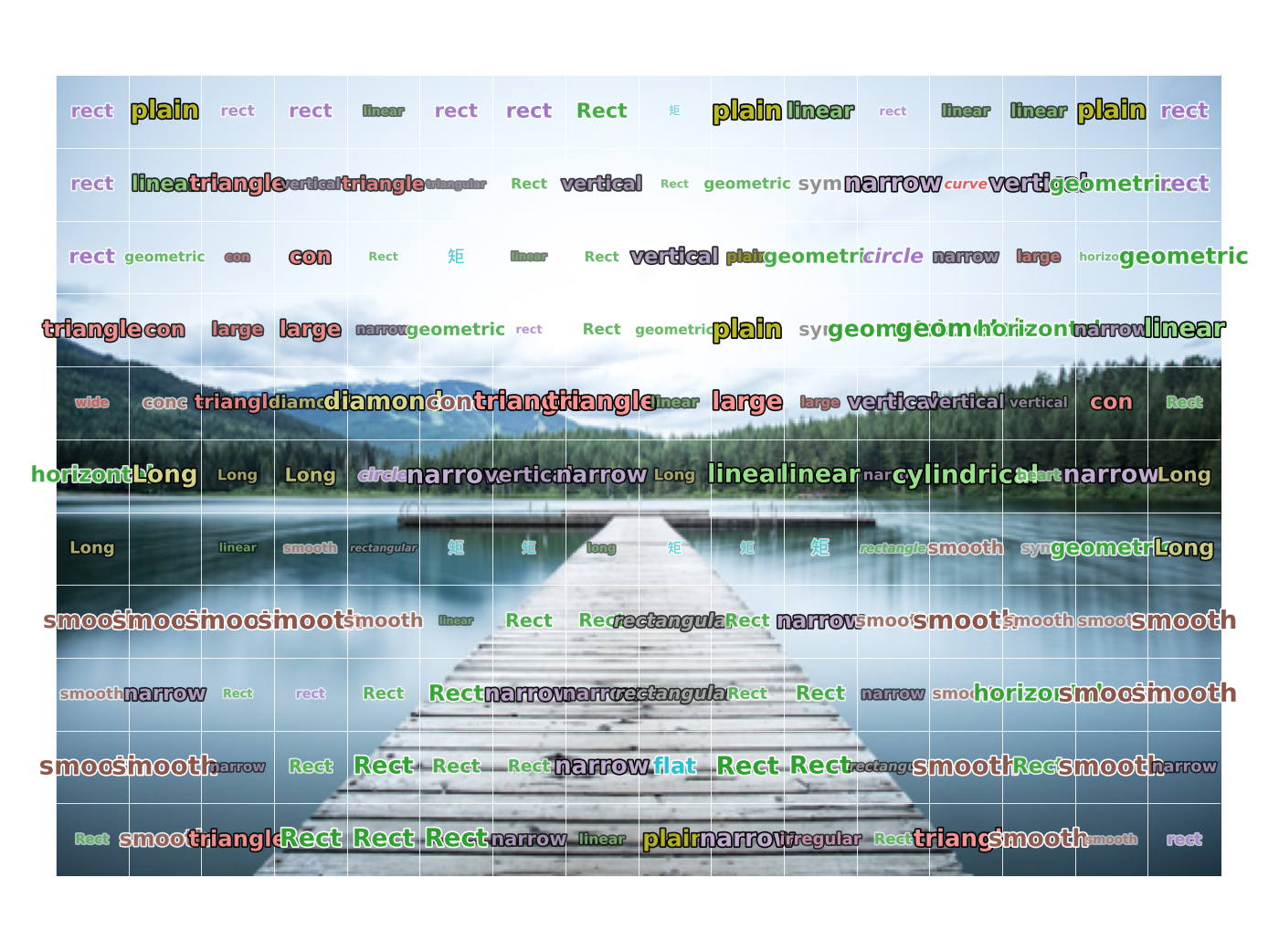}\hfill
    \includegraphics[height=0.364\linewidth]{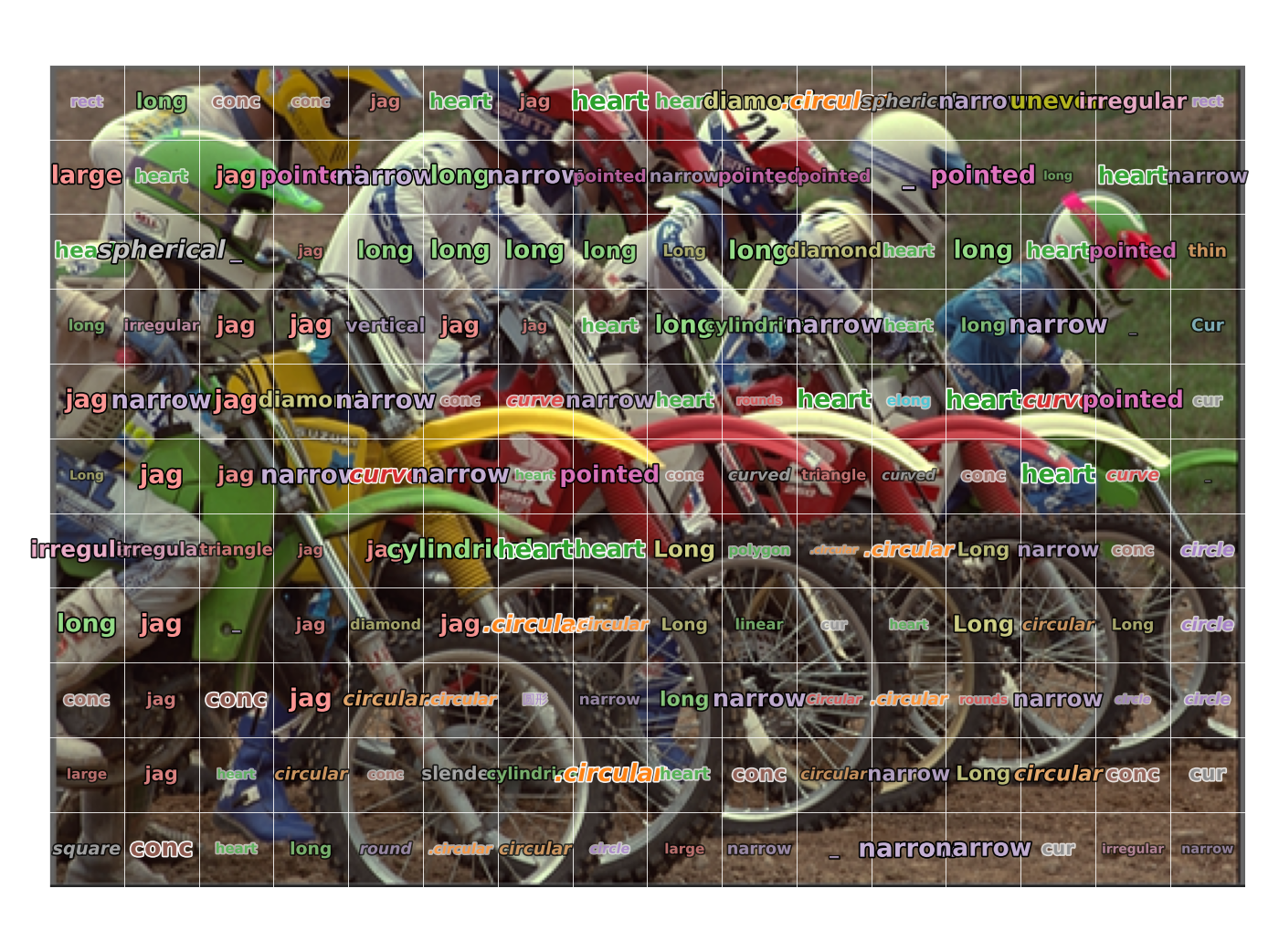}\\[4pt]
    \includegraphics[height=0.289\linewidth]{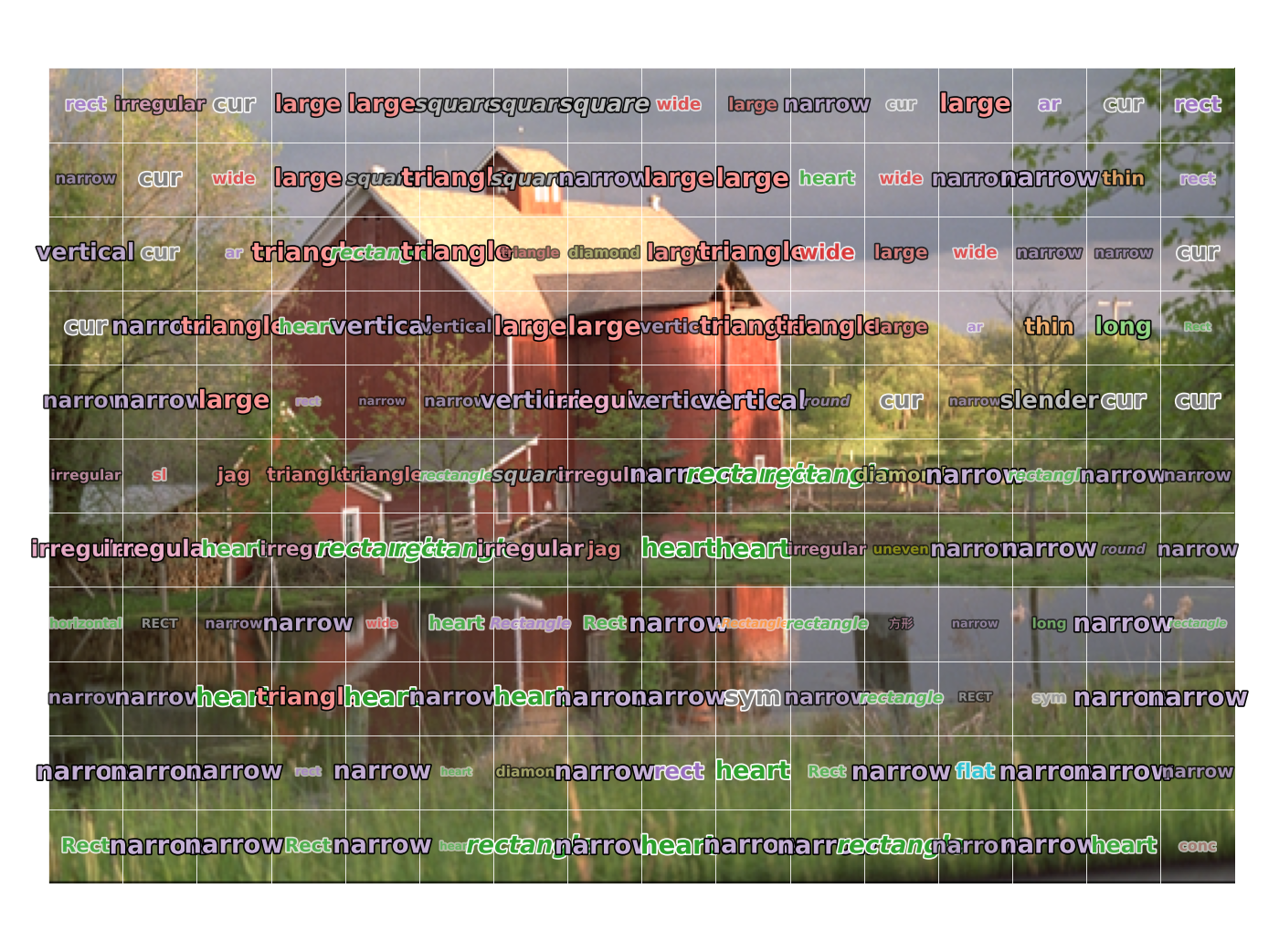}\hfill
    \includegraphics[height=0.289\linewidth]{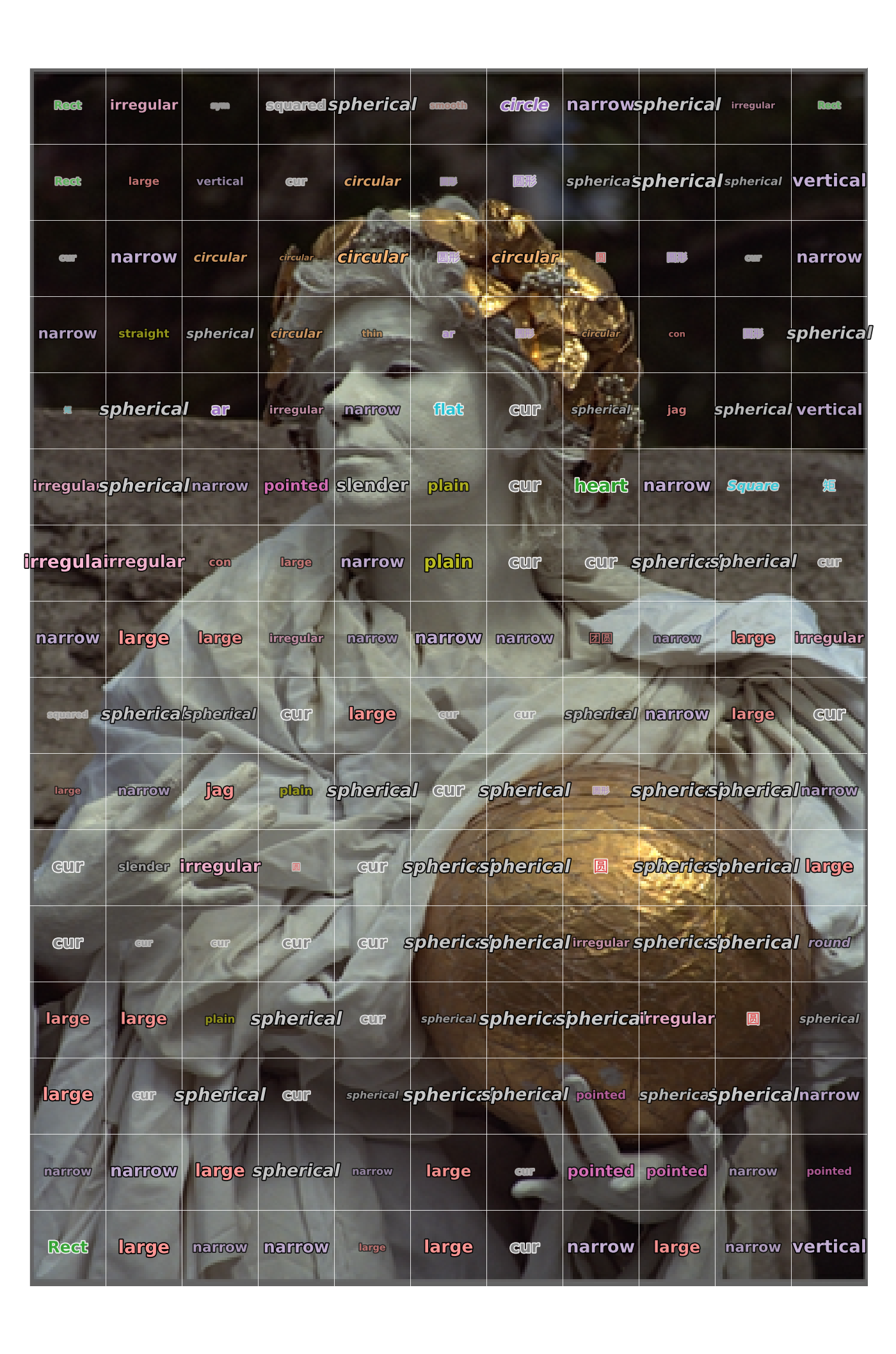}\hfill
    \includegraphics[height=0.289\linewidth]{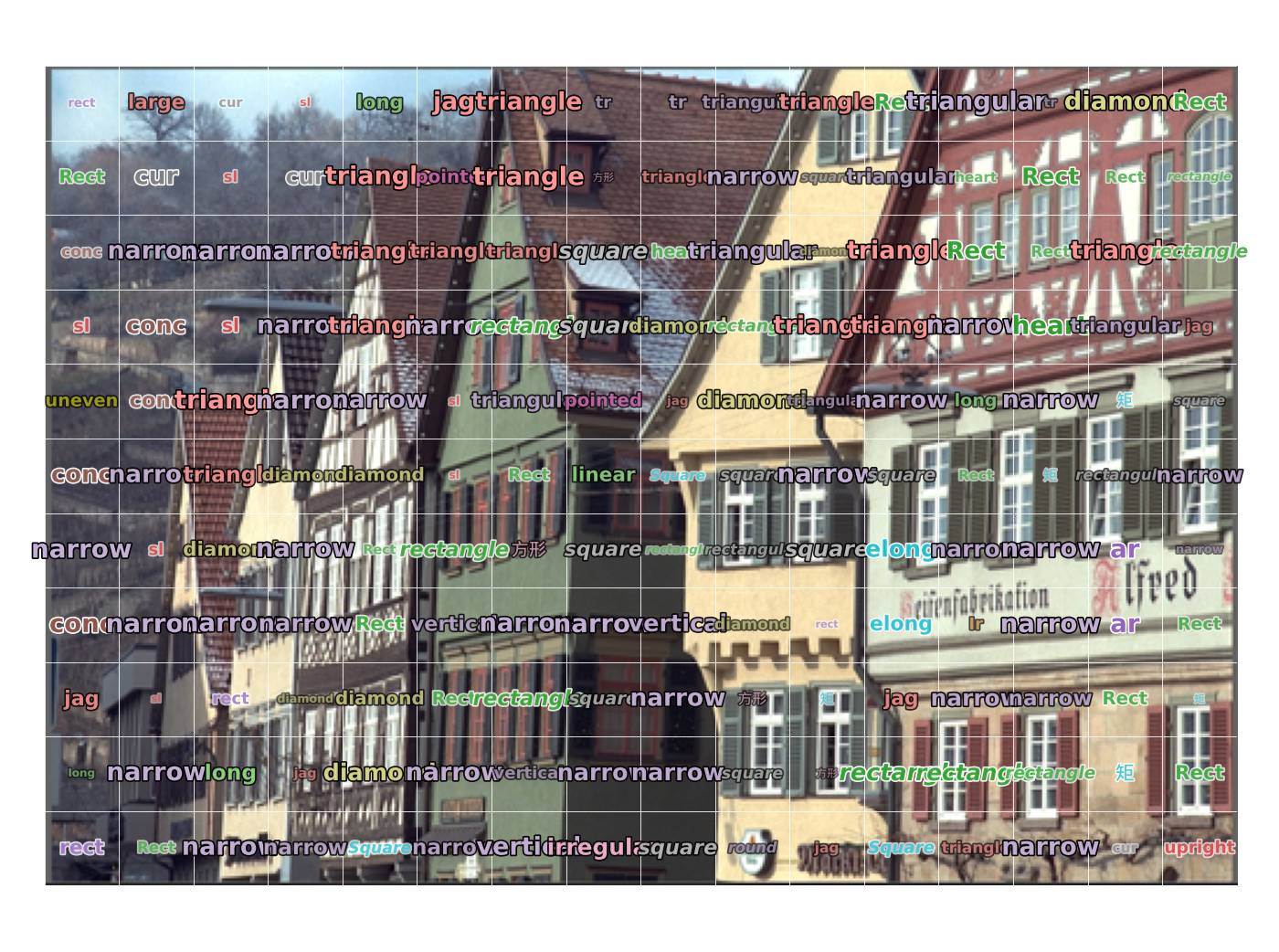}
    \caption{\textbf{Patch-level readouts of shapes.}
    Top-1 token prediction of DaP estimated with \protect\prompt{Which word best describes the shape of the \protect\ph{object name}?} on each visual token, as in \cref{fig:supp_readout_object}.}
    \label{fig:supp_readout_shape}
\end{figure*}

\Cref{fig:supp_readout_object,fig:supp_readout_color,fig:supp_readout_material,fig:supp_readout_shape} show more patch-level readouts.
As in \cref{fig:patch_readout}, DaP decodes each visual token separately ($\mathcal{M} = \{i\}$), here on every visual token of the full image.
For Object Names, the gate keeps $k = 5000$ tokens, a set large enough to also contain color words. To get better qualitative results, we remove the tokens of the color gate from it.
Word size and opacity are scaled between the lowest and the highest probability of each image.
DaP gives an answer at every visual token, including tokens where the concept does not apply, such as a shape for the sky.
The images come from the Kodak set, COCO and the LLaVA and Qwen-VL demos, while DaP is estimated on VAW training instances.

\subsection{Steering examples} \label{sec:app:steered_examples}

\Cref{fig:supp_steer_types,fig:supp_steer_attributes} show more examples of steering with DaP, following \cref{eq:steer}.
The images come from the Kodak set, COCO and the LLaVA and Qwen-VL demos, while DaP is estimated on VAW training instances.
For each example, the mask region $\mathcal{M}$ is a set of visual patches selected by hand.
Each question ends with \prompt{Answer in one sentence.}
We test several steering strengths per example and report the smallest $\alpha$ that changes the answer.

\newcommand{\steercard}[1]{\begin{minipage}[t]{0.49\linewidth}\vspace{0pt}\includegraphics[width=\linewidth]{assets/supp/steer/#1.pdf}\end{minipage}}

\begin{figure*}[p]
    \centering
    \steercard{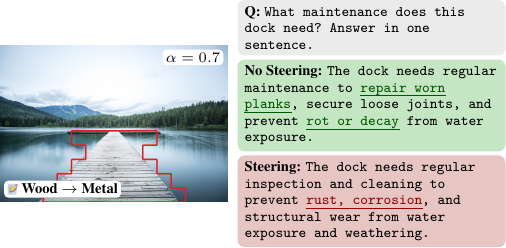}\hfill
    \steercard{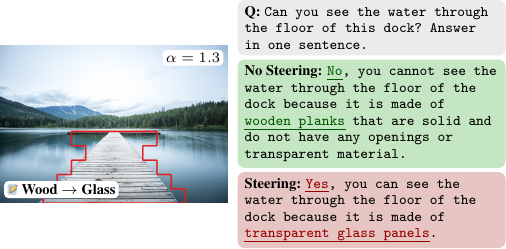}\\[8pt]
    \steercard{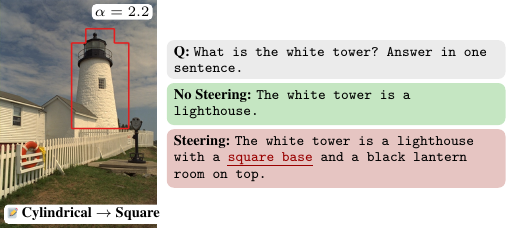}\hfill
    \steercard{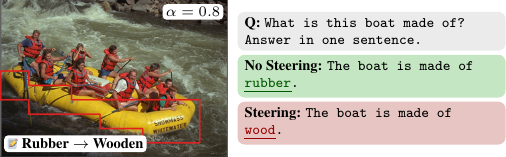}\\[8pt]
    \steercard{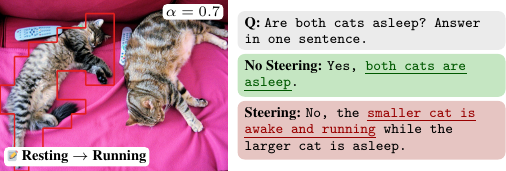}\hfill
    \steercard{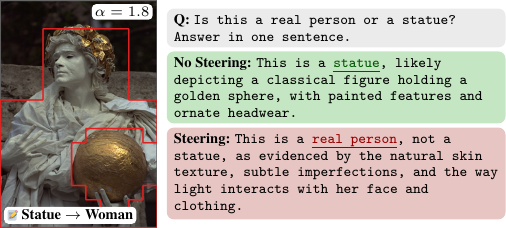}\\[8pt]
    \steercard{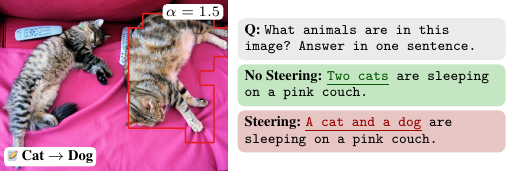}\hfill
    \steercard{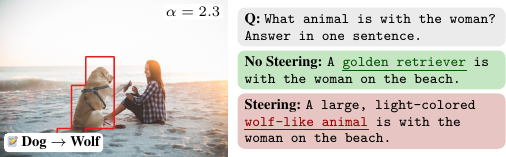}
    \caption{\textbf{Steering Attribute Types and Object Names with DaP.}
    As in \cref{fig:steer}, \nosteerhl{green} marks the answer of the model without steering, and \steerhl{red} the answer of the steered model.
    The red outline marks the visual patches where the steering is performed.
    Each edit applies \cref{eq:steer} from the concept that DaP reads on these patches to the target concept, with the steering strength $\alpha$ given on each image.
    Underlined words change with the steering.}
    \label{fig:supp_steer_types}
\end{figure*}

\begin{figure*}[p]
    \centering
    \steercard{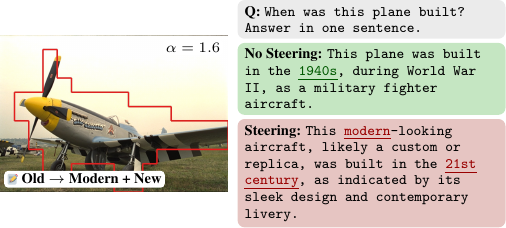}\hfill
    \steercard{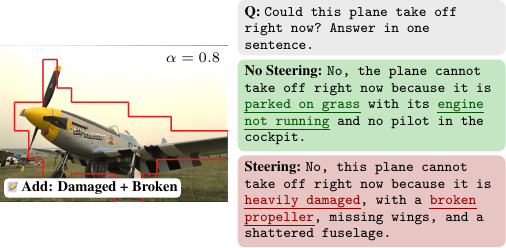}\\[8pt]
    \steercard{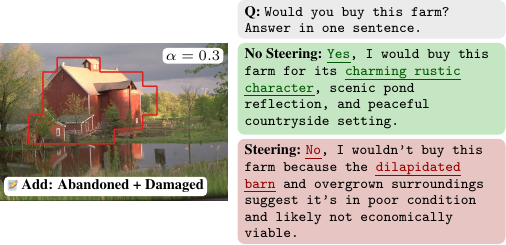}\hfill
    \steercard{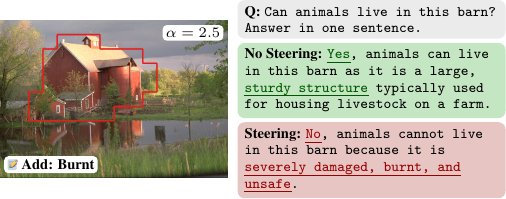}\\[8pt]
    \steercard{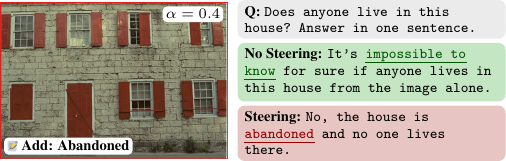}\hfill
    \steercard{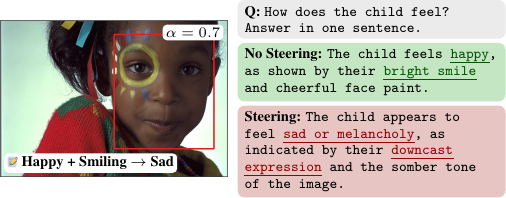}\\[8pt]
    \steercard{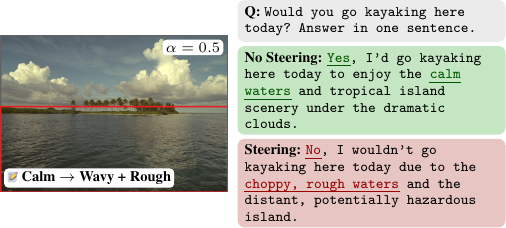}\hfill
    \steercard{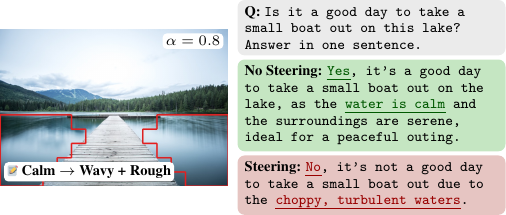}
    \caption{\textbf{Steering All Attributes with DaP.}
    Same layout as \cref{fig:supp_steer_types}.
    Each attribute after \textit{Add} or right of $\rightarrow$ is steered toward \protect\tok{yes}, and each attribute left of $\rightarrow$ toward \protect\tok{no}.
    The unit steering directions $\delta / \lVert \delta \rVert$ of several attributes are summed.}
    \label{fig:supp_steer_attributes}
\end{figure*}

\clearpage

\subsection{Transferability} \label{sec:app:transfer}
In \cref{sec:method}, we noticed that the bias plays the role of the attribute selector.
In that case, an interesting question is whether the Jacobian estimate is the same for all attribute types and object names.
If so, we would need only one Jacobian estimate for a model, which gives us the universal mapping from the input visual tokens to the unembedding matrix.
This is corroborated by the fact that the Jacobians estimated by DaP have a relatively high rank (their participation ratio $(\sum_i \sigma_i^2)^2 / \sum_i \sigma_i^4$ is 79, 107, 12, 100 and 98 for color, material, shape, activity and object names, respectively), so unrelated attributes are not filtered out by the Jacobian, explaining why some of them come up as answers.

\begin{figure*}[t]
    \centering
    \begin{minipage}[t]{0.49\textwidth}\centering
        \includegraphics[width=\linewidth]{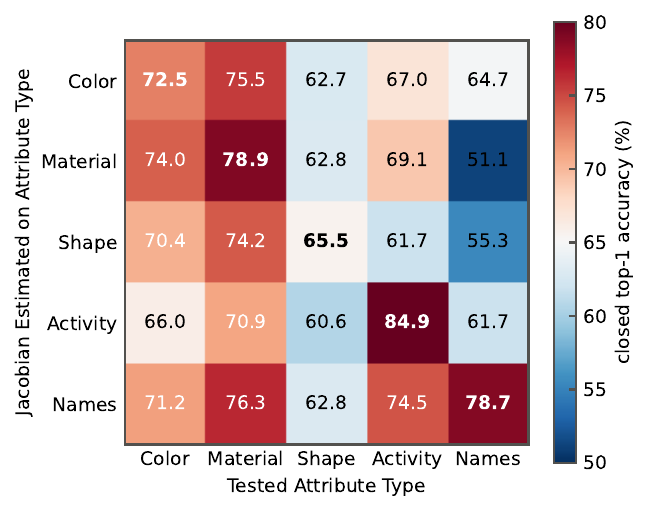}
    \end{minipage}\hfill
    \begin{minipage}[t]{0.49\textwidth}\centering
        \includegraphics[width=\linewidth]{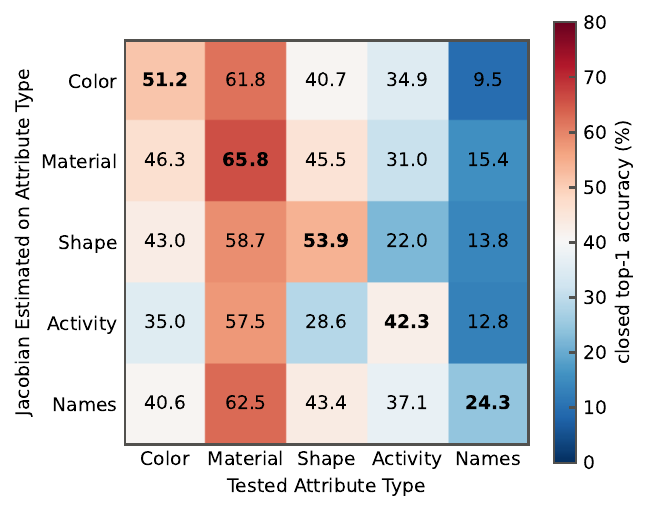}
    \end{minipage}\\[6pt]
    \begin{minipage}[t]{0.49\textwidth}\centering
        \includegraphics[width=\linewidth]{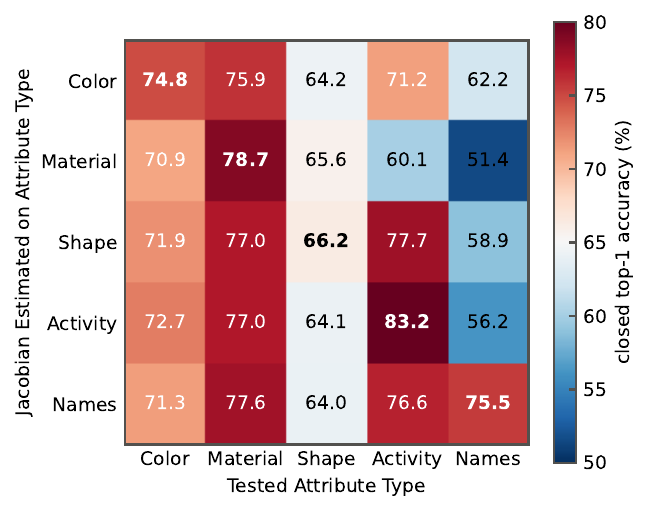}
    \end{minipage}\hfill
    \begin{minipage}[t]{0.49\textwidth}\centering
        \includegraphics[width=\linewidth]{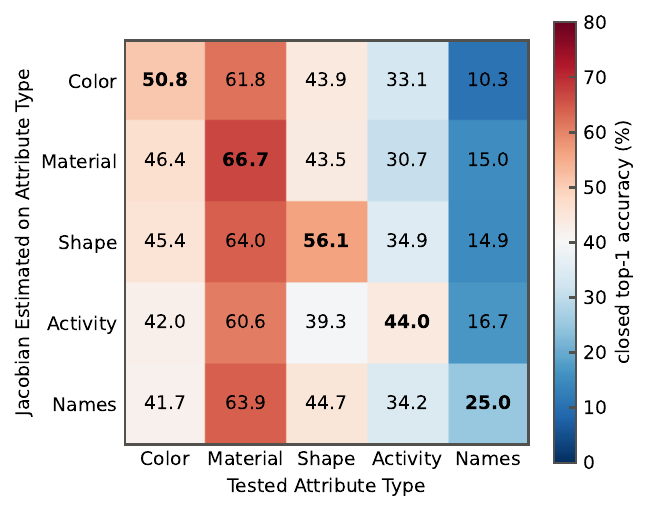}
    \end{minipage}
    \caption{\textbf{Transfer of $J_\pi$ across attribute types.}
    Closed top-1 accuracy (\%) of DaP on VAW.
    Each row estimates $J_\pi$ with the prompt of one attribute type, and each column tests it on another attribute type.
    Top, each row estimates $J_\pi$ on 16 shots of each seen concept of its attribute type.
    Bottom, every row estimates $J_\pi$ on the same 64 object instances drawn at random from the VAW training split, so the rows differ only by the prompt.
    Left, the test covers the 4 seen concepts of the tested attribute type.
    Right, it covers the unseen concepts of the tested attribute type.}
    \label{fig:cross_domain}
\end{figure*}

To test that, we measure the cross-domain transfer of each Jacobian (\cref{fig:cross_domain}).
Since the bias plays the role of the attribute selector, restricting the answers to the concepts of the tested domain replaces it, so the accuracy only measures how well the Jacobian separates the concepts of the tested domain.
The results show that the Jacobian is partly universal and partly specific to its domain.
On the one hand, every Jacobian transfers to every other domain well above chance.
On the other hand, activity and object names lose up to 28 points with the Jacobian of another domain, and on unseen concepts, the Jacobian of the tested domain gives the best accuracy in every domain.
More interestingly, when every Jacobian is estimated on the same object instances and only the prompt differs (bottom of \cref{fig:cross_domain}), we still see a gap between domains, which shows that the prompt plays an important role in the estimation of DaP.

\refstepcounter{section}\label{sec:app:comparison_with_other_methdos}

\section{Other LVLMs} \label{sec:app:models}
\Cref{tab:models_acc,tab:models_faith,tab:models_steer} report the accuracy, the faithfulness and the steering success of each linear probe on Qwen3-VL-8B, InternVL3.5-8B, Idefics3-8B and Ministral 3 8B.

\begin{table*}
\centering
\small\renewcommand{\arraystretch}{0.95}\setlength{\tabcolsep}{4pt}
\begin{tabular}{llc|c|cccc|c}
\toprule
 & Method & Shots & All Attributes & Color & Material & Shape & Activity & Names \\
\midrule
 & \# unseen concepts & & -- & 27 & 10 & 17 & 21 & 1313 \\
\midrule[\heavyrulewidth]
\multicolumn{9}{c}{\textbf{Qwen3-VL-8B}} \\
\midrule
 & NCC~\cite{panickssery2024caa} & 16 & 68.4 & 56.6 & 64.2 & 44.2 & 70.7 & \underline{84.7} \\
 & Probe~\cite{savietto2026geometry} & 16 & \underline{69.7} & 62.3 & 67.3 & 50.3 & \underline{76.4} & \textbf{85.7} \\
 & Emb-Lens~\cite{fan2026visualtokens} & 0 & -- & 38.3 & 49.7 & 47.7 & 53.1 & 35.3 \\
 & \cellcolor{gray!15}DaP & \cellcolor{gray!15}1 & \cellcolor{gray!15}62.1 & \cellcolor{gray!15}\underline{65.6} & \cellcolor{gray!15}\underline{68.0} & \cellcolor{gray!15}\underline{60.9} & \cellcolor{gray!15}58.9 & \cellcolor{gray!15}42.5 \\
 & \cellcolor{gray!15}DaP & \cellcolor{gray!15}16 & \cellcolor{gray!15}\textbf{72.7} & \cellcolor{gray!15}\textbf{73.8} & \cellcolor{gray!15}\textbf{77.6} & \cellcolor{gray!15}\textbf{65.2} & \cellcolor{gray!15}\textbf{81.0} & \cellcolor{gray!15}76.0 \\
\cmidrule(l){2-9}
\multirow{-6}{*}{\rotatebox[origin=c]{90}{\shortstack{\textit{Seen (4-way)}}}} & \textit{Model} & \textit{0} & \textit{82.4} & \textit{89.8} & \textit{89.5} & \textit{76.6} & \textit{91.1} & \textit{82.9} \\
\midrule\midrule
 & Emb-Lens~\cite{fan2026visualtokens} & 0 & -- & 8.7 & 37.8 & 17.8 & 14.0 & 2.9 \\
 & \cellcolor{gray!15}DaP & \cellcolor{gray!15}1 & \cellcolor{gray!15}-- & \cellcolor{gray!15}\underline{33.7} & \cellcolor{gray!15}\underline{45.2} & \cellcolor{gray!15}\underline{43.9} & \cellcolor{gray!15}\underline{21.8} & \cellcolor{gray!15}\underline{6.2} \\
 & \cellcolor{gray!15}DaP & \cellcolor{gray!15}16 & \cellcolor{gray!15}-- & \cellcolor{gray!15}\textbf{53.0} & \cellcolor{gray!15}\textbf{58.2} & \cellcolor{gray!15}\textbf{57.7} & \cellcolor{gray!15}\textbf{42.0} & \cellcolor{gray!15}\textbf{24.1} \\
\cmidrule(l){2-9}
\multirow{-4}{*}{\rotatebox[origin=c]{90}{\shortstack{\textit{Unseen}}}} & \textit{Model} & \textit{0} & \textit{--} & \textit{74.3} & \textit{77.5} & \textit{58.4} & \textit{65.5} & \textit{27.7} \\
\midrule[\heavyrulewidth]
\multicolumn{9}{c}{\textbf{InternVL3.5-8B}} \\
\midrule
 & NCC~\cite{panickssery2024caa} & 16 & 66.9 & 54.8 & 60.7 & 48.6 & 71.6 & \underline{87.6} \\
 & Probe~\cite{savietto2026geometry} & 16 & \underline{68.9} & 62.7 & \underline{63.3} & 49.4 & \textbf{74.5} & \textbf{90.3} \\
 & Emb-Lens~\cite{fan2026visualtokens} & 0 & -- & 32.5 & 45.8 & 41.7 & 22.5 & 25.8 \\
 & \cellcolor{gray!15}DaP & \cellcolor{gray!15}1 & \cellcolor{gray!15}62.0 & \cellcolor{gray!15}\underline{68.8} & \cellcolor{gray!15}58.6 & \cellcolor{gray!15}\textbf{63.5} & \cellcolor{gray!15}63.7 & \cellcolor{gray!15}35.2 \\
 & \cellcolor{gray!15}DaP & \cellcolor{gray!15}16 & \cellcolor{gray!15}\textbf{71.1} & \cellcolor{gray!15}\textbf{74.2} & \cellcolor{gray!15}\textbf{74.2} & \cellcolor{gray!15}\underline{62.8} & \cellcolor{gray!15}\underline{73.8} & \cellcolor{gray!15}78.2 \\
\cmidrule(l){2-9}
\multirow{-6}{*}{\rotatebox[origin=c]{90}{\shortstack{\textit{Seen (4-way)}}}} & \textit{Model} & \textit{0} & \textit{75.1} & \textit{88.7} & \textit{86.9} & \textit{69.6} & \textit{90.4} & \textit{83.2} \\
\midrule\midrule
 & Emb-Lens~\cite{fan2026visualtokens} & 0 & -- & 5.6 & 14.7 & 8.8 & 10.9 & 1.4 \\
 & \cellcolor{gray!15}DaP & \cellcolor{gray!15}1 & \cellcolor{gray!15}-- & \cellcolor{gray!15}\underline{27.3} & \cellcolor{gray!15}\underline{42.2} & \cellcolor{gray!15}\underline{34.7} & \cellcolor{gray!15}\underline{20.4} & \cellcolor{gray!15}\underline{3.3} \\
 & \cellcolor{gray!15}DaP & \cellcolor{gray!15}16 & \cellcolor{gray!15}-- & \cellcolor{gray!15}\textbf{42.3} & \cellcolor{gray!15}\textbf{59.3} & \cellcolor{gray!15}\textbf{43.8} & \cellcolor{gray!15}\textbf{32.1} & \cellcolor{gray!15}\textbf{17.7} \\
\cmidrule(l){2-9}
\multirow{-4}{*}{\rotatebox[origin=c]{90}{\shortstack{\textit{Unseen}}}} & \textit{Model} & \textit{0} & \textit{--} & \textit{67.1} & \textit{76.7} & \textit{58.3} & \textit{64.0} & \textit{22.7} \\
\midrule[\heavyrulewidth]
\multicolumn{9}{c}{\textbf{Idefics3-8B}} \\
\midrule
 & NCC~\cite{panickssery2024caa} & 16 & 67.3 & 49.7 & 58.2 & 40.3 & \textbf{73.5} & \textbf{86.3} \\
 & Probe~\cite{savietto2026geometry} & 16 & \underline{67.5} & 49.6 & 59.5 & 43.6 & \underline{73.1} & \underline{85.6} \\
 & Emb-Lens~\cite{fan2026visualtokens} & 0 & -- & 31.2 & 21.8 & 43.6 & 39.7 & 47.2 \\
 & \cellcolor{gray!15}DaP & \cellcolor{gray!15}1 & \cellcolor{gray!15}59.8 & \cellcolor{gray!15}\underline{50.4} & \cellcolor{gray!15}\underline{63.2} & \cellcolor{gray!15}\underline{52.5} & \cellcolor{gray!15}59.4 & \cellcolor{gray!15}52.4 \\
 & \cellcolor{gray!15}DaP & \cellcolor{gray!15}16 & \cellcolor{gray!15}\textbf{69.8} & \cellcolor{gray!15}\textbf{62.0} & \cellcolor{gray!15}\textbf{75.8} & \cellcolor{gray!15}\textbf{58.1} & \cellcolor{gray!15}71.4 & \cellcolor{gray!15}68.0 \\
\cmidrule(l){2-9}
\multirow{-6}{*}{\rotatebox[origin=c]{90}{\shortstack{\textit{Seen (4-way)}}}} & \textit{Model} & \textit{0} & \textit{79.5} & \textit{86.4} & \textit{84.8} & \textit{67.7} & \textit{88.3} & \textit{65.3} \\
\midrule\midrule
 & Emb-Lens~\cite{fan2026visualtokens} & 0 & -- & 7.1 & 10.4 & 6.4 & 14.7 & 0.3 \\
 & \cellcolor{gray!15}DaP & \cellcolor{gray!15}1 & \cellcolor{gray!15}-- & \cellcolor{gray!15}\underline{20.7} & \cellcolor{gray!15}\underline{42.0} & \cellcolor{gray!15}\underline{15.8} & \cellcolor{gray!15}\underline{15.8} & \cellcolor{gray!15}\underline{2.7} \\
 & \cellcolor{gray!15}DaP & \cellcolor{gray!15}16 & \cellcolor{gray!15}-- & \cellcolor{gray!15}\textbf{32.8} & \cellcolor{gray!15}\textbf{53.1} & \cellcolor{gray!15}\textbf{26.5} & \cellcolor{gray!15}\textbf{32.9} & \cellcolor{gray!15}\textbf{15.2} \\
\cmidrule(l){2-9}
\multirow{-4}{*}{\rotatebox[origin=c]{90}{\shortstack{\textit{Unseen}}}} & \textit{Model} & \textit{0} & \textit{--} & \textit{68.7} & \textit{75.6} & \textit{53.0} & \textit{61.9} & \textit{20.4} \\
\midrule[\heavyrulewidth]
\multicolumn{9}{c}{\textbf{Ministral 3 8B}} \\
\midrule
 & NCC~\cite{panickssery2024caa} & 16 & \textbf{66.8} & \underline{45.1} & \underline{53.9} & 44.0 & \textbf{71.2} & \textbf{74.6} \\
 & Probe~\cite{savietto2026geometry} & 16 & \underline{65.9} & 44.2 & 53.8 & 41.6 & \underline{69.5} & \underline{70.6} \\
 & Emb-Lens~\cite{fan2026visualtokens} & 0 & -- & 28.4 & 36.6 & \underline{45.4} & 29.8 & 23.3 \\
 & \cellcolor{gray!15}DaP & \cellcolor{gray!15}1 & \cellcolor{gray!15}57.0 & \cellcolor{gray!15}43.0 & \cellcolor{gray!15}34.2 & \cellcolor{gray!15}44.2 & \cellcolor{gray!15}58.7 & \cellcolor{gray!15}44.0 \\
 & \cellcolor{gray!15}DaP & \cellcolor{gray!15}16 & \cellcolor{gray!15}65.4 & \cellcolor{gray!15}\textbf{63.9} & \cellcolor{gray!15}\textbf{64.1} & \cellcolor{gray!15}\textbf{53.2} & \cellcolor{gray!15}46.0 & \cellcolor{gray!15}51.2 \\
\cmidrule(l){2-9}
\multirow{-6}{*}{\rotatebox[origin=c]{90}{\shortstack{\textit{Seen (4-way)}}}} & \textit{Model} & \textit{0} & \textit{72.8} & \textit{83.0} & \textit{82.1} & \textit{67.7} & \textit{82.8} & \textit{75.9} \\
\midrule\midrule
 & Emb-Lens~\cite{fan2026visualtokens} & 0 & -- & 2.9 & 19.5 & \underline{14.9} & 5.5 & 0.2 \\
 & \cellcolor{gray!15}DaP & \cellcolor{gray!15}1 & \cellcolor{gray!15}-- & \cellcolor{gray!15}\underline{15.6} & \cellcolor{gray!15}\underline{23.6} & \cellcolor{gray!15}14.0 & \cellcolor{gray!15}\underline{10.5} & \cellcolor{gray!15}\underline{0.5} \\
 & \cellcolor{gray!15}DaP & \cellcolor{gray!15}16 & \cellcolor{gray!15}-- & \cellcolor{gray!15}\textbf{25.8} & \cellcolor{gray!15}\textbf{40.9} & \cellcolor{gray!15}\textbf{27.1} & \cellcolor{gray!15}\textbf{20.4} & \cellcolor{gray!15}\textbf{5.7} \\
\cmidrule(l){2-9}
\multirow{-4}{*}{\rotatebox[origin=c]{90}{\shortstack{\textit{Unseen}}}} & \textit{Model} & \textit{0} & \textit{--} & \textit{49.7} & \textit{69.5} & \textit{54.4} & \textit{59.3} & \textit{19.4} \\
\bottomrule
\end{tabular}
\caption{\textbf{Linear probe accuracy (\%) of other LVLMs}. Rows and columns as in \cref{tab:benchmark}. NCC and Probe use the configurations chosen on the validation set for Qwen3-VL-4B. \textbf{Bold}: best, \underline{underlined}: second best.}
\label{tab:models_acc}
\end{table*}

\begin{table*}
\centering
\small\renewcommand{\arraystretch}{0.95}\setlength{\tabcolsep}{4pt}
\begin{tabular}{llc|c|cccc|c}
\toprule
 & Method & Shots & All Attributes & Color & Material & Shape & Activity & Names \\
\midrule
 & \# unseen concepts & & -- & 27 & 10 & 17 & 21 & 1313 \\
\midrule[\heavyrulewidth]
\multicolumn{9}{c}{\textbf{Qwen3-VL-8B}} \\
\midrule
 & NCC~\cite{panickssery2024caa} & 16 & 67.5 & 55.2 & 63.6 & 39.1 & 72.0 & \textbf{78.7} \\
 & Probe~\cite{savietto2026geometry} & 16 & \underline{68.7} & 61.4 & 67.0 & 47.7 & \underline{77.7} & \underline{77.5} \\
 & Emb-Lens~\cite{fan2026visualtokens} & 0 & -- & 38.7 & 50.1 & 56.2 & 54.4 & 29.5 \\
 & \cellcolor{gray!15}DaP & \cellcolor{gray!15}1 & \cellcolor{gray!15}63.9 & \cellcolor{gray!15}\underline{65.3} & \cellcolor{gray!15}\underline{69.1} & \cellcolor{gray!15}\underline{66.1} & \cellcolor{gray!15}58.8 & \cellcolor{gray!15}39.4 \\
 & \cellcolor{gray!15}DaP & \cellcolor{gray!15}16 & \cellcolor{gray!15}\textbf{74.0} & \cellcolor{gray!15}\textbf{74.3} & \cellcolor{gray!15}\textbf{78.3} & \cellcolor{gray!15}\textbf{72.8} & \cellcolor{gray!15}\textbf{82.8} & \cellcolor{gray!15}72.3 \\
\cmidrule(l){2-9}
\multirow{-6}{*}{\rotatebox[origin=c]{90}{\shortstack{\textit{Seen (4-way)}}}} & \textit{Chance} & \textit{--} & \textit{60.1} & \textit{27.8} & \textit{37.6} & \textit{54.4} & \textit{47.3} & \textit{61.5} \\
\midrule\midrule
 & Emb-Lens~\cite{fan2026visualtokens} & 0 & -- & 8.1 & 34.6 & 10.7 & 12.8 & 3.2 \\
 & \cellcolor{gray!15}DaP & \cellcolor{gray!15}1 & \cellcolor{gray!15}-- & \cellcolor{gray!15}\underline{35.7} & \cellcolor{gray!15}\underline{50.9} & \cellcolor{gray!15}\underline{44.8} & \cellcolor{gray!15}\underline{27.6} & \cellcolor{gray!15}\underline{9.2} \\
 & \cellcolor{gray!15}DaP & \cellcolor{gray!15}16 & \cellcolor{gray!15}-- & \cellcolor{gray!15}\textbf{57.0} & \cellcolor{gray!15}\textbf{63.5} & \cellcolor{gray!15}\textbf{52.7} & \cellcolor{gray!15}\textbf{47.6} & \cellcolor{gray!15}\textbf{37.5} \\
\cmidrule(l){2-9}
\multirow{-4}{*}{\rotatebox[origin=c]{90}{\shortstack{\textit{Unseen}}}} & \textit{Chance} & \textit{--} & \textit{--} & \textit{24.4} & \textit{23.6} & \textit{17.6} & \textit{13.9} & \textit{3.6} \\
\midrule[\heavyrulewidth]
\multicolumn{9}{c}{\textbf{InternVL3.5-8B}} \\
\midrule
 & NCC~\cite{panickssery2024caa} & 16 & 65.3 & 53.7 & 60.1 & 43.4 & 68.5 & \underline{78.8} \\
 & Probe~\cite{savietto2026geometry} & 16 & \underline{67.1} & 61.6 & \underline{61.3} & 45.7 & \underline{72.7} & \textbf{79.6} \\
 & Emb-Lens~\cite{fan2026visualtokens} & 0 & -- & 31.1 & 43.3 & 35.8 & 22.5 & 22.7 \\
 & \cellcolor{gray!15}DaP & \cellcolor{gray!15}1 & \cellcolor{gray!15}62.3 & \cellcolor{gray!15}\underline{67.4} & \cellcolor{gray!15}56.1 & \cellcolor{gray!15}\underline{66.1} & \cellcolor{gray!15}65.7 & \cellcolor{gray!15}35.7 \\
 & \cellcolor{gray!15}DaP & \cellcolor{gray!15}16 & \cellcolor{gray!15}\textbf{69.2} & \cellcolor{gray!15}\textbf{73.5} & \cellcolor{gray!15}\textbf{73.9} & \cellcolor{gray!15}\textbf{68.2} & \cellcolor{gray!15}\textbf{76.1} & \cellcolor{gray!15}73.2 \\
\cmidrule(l){2-9}
\multirow{-6}{*}{\rotatebox[origin=c]{90}{\shortstack{\textit{Seen (4-way)}}}} & \textit{Chance} & \textit{--} & \textit{70.4} & \textit{28.1} & \textit{36.0} & \textit{48.8} & \textit{53.5} & \textit{51.5} \\
\midrule\midrule
 & Emb-Lens~\cite{fan2026visualtokens} & 0 & -- & 4.9 & 14.9 & 10.4 & 13.7 & 0.7 \\
 & \cellcolor{gray!15}DaP & \cellcolor{gray!15}1 & \cellcolor{gray!15}-- & \cellcolor{gray!15}\underline{27.5} & \cellcolor{gray!15}\underline{41.8} & \cellcolor{gray!15}\underline{36.7} & \cellcolor{gray!15}\underline{19.8} & \cellcolor{gray!15}\underline{3.4} \\
 & \cellcolor{gray!15}DaP & \cellcolor{gray!15}16 & \cellcolor{gray!15}-- & \cellcolor{gray!15}\textbf{45.2} & \cellcolor{gray!15}\textbf{62.6} & \cellcolor{gray!15}\textbf{40.9} & \cellcolor{gray!15}\textbf{29.2} & \cellcolor{gray!15}\textbf{19.0} \\
\cmidrule(l){2-9}
\multirow{-4}{*}{\rotatebox[origin=c]{90}{\shortstack{\textit{Unseen}}}} & \textit{Chance} & \textit{--} & \textit{--} & \textit{17.7} & \textit{23.2} & \textit{19.0} & \textit{13.7} & \textit{2.3} \\
\midrule[\heavyrulewidth]
\multicolumn{9}{c}{\textbf{Idefics3-8B}} \\
\midrule
 & NCC~\cite{panickssery2024caa} & 16 & 64.8 & 47.9 & 57.6 & 33.0 & \underline{72.8} & \underline{62.2} \\
 & Probe~\cite{savietto2026geometry} & 16 & \underline{65.1} & 47.4 & 58.7 & 36.7 & 72.7 & \textbf{62.9} \\
 & Emb-Lens~\cite{fan2026visualtokens} & 0 & -- & 31.5 & 25.1 & 47.9 & 38.0 & 37.0 \\
 & \cellcolor{gray!15}DaP & \cellcolor{gray!15}1 & \cellcolor{gray!15}60.6 & \cellcolor{gray!15}\underline{48.9} & \cellcolor{gray!15}\underline{60.8} & \cellcolor{gray!15}\underline{57.9} & \cellcolor{gray!15}61.7 & \cellcolor{gray!15}48.5 \\
 & \cellcolor{gray!15}DaP & \cellcolor{gray!15}16 & \cellcolor{gray!15}\textbf{71.8} & \cellcolor{gray!15}\textbf{60.1} & \cellcolor{gray!15}\textbf{73.1} & \cellcolor{gray!15}\textbf{62.8} & \cellcolor{gray!15}\textbf{74.1} & \cellcolor{gray!15}55.3 \\
\cmidrule(l){2-9}
\multirow{-6}{*}{\rotatebox[origin=c]{90}{\shortstack{\textit{Seen (4-way)}}}} & \textit{Chance} & \textit{--} & \textit{59.2} & \textit{30.2} & \textit{34.8} & \textit{55.7} & \textit{50.9} & \textit{34.1} \\
\midrule\midrule
 & Emb-Lens~\cite{fan2026visualtokens} & 0 & -- & 6.6 & 10.6 & 6.2 & 11.2 & 0.6 \\
 & \cellcolor{gray!15}DaP & \cellcolor{gray!15}1 & \cellcolor{gray!15}-- & \cellcolor{gray!15}\underline{22.0} & \cellcolor{gray!15}\underline{40.5} & \cellcolor{gray!15}\underline{14.8} & \cellcolor{gray!15}\underline{14.8} & \cellcolor{gray!15}\underline{2.5} \\
 & \cellcolor{gray!15}DaP & \cellcolor{gray!15}16 & \cellcolor{gray!15}-- & \cellcolor{gray!15}\textbf{33.0} & \cellcolor{gray!15}\textbf{54.4} & \cellcolor{gray!15}\textbf{26.5} & \cellcolor{gray!15}\textbf{38.8} & \cellcolor{gray!15}\textbf{19.6} \\
\cmidrule(l){2-9}
\multirow{-4}{*}{\rotatebox[origin=c]{90}{\shortstack{\textit{Unseen}}}} & \textit{Chance} & \textit{--} & \textit{--} & \textit{28.9} & \textit{24.0} & \textit{21.4} & \textit{17.3} & \textit{3.0} \\
\midrule[\heavyrulewidth]
\multicolumn{9}{c}{\textbf{Ministral 3 8B}} \\
\midrule
 & NCC~\cite{panickssery2024caa} & 16 & \underline{63.5} & 41.4 & \underline{53.1} & 41.1 & \textbf{72.1} & \textbf{62.1} \\
 & Probe~\cite{savietto2026geometry} & 16 & 62.5 & 40.6 & 52.8 & 39.6 & \underline{70.5} & \underline{57.3} \\
 & Emb-Lens~\cite{fan2026visualtokens} & 0 & -- & 31.9 & 36.6 & 39.7 & 31.9 & 27.8 \\
 & \cellcolor{gray!15}DaP & \cellcolor{gray!15}1 & \cellcolor{gray!15}58.6 & \cellcolor{gray!15}\underline{42.3} & \cellcolor{gray!15}33.8 & \cellcolor{gray!15}\underline{42.1} & \cellcolor{gray!15}59.5 & \cellcolor{gray!15}45.5 \\
 & \cellcolor{gray!15}DaP & \cellcolor{gray!15}16 & \cellcolor{gray!15}\textbf{65.1} & \cellcolor{gray!15}\textbf{61.8} & \cellcolor{gray!15}\textbf{65.0} & \cellcolor{gray!15}\textbf{57.0} & \cellcolor{gray!15}46.3 & \cellcolor{gray!15}52.2 \\
\cmidrule(l){2-9}
\multirow{-6}{*}{\rotatebox[origin=c]{90}{\shortstack{\textit{Seen (4-way)}}}} & \textit{Chance} & \textit{--} & \textit{69.9} & \textit{30.9} & \textit{40.4} & \textit{43.1} & \textit{43.0} & \textit{50.9} \\
\midrule\midrule
 & Emb-Lens~\cite{fan2026visualtokens} & 0 & -- & 4.1 & 18.7 & 11.4 & 7.6 & 0.4 \\
 & \cellcolor{gray!15}DaP & \cellcolor{gray!15}1 & \cellcolor{gray!15}-- & \cellcolor{gray!15}\underline{11.1} & \cellcolor{gray!15}\underline{20.8} & \cellcolor{gray!15}\underline{12.1} & \cellcolor{gray!15}\underline{10.0} & \cellcolor{gray!15}\underline{0.8} \\
 & \cellcolor{gray!15}DaP & \cellcolor{gray!15}16 & \cellcolor{gray!15}-- & \cellcolor{gray!15}\textbf{22.3} & \cellcolor{gray!15}\textbf{44.1} & \cellcolor{gray!15}\textbf{23.5} & \cellcolor{gray!15}\textbf{21.1} & \cellcolor{gray!15}\textbf{7.3} \\
\cmidrule(l){2-9}
\multirow{-4}{*}{\rotatebox[origin=c]{90}{\shortstack{\textit{Unseen}}}} & \textit{Chance} & \textit{--} & \textit{--} & \textit{13.5} & \textit{30.1} & \textit{15.5} & \textit{12.8} & \textit{2.0} \\
\bottomrule
\end{tabular}
\caption{\textbf{Agreement (\%) with the model of other LVLMs}. Rows and columns as in \cref{tab:faithfulness}. Bold: best, underlined: second best (chance excluded).}
\label{tab:models_faith}
\end{table*}

\begin{table*}
\centering
\small\renewcommand{\arraystretch}{0.95}\setlength{\tabcolsep}{4pt}
\begin{tabular}{llc|c|cccc|c}
\toprule
 & Method & Shots & All Attributes & Color & Material & Shape & Activity & Names \\
\midrule
 & \# unseen targets & & -- & 27 & 10 & 17 & 21 & 1313 \\
\midrule[\heavyrulewidth]
\multicolumn{9}{c}{\textbf{Qwen3-VL-8B}} \\
\midrule
 & NCC & 16 & 39.2 & 17.5 & 27.5 & 24.0 & 10.6 & 10.5 \\
 & NCC & 256 & -- & 38.2 & 37.2 & 41.0 & 21.6 & \underline{16.0} \\
 & Probe & 16 & 39.4 & 21.8 & 26.5 & 24.2 & 11.1 & 10.2 \\
 & Probe & 256 & -- & 50.2 & 42.2 & 44.5 & \underline{33.2} & 8.5 \\
 & Emb-Lens & 0 & -- & 3.2 & 6.8 & 10.0 & 1.9 & 3.0 \\
 & \cellcolor{gray!15}DaP & \cellcolor{gray!15}1 & \cellcolor{gray!15}\underline{60.4} & \cellcolor{gray!15}\underline{57.0} & \cellcolor{gray!15}\underline{55.0} & \cellcolor{gray!15}\underline{78.5} & \cellcolor{gray!15}31.2 & \cellcolor{gray!15}6.0 \\
\multirow{-7}{*}{\rotatebox[origin=c]{90}{\shortstack{\textit{Seen}}}} & \cellcolor{gray!15}DaP & \cellcolor{gray!15}16 & \cellcolor{gray!15}\textbf{82.0} & \cellcolor{gray!15}\textbf{73.2} & \cellcolor{gray!15}\textbf{71.0} & \cellcolor{gray!15}\textbf{85.5} & \cellcolor{gray!15}\textbf{64.9} & \cellcolor{gray!15}\textbf{30.8} \\
\midrule\midrule
 & Emb-Lens & 0 & -- & 0.5 & 0.8 & 0.2 & 0.0 & 0.0 \\
 & \cellcolor{gray!15}DaP & \cellcolor{gray!15}1 & \cellcolor{gray!15}-- & \cellcolor{gray!15}\underline{11.2} & \cellcolor{gray!15}\underline{14.8} & \cellcolor{gray!15}\underline{19.5} & \cellcolor{gray!15}\underline{2.9} & \cellcolor{gray!15}\underline{2.2} \\
\multirow{-3}{*}{\rotatebox[origin=c]{90}{\shortstack{\textit{Unseen}}}} & \cellcolor{gray!15}DaP & \cellcolor{gray!15}16 & \cellcolor{gray!15}-- & \cellcolor{gray!15}\textbf{19.2} & \cellcolor{gray!15}\textbf{21.5} & \cellcolor{gray!15}\textbf{43.5} & \cellcolor{gray!15}\textbf{20.2} & \cellcolor{gray!15}\textbf{7.5} \\
\midrule[\heavyrulewidth]
\multicolumn{9}{c}{\textbf{InternVL3.5-8B}} \\
\midrule
 & NCC & 16 & 39.3 & 11.5 & 23.5 & 25.2 & 19.7 & 15.5 \\
 & NCC & 256 & -- & 33.5 & 35.0 & 37.5 & 36.1 & \textbf{18.0} \\
 & Probe & 16 & 39.3 & 15.0 & 23.2 & 20.2 & 17.3 & 13.8 \\
 & Probe & 256 & -- & 51.7 & \underline{44.0} & 36.5 & 37.0 & 9.0 \\
 & Emb-Lens & 0 & -- & 3.2 & 8.8 & 10.0 & 1.9 & 1.5 \\
 & \cellcolor{gray!15}DaP & \cellcolor{gray!15}1 & \cellcolor{gray!15}\underline{56.0} & \cellcolor{gray!15}\underline{59.8} & \cellcolor{gray!15}42.5 & \cellcolor{gray!15}\underline{54.0} & \cellcolor{gray!15}\underline{47.6} & \cellcolor{gray!15}5.0 \\
\multirow{-7}{*}{\rotatebox[origin=c]{90}{\shortstack{\textit{Seen}}}} & \cellcolor{gray!15}DaP & \cellcolor{gray!15}16 & \cellcolor{gray!15}\textbf{78.1} & \cellcolor{gray!15}\textbf{72.5} & \cellcolor{gray!15}\textbf{73.8} & \cellcolor{gray!15}\textbf{66.0} & \cellcolor{gray!15}\textbf{68.8} & \cellcolor{gray!15}\underline{17.2} \\
\midrule\midrule
 & Emb-Lens & 0 & -- & 0.2 & 0.8 & 0.5 & \underline{0.5} & 0.0 \\
 & \cellcolor{gray!15}DaP & \cellcolor{gray!15}1 & \cellcolor{gray!15}-- & \cellcolor{gray!15}\underline{10.8} & \cellcolor{gray!15}\underline{7.0} & \cellcolor{gray!15}\underline{7.5} & \cellcolor{gray!15}0.0 & \cellcolor{gray!15}\underline{1.0} \\
\multirow{-3}{*}{\rotatebox[origin=c]{90}{\shortstack{\textit{Unseen}}}} & \cellcolor{gray!15}DaP & \cellcolor{gray!15}16 & \cellcolor{gray!15}-- & \cellcolor{gray!15}\textbf{22.2} & \cellcolor{gray!15}\textbf{19.8} & \cellcolor{gray!15}\textbf{20.8} & \cellcolor{gray!15}\textbf{4.3} & \cellcolor{gray!15}\textbf{2.0} \\
\midrule[\heavyrulewidth]
\multicolumn{9}{c}{\textbf{Idefics3-8B}} \\
\midrule
 & NCC & 16 & 45.5 & 7.2 & 30.5 & 23.5 & 27.9 & 14.0 \\
 & NCC & 256 & -- & 22.2 & 43.2 & 35.8 & 38.9 & 15.0 \\
 & Probe & 16 & 45.7 & 7.8 & 30.5 & 23.8 & 28.8 & 14.5 \\
 & Probe & 256 & -- & 22.5 & 43.5 & 36.0 & 41.3 & 17.0 \\
 & Emb-Lens & 0 & -- & 4.0 & 11.5 & 16.5 & 2.9 & 1.2 \\
 & \cellcolor{gray!15}DaP & \cellcolor{gray!15}1 & \cellcolor{gray!15}\underline{82.7} & \cellcolor{gray!15}\underline{52.8} & \cellcolor{gray!15}\underline{65.8} & \cellcolor{gray!15}\underline{60.5} & \cellcolor{gray!15}\underline{67.8} & \cellcolor{gray!15}\underline{18.8} \\
\multirow{-7}{*}{\rotatebox[origin=c]{90}{\shortstack{\textit{Seen}}}} & \cellcolor{gray!15}DaP & \cellcolor{gray!15}16 & \cellcolor{gray!15}\textbf{93.1} & \cellcolor{gray!15}\textbf{77.0} & \cellcolor{gray!15}\textbf{82.2} & \cellcolor{gray!15}\textbf{88.8} & \cellcolor{gray!15}\textbf{92.3} & \cellcolor{gray!15}\textbf{48.2} \\
\midrule\midrule
 & Emb-Lens & 0 & -- & 1.2 & 0.5 & 1.2 & 0.0 & 0.2 \\
 & \cellcolor{gray!15}DaP & \cellcolor{gray!15}1 & \cellcolor{gray!15}-- & \cellcolor{gray!15}\underline{11.2} & \cellcolor{gray!15}\underline{27.5} & \cellcolor{gray!15}\underline{16.0} & \cellcolor{gray!15}\underline{27.4} & \cellcolor{gray!15}\underline{1.2} \\
\multirow{-3}{*}{\rotatebox[origin=c]{90}{\shortstack{\textit{Unseen}}}} & \cellcolor{gray!15}DaP & \cellcolor{gray!15}16 & \cellcolor{gray!15}-- & \cellcolor{gray!15}\textbf{22.8} & \cellcolor{gray!15}\textbf{55.0} & \cellcolor{gray!15}\textbf{45.2} & \cellcolor{gray!15}\textbf{61.5} & \cellcolor{gray!15}\textbf{8.8} \\
\midrule[\heavyrulewidth]
\multicolumn{9}{c}{\textbf{Ministral 3 8B}} \\
\midrule
 & NCC & 16 & 45.8 & 13.2 & 22.5 & 20.5 & 12.5 & 13.0 \\
 & NCC & 256 & -- & 27.3 & \underline{33.5} & 36.2 & 17.8 & 12.2 \\
 & Probe & 16 & 45.9 & 13.5 & 22.5 & 20.2 & 12.5 & 13.2 \\
 & Probe & 256 & -- & 28.2 & 33.0 & \underline{36.8} & 19.2 & \underline{14.0} \\
 & Emb-Lens & 0 & -- & 5.2 & 11.8 & 10.8 & 4.3 & 1.0 \\
 & \cellcolor{gray!15}DaP & \cellcolor{gray!15}1 & \cellcolor{gray!15}\underline{62.7} & \cellcolor{gray!15}\underline{31.2} & \cellcolor{gray!15}29.5 & \cellcolor{gray!15}25.0 & \cellcolor{gray!15}\underline{40.4} & \cellcolor{gray!15}2.0 \\
\multirow{-7}{*}{\rotatebox[origin=c]{90}{\shortstack{\textit{Seen}}}} & \cellcolor{gray!15}DaP & \cellcolor{gray!15}16 & \cellcolor{gray!15}\textbf{85.3} & \cellcolor{gray!15}\textbf{69.8} & \cellcolor{gray!15}\textbf{59.2} & \cellcolor{gray!15}\textbf{52.0} & \cellcolor{gray!15}\textbf{67.8} & \cellcolor{gray!15}\textbf{19.0} \\
\midrule\midrule
 & Emb-Lens & 0 & -- & 0.8 & 1.5 & 0.2 & 0.0 & 0.5 \\
 & \cellcolor{gray!15}DaP & \cellcolor{gray!15}1 & \cellcolor{gray!15}-- & \cellcolor{gray!15}\underline{4.5} & \cellcolor{gray!15}\underline{3.2} & \cellcolor{gray!15}\underline{2.5} & \cellcolor{gray!15}\underline{1.9} & \cellcolor{gray!15}\underline{1.0} \\
\multirow{-3}{*}{\rotatebox[origin=c]{90}{\shortstack{\textit{Unseen}}}} & \cellcolor{gray!15}DaP & \cellcolor{gray!15}16 & \cellcolor{gray!15}-- & \cellcolor{gray!15}\textbf{9.0} & \cellcolor{gray!15}\textbf{18.0} & \cellcolor{gray!15}\textbf{13.8} & \cellcolor{gray!15}\textbf{9.1} & \cellcolor{gray!15}\textbf{4.8} \\
\bottomrule
\end{tabular}
\caption{\textbf{Steering success (\%) at $\alpha=1$ of other LVLMs}. Rows and columns as in \cref{tab:steering}. Probe uses the regularization chosen on the validation set for Qwen3-VL-4B. \textbf{Bold}: best, \underline{underlined}: second best.}
\label{tab:models_steer}
\end{table*}

\clearpage

\onecolumn

\section{Steering in Another Language} \label{sec:app:steering_language}
\Cref{tab:steer_zh} reports the steering success when the steered model is asked the question in Chinese.
{\makeatletter\def\fps@table{!htbp}\makeatother
\begin{table*}
\centering
\small\renewcommand{\arraystretch}{1}\setlength{\tabcolsep}{4pt}
\begin{tabular}{llc|cccc|c}
\toprule
 & Method & Shots & Color & Material & Shape & Activity & Names \\
\midrule
 & NCC & 16 & 19.8 & 27.0 & 21.8 & 14.9 & 21.3 \\
 & NCC & 256 & 39.8 & 42.3 & 35.0 & 31.7 & \underline{28.0} \\
 & Probe & 16 & 26.0 & 22.5 & 23.3 & 15.4 & 9.7 \\
 & Probe & 256 & 57.0 & 50.7 & 42.8 & 44.2 & 5.0 \\
 & Emb-Lens & 0 & 6.8 & 15.0 & 15.3 & 4.3 & 4.7 \\
 & \cellcolor{gray!15}DaP & \cellcolor{gray!15}1 & \cellcolor{gray!15}\underline{75.2} & \cellcolor{gray!15}\underline{76.5} & \cellcolor{gray!15}\underline{82.0} & \cellcolor{gray!15}\underline{57.2} & \cellcolor{gray!15}27.5 \\
\multirow{-7}{*}{\rotatebox[origin=c]{90}{\shortstack{\textit{English}}}} & \cellcolor{gray!15}DaP & \cellcolor{gray!15}16 & \cellcolor{gray!15}\textbf{83.5} & \cellcolor{gray!15}\textbf{82.2} & \cellcolor{gray!15}\textbf{86.3} & \cellcolor{gray!15}\textbf{80.3} & \cellcolor{gray!15}\textbf{43.8} \\
\midrule\midrule
 & NCC & 16 & 26.0 & 49.8 & 34.0 & 24.5 & 6.8 \\
 & NCC & 256 & 49.3 & 70.7 & 44.5 & 41.8 & \underline{10.8} \\
 & Probe & 16 & 29.5 & 36.5 & 31.5 & 21.2 & 1.8 \\
 & Probe & 256 & 65.2 & 76.7 & 46.3 & 49.0 & 0.0 \\
 & Emb-Lens & 0 & 11.2 & 28.5 & 20.0 & 3.8 & 0.0 \\
 & \cellcolor{gray!15}DaP & \cellcolor{gray!15}1 & \cellcolor{gray!15}\underline{73.0} & \cellcolor{gray!15}\underline{88.2} & \cellcolor{gray!15}\underline{77.0} & \cellcolor{gray!15}\underline{58.2} & \cellcolor{gray!15}6.0 \\
\multirow{-7}{*}{\rotatebox[origin=c]{90}{\shortstack{\textit{Chinese}}}} & \cellcolor{gray!15}DaP & \cellcolor{gray!15}16 & \cellcolor{gray!15}\textbf{83.0} & \cellcolor{gray!15}\textbf{93.8} & \cellcolor{gray!15}\textbf{83.5} & \cellcolor{gray!15}\textbf{86.1} & \cellcolor{gray!15}\textbf{11.7} \\
\bottomrule
\end{tabular}
\caption{\textbf{Steering success (\%) at $\alpha=1$ with a question in Chinese}. The directions are those of \cref{tab:steering}, estimated with the English prompt. English rows ask the steering question of \cref{tab:steering}. Chinese rows ask it in Chinese and read the answer on the Chinese words of the seen concepts. Seen targets only. \textbf{Bold}: best, \underline{underlined}: second best.}
\label{tab:steer_zh}
\end{table*}
}
\twocolumn
\clearpage

\section{Failing Cases} \label{sec:app:failing}
Here we show some of the failing cases we encountered using DaP.

\subsection{Nonlinearly Decodable}

The LVLM answers some concepts correctly that no linear probe decodes from $\hat{\mathbf{v}}$.
\Cref{tab:nonlinear} lists the five concepts of All Attributes with the largest gap between the accuracy of the LVLM and the best accuracy among NCC, the trained probe and DaP, at 16 shots.
The LVLM reaches at least 93\% on each of them, while NCC, the trained probe and DaP reach at most 67\%.
Three of these concepts describe a part of a person, such as the skin, the hair or the face, which covers only a few visual tokens of $\mathcal{M}$.
Averaging the visual tokens of $\mathcal{M}$ into $\hat{\mathbf{v}}$ may thus dilute the information that the LVLM reads from these few tokens.

\begin{table}[t]
\centering
\small\renewcommand{\arraystretch}{1}\setlength{\tabcolsep}{5pt}
\begin{tabular}{l|ccc|c}
\toprule
Concept & NCC & Probe & DaP & \textit{Model} \\
\midrule
\concept{caucasian} & 51.4 & 51.4 & 59.5 & \textit{93.2} \\
\concept{black haired} & 56.1 & 58.5 & 61.0 & \textit{93.9} \\
\concept{laughing} & 63.4 & 61.0 & 65.9 & \textit{95.1} \\
\concept{pale blue} & 65.2 & 65.2 & 65.2 & \textit{93.5} \\
\concept{bright blue} & 64.8 & 58.0 & 67.0 & \textit{94.3} \\
\bottomrule
\end{tabular}
\caption{\textbf{Nonlinearly decodable concepts.} Accuracy (\%) at 16 shots of the five concepts of All Attributes with the largest gap between the LVLM and the best linear probe.}
\label{tab:nonlinear}
\end{table}

\subsection{Prompt Precision}

DaP reads the visual embeddings through the question as the LVLM understands it, which may differ from the intended question.
For the state attribute type of VAW (\eg, \concept{open}, \concept{empty}, \concept{wet}), the LVLM reads the question \prompt{What state is this in?} as a question about a geographic state.
It then answers \tok{none}, \tok{not} or US states such as \tok{California}, and its open accuracy falls to 0.1\%.
DaP estimated with this question inherits this reading.
The bias gate then holds US states such as \tok{California}, \tok{Texas} and \tok{Florida} among its first ten tokens, and gated DaP answers US states such as \tok{Hawaii} on some of the training instances.
With the prompt of the attribute types, \prompt{Which word best describes the state of the \ph{object name}?}, the bias gate contains no US state.
Thus, DaP is only as precise as the prompt, since it reads the attribute that the LVLM understands.

\subsection{Words with Two Senses}

The word \concept{orange} names both a color and a fruit.
The token \tok{orange} thus ranks 27th among the $k = 200$ tokens of the bias gate for colors and 113th among the $k = 5000$ tokens of the bias gate for object names, so the gate keeps it for both questions.
When asked for the object name, DaP answers \tok{orange} on 20\% of the objects labeled with the color orange that are not oranges, against 4\% for the LVLM.
The bias gate cannot separate the two senses of the token, so DaP reads the orange color of an object as its name.

\begin{table*}
\centering
\small\renewcommand{\arraystretch}{1}\setlength{\tabcolsep}{4pt}
\begin{tabular}{llcc|c|cccc|c}
\toprule
 & Method & Shots & Labels & All Attributes & Color & Material & Shape & Activity & Names \\
\midrule
 & DaP LRC-Style~\cite{chanin2024lrc} & 1 & \cmark & -- & 38.2 & 62.7 & 31.4 & 47.6 & 51.2 \\
 & DaP LRC-Style~\cite{chanin2024lrc} & 16 & \cmark & -- & 64.5 & 73.0 & 57.5 & 79.9 & 57.6 \\
 & \cellcolor{gray!15}DaP & \cellcolor{gray!15}1 & \cellcolor{gray!15}\xmark & \cellcolor{gray!15}\underline{63.6} & \cellcolor{gray!15}\underline{69.3} & \cellcolor{gray!15}\underline{73.5} & \cellcolor{gray!15}\underline{63.3} & \cellcolor{gray!15}\underline{82.0} & \cellcolor{gray!15}\underline{77.6} \\
\multirow{-4}{*}{\rotatebox[origin=c]{90}{\shortstack{\textit{Seen (4-way)}}}} & \cellcolor{gray!15}DaP & \cellcolor{gray!15}16 & \cellcolor{gray!15}\xmark & \cellcolor{gray!15}\textbf{73.7} & \cellcolor{gray!15}\textbf{72.5} & \cellcolor{gray!15}\textbf{78.9} & \cellcolor{gray!15}\textbf{65.5} & \cellcolor{gray!15}\textbf{84.9} & \cellcolor{gray!15}\textbf{78.7} \\
\midrule\midrule
 & \# classes &  &  & -- & 27 & 10 & 17 & 21 & 1313 \\
 & DaP LRC-Style~\cite{chanin2024lrc} & 1 & \cmark & -- & -- & -- & -- & -- & -- \\
 & DaP LRC-Style~\cite{chanin2024lrc} & 16 & \cmark & -- & -- & -- & -- & -- & -- \\
 & \cellcolor{gray!15}DaP & \cellcolor{gray!15}1 & \cellcolor{gray!15}\xmark & \cellcolor{gray!15}-- & \cellcolor{gray!15}\underline{44.3} & \cellcolor{gray!15}\underline{59.0} & \cellcolor{gray!15}\underline{52.9} & \cellcolor{gray!15}\underline{23.6} & \cellcolor{gray!15}\underline{20.0} \\
\multirow{-5}{*}{\rotatebox[origin=c]{90}{\shortstack{\textit{Unseen}}}} & \cellcolor{gray!15}DaP & \cellcolor{gray!15}16 & \cellcolor{gray!15}\xmark & \cellcolor{gray!15}-- & \cellcolor{gray!15}\textbf{51.2} & \cellcolor{gray!15}\textbf{65.8} & \cellcolor{gray!15}\textbf{53.9} & \cellcolor{gray!15}\textbf{42.3} & \cellcolor{gray!15}\textbf{24.3} \\
\bottomrule
\end{tabular}
\caption{Top-1 accuracy (\%) of DaP and DaP LRC-Style, the LRC~\cite{chanin2024lrc} concept directions computed with the Jacobian of DaP (rank-128 pseudo-inverse, class centroids of the estimation images). Columns as in Table~\ref{tab:benchmark}. DaP LRC-Style needs the labels of the estimation images, so it cannot read unseen concepts. --: not available. Bold: best, underlined: second best.}
\label{tab:lrc}
\end{table*}

\subsection{Tokenization}

Open DaP reads a single token, the first one of the answer.
\concept{riding horse} and \concept{riding bike} both start with \tok{riding}, so open DaP cannot tell them apart.
Similarly, \concept{maroon} has no token of its own and starts with \tok{mar}, so DaP reads \concept{maroon} through \tok{mar}.
DaP steering toward \concept{maroon} at $\alpha = 1$ never succeeds, and the LVLM answers \tok{brown} for 11 of the 19 steered objects.

\section{Previous Linearization Methods} \label{sec:app:linearization}

In this section, we dive deeper into the comparison of our proposed framework against past work on linearization in language models. As the objective of our work is not to propose a novel linearization method but to bring the tool to the vision community, this is not part of the main paper. 

LRE~\cite{hernandez2024lre} approximates how an LLM decodes a relation (\eg, \emph{plays the instrument}) from the hidden state of its subject (\eg, \emph{Miles Davis}).
It averages the Jacobian and the bias of a few examples, from the last subject token at a middle layer to the last layer, and reads the object with the unembedding matrix.
LRC~\cite{chanin2024lrc} inverts the Jacobian of LRE to get one direction per concept (\eg, \concept{located in France}).
J-Lens~\cite{gurnee2026workspace} reads the hidden states of any layer with the unembedding matrix through the Jacobian only, without a bias.
However, the setup in which the Jacobian is estimated is completely different, as it is estimated to predict future tokens on a corpus of next-token predictions.
We detail here the four ways in which DaP adapts these methods to visual representations.

\paragraph{(1) Questions instead of completions.}
We propose a framework in which the prompt is an open or closed question, so that any concept can be probed.
LRE and LRC use sentence completion, which is restrictive, since the concept must be the next token of the sentence.
A question is practical, since probing a new concept only requires asking about it, which makes probing much easier.
The closed question in particular has no equivalent in previous linearization methods. The closed yes-no question can query the direction of multi-token concepts and of any concept in general.

\paragraph{(2) Objects that span several visual tokens.}
LRE and LRC map the hidden state of a single text token at a middle layer.
DaP maps the visual tokens of the object, averaged over the object region $\mathcal{M}$, from the input of the LLM Decoder to its last layers.
Since the modality, the input and the depth of the map all differ, the results of LRE and LRC on text are hard to directly compare with ours.

\paragraph{(3) The unembedding matrix as concept anchors.}
LRC maps the mean hidden state of the examples of each concept back through the inverse of the Jacobian, so it needs labeled examples of every concept.
To steer toward a concept, LRE needs the hidden state of an example of that concept.
Instead, we use the unembedding matrix as a dictionary of concept direction anchors, which reduces the reliance on labels.
DaP thus needs no labels at estimation and generalizes to unseen concepts.
\Cref{tab:lrc} compares DaP with DaP LRC-Style, which replaces the rows of $W_U$ by the mean $\mathbf{h}$ of the estimation images of each concept, as in LRC, and keeps everything else of DaP.

\paragraph{(4) The bias as a gate.}
LRE and LRC use the bias.
Recently, J-Lens~\cite{gurnee2026workspace} showed that the bias is not needed, especially to read out from the full vocabulary.
To improve performance, we add the bias back as a gate on the vocabulary in open classification.




\end{document}